\documentclass[final,3p,times,authoryear]{elsarticle}
\pdfoutput=1

\usepackage{hyperref}

\usepackage[symbol]{footmisc}

\usepackage[inline,shortlabels]{enumitem}
\usepackage{tablefootnote}
\usepackage{pdflscape}
\usepackage{makecell}
\usepackage{cellspace}
\usepackage{lmodern}
\usepackage{caption}
\usepackage{tikz} 
\usepackage[textsize=small]{todonotes}
\usepackage{multirow}
\usepackage{pifont}
\usepackage{booktabs}
\usepackage{subcaption}
\usepackage{bbm}
\usepackage{float}
\usepackage{longtable}
\usepackage{tabularx}
\usepackage{geometry}
\usepackage{bm}
\usepackage{siunitx}
\usepackage[normalem]{ulem}

\usepackage{lineno}
\usepackage{amsmath} 
\usepackage{etoolbox} 

\newcommand*\linenomathpatch[1]{%
  \cspreto{#1}{\linenomath}%
  \cspreto{#1*}{\linenomath}%
  \cspreto{end#1}{\endlinenomath}%
  \cspreto{end#1*}{\endlinenomath}%
}
\linenomathpatch{equation}
\linenomathpatch{gather}
\linenomathpatch{align}

\graphicspath{{figures/}}

\usepackage{xcolor}

\usepackage{algorithm}
\usepackage{algpseudocode}

\algnewcommand{\LeftComment}[1]{\Statex \(\triangleright\) #1}

\usepackage{colortbl}
\usetikzlibrary{calc}
\usepackage{pgfplots}
\pgfplotsset{/pgfplots/error bars/error bar style={thick}}  
\pgfplotsset{compat = 1.3}

\usepackage{chngcntr} 

\definecolor{darkspringgreen}{rgb}{0.09, 0.45, 0.27}
\colorlet{presentattr}{darkspringgreen!40}
\colorlet{ourdataset}{darkspringgreen!30}

\definecolor{deepcarmine}{rgb}{0.66, 0.13, 0.24}
\colorlet{absentattr}{deepcarmine!40}

\definecolor{Gainsboro}{rgb}{0.86,0.86,0.86}
\colorlet{graycell}{Gainsboro}

\newcommand{\datasetname}{DWIE}
\newcommand{\cmark}{\ding{51}}%
\newcommand{\xmark}{\ding{55}}%
\newcommand{\propformat}[1]{\texttt{#1}}

\newcommand{\eg}{e.g., }
\newcommand{\cf}{cf.\ }
\newcommand{\ie}{i.e., }

\newcommand{\topicname}[1]{\textbf{#1}}
\newcommand{\entityname}[1]{\textbf{#1}}

\newcommand{\figref}[1]{Fig.~\ref{#1}}    
\newcommand{\Figref}[1]{Figure~\ref{#1}}  
\newcommand{\tabref}[1]{Table~\ref{#1}}
\newcommand{\Tabref}[1]{Table~\ref{#1}}
\newcommand{\Algref}[1]{Algorithm~\ref{#1}}
\newcommand{\secref}[1]{Section~\ref{#1}}
\newcommand{\equref}[1]{Eq.~(\ref{#1})}
\newcommand{\eqsref}[2]{Eq.~(\ref{#1})--(\ref{#2})}
\newcommand{\appref}[1]{\ref{#1}} 

\newcommand{\relation}[3]{$\textsf{#1}\langle\textit{#2},$ $\textit{#3}\rangle$}

\newcommand{\relationalign}[3]{\textsf{#1}\langle\textit{#2}, \textit{#3}\rangle}
\newcommand{\relationalignsingle}[2]{\textsf{#1}\langle\textit{#2}\rangle}

\newcommand{\boldpartitle}[1]{\noindent \textbf{#1} ---}

\DeclareMathAlphabet\mathbfcal{OMS}{cmsy}{b}{n}

\makeatletter
\let\UrlSpecialsOld\UrlSpecials
\def\UrlSpecials{\UrlSpecialsOld\do\/{\Url@slash}\do\_{\Url@underscore}}%
\def\Url@slash{\@ifnextchar/{\kern-.11em\mathchar47\kern-.2em}%
    {\kern-.0em\mathchar47\kern-.08em\penalty\UrlBigBreakPenalty}}
\def\Url@underscore{\nfss@text{\leavevmode \kern.06em\vbox{\hrule\@width.3em}}}
\makeatother

\modulolinenumbers[1]

\journal{Journal of Information Processing and Management}

\biboptions{sort}

\begin{document}

\begin{frontmatter}

\title{{\datasetname}: an entity-centric dataset for \\multi-task document-level information extraction.}

\author[]{Klim Zaporojets\corref{cor1}\fnref{fn1}}
\ead{klim.zaporojets@ugent.be}
\author[]{Johannes Deleu\fnref{fn1}}
\ead{johannes.deleu@ugent.be}
\author[]{Chris Develder\fnref{fn1}}
\ead{chris.develder@ugent.be}
\author[]{Thomas Demeester\fnref{fn1}}
\ead{thomas.demeester@ugent.be}
\cortext[cor1]{Corresponding author}

\address{Ghent University -- imec, IDLab, Department of Information Technology,\\
Technologiepark Zwijnaarde 15, 9052 Ghent, Belgium
}
\fntext[fn1]{\textit{URL:} \url{https://ugentt2k.github.io/}}

\begin{abstract}
This paper presents \datasetname, the `Deutsche Welle corpus for Information Extraction', 
a newly created 
multi-task dataset that combines four main Information Extraction (IE) annotation subtasks:
\begin{enumerate*}[(i)]
\item Named Entity Recognition (NER),
\item Coreference Resolution,
\item Relation Extraction (RE), and
\item Entity Linking.
\end{enumerate*}
DWIE is conceived as an \emph{entity-centric} dataset that describes interactions and properties of conceptual entities on %
the level of the complete document.
This contrasts with currently dominant \emph{mention-driven} approaches 
that start from the detection and classification of 
named entity mentions in
individual sentences.
Further, {\datasetname} presented two main challenges when building and evaluating IE models for it. \textit{First}, the use of traditional 
mention-level 
evaluation metrics for NER and RE tasks on entity-centric \datasetname~dataset can result in measurements dominated by predictions on more frequently mentioned entities. We tackle this issue by proposing a new entity-driven metric that takes into account the number of mentions that compose each of the predicted and ground truth entities. 
\textit{Second}, the document-level multi-task annotations require the models to transfer information between entity mentions located in different parts of the document, as well as between different tasks, in a joint learning setting. To realize this, we propose to use graph-based neural message passing techniques between document-level mention spans. Our experiments show an improvement of up to 5.5 F$_\text{1}$ percentage points when incorporating neural graph propagation into our joint model. 
This demonstrates \datasetname's potential to stimulate further research 
in graph neural networks for  representation learning in multi-task IE. 
We make DWIE publicly available at \url{https://github.com/klimzaporojets/DWIE}.
\end{abstract}


\end{frontmatter}


\section{Introduction}
\label{sec:introduction}
Information Extraction (IE) plays a fundamental role as a backbone component in many downstream applications. For example, 
an application
such as question answering
may be improved by relying on relation extraction (RE) \citep{yu2017improved,hu2019language}, coreference resolution \citep{gao2019interconnected,bhattacharjee2020investigating}, named entity recognition (NER) \citep{molla2006named,singh2018reinvent}, and entity linking (EL) 
\citep{chen2017reading,broscheit2019investigating} 
 components. 
This also holds for other applications such as
personalized news recommendation \citep{wang2018dkn,karimi2018news,wang2019multi}, 
fact checking \citep{thorne2018automated,zhang2020overview}, 
opinion mining \citep{sun2017review}, 
semantic search \citep{cifariello2019wiser}, and 
conversational agents \citep{roller2020recipes}. 
The last decade has shown a growing interest in IE datasets suitably annotated for
developing multi-task models where each of the tasks (\eg NER, RE, etc.) would benefit from the interaction with (an)other task(s) \citep{bekoulis2018joint,fei2020boundaries,lee2017end,lee2018higher,luan2019general}, to boost their performance.
However, the currently widely used IE datasets to build such multi-task models exhibit three major limitations. 
\textit{First}, the annotation schema adopted in most of 
these datasets
is mention-driven, focusing on annotating elements (\eg relations, entity types) that involve specific entity mentions explicitly mentioned in the text. 
This produces very localized annotations (\eg sentence-based relations between entity mentions) that do not reflect meaning that can be inferred on a more general 
document-level.
\textit{Second}, the number of annotated extraction tasks in most of the IE datasets is rather limited. Most of them focus on a single or at most a few different tasks. 
Furthermore, some other datasets, including the well-known TAC-KBPs \citep{ji2010overview,ellis2014overview,ji2015overview,ellis2015overview,ji2017overview}, use different non-overlapping corpora for each of the tracks that group a few related tasks. Consequently, current models addressing multiple IE tasks together often use multi-tasking (with different datasets per task) rather than really joint modeling approaches.
\textit{Finally}, the annotation of currently widely used IE datasets is driven by either relying on a priori defined annotation schemas \citep{doddington2004automatic, walker2006ace,song2015light,augenstein2017semeval,zhang2017position,hendrickx2010semeval} or on distantly supervised labeling techniques \citep{han2018fewrel,yao2019docred,riedel2010modeling,quirk2017distant,peng2017cross}. In consequence, the resulting annotations
are not necessarily representative of the actual information contained in the 
annotated 
corpus.

\begin{figure}[t]
\centering
\includegraphics[width=1.0\columnwidth, trim={0cm 0cm 0cm 0cm},clip]{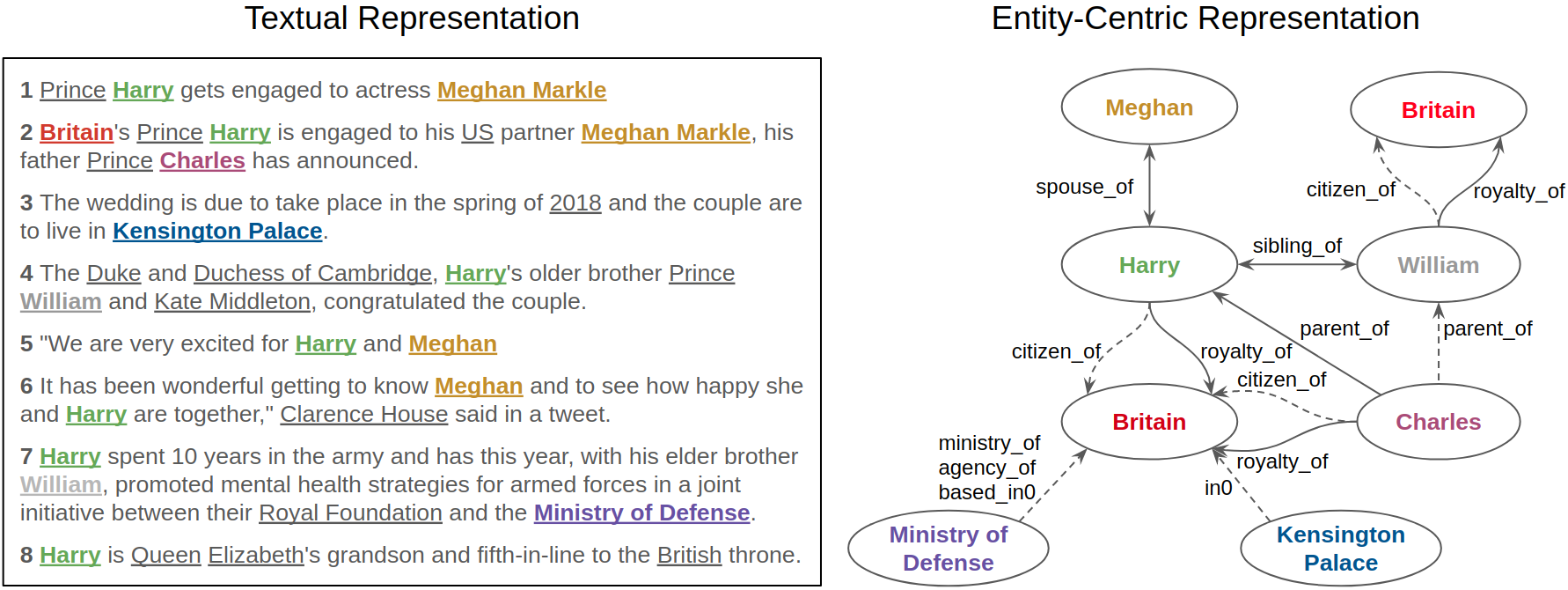}
\captionsetup{singlelinecheck=off}
\caption[test]{An example from the \datasetname~dataset with entity mentions underlined.
We show 8 of the 29 entities in the graph on the right.
It illustrates the relations that can be derived from the content of the article. The relations that are explicitly mentioned in the text (trigger-based) are depicted by solid arrows. Conversely, the relations that are implicit and/or need the whole document context (document-based) to be derived are represented by dashed arrows.}
\label{fig:example}
\end{figure}

In this work, we tackle the aforementioned limitations 
of IE datasets 
by introducing a new 
dataset named DWIE. It consists of 802 general news articles in English, selected randomly from a corpus collected from Deutsche Welle\footnote{\url{https://www.dw.com}} between 2002 and 2018, as part of the CPN project.\footnote{\url{https://www.projectcpn.eu}} We focus on annotating four main IE tasks: \begin{enumerate*}[(i)]
\item Named Entity Recognition (NER),
\item Coreference Resolution,
\item Relation Extraction (RE), and
\item Entity Linking.\footnote{The linking is done to Wikipedia version 20181115.}
\end{enumerate*}
\Figref{fig:example} 
shows an example snippet from the \datasetname~corpus.
We adopt an \textit{entity-centric} approach where all annotations (\ie for NER, RE and Entity Linking tasks) are made on the entity\footnote{Also referred to as \textit{entity cluster} or just \textit{cluster}.} level. Each of the entities is composed by the coreferenced entity mentions from the entire document (\eg the entity \textit{Meghan} in \figref{fig:example} clusters the entity mentions ``Meghan Markle'' and ``Meghan'' across the whole document). 
This entity-centric approach contrasts with mention-driven annotations in widely used IE datasets \citep{doddington2004automatic,ji2015overview,song2015light,han2018fewrel,zhang2017position,hendrickx2010semeval,luan2018multi} where the annotation process is 
biased towards 
considering only local 
explicit textual evidence to annotate elements such as relations and entity types (\eg the relation \relation{spouse\_of}{Meghan}{Harry} that can be extracted from the 1st sentence in \figref{fig:example}). Consequently, our {\datasetname} dataset paves the way for research on more complex document-level reasoning that goes beyond only the local textual context directly surrounding individual entity mentions. For example, consider the relation \relation{ministry\_of}{Ministry of Defense}{Britain} in \figref{fig:example}: while the text of the document does not directly state such a relation, it can be deduced from a more general document-level entity-centric vision of the article, \ie combining the information involving the entities \textit{Ministry of Defense} and \textit{Harry} in sentence 7 with the one involving \textit{Britain} and \textit{Harry} in sentence 2.
Finally, the entity-centric approach provides entity linking annotations that are consistent across the document: by clustering mentions of the same entity, and then providing links to the Wikidata KB (or NIL if the entity does not appear there) for the whole cluster at once, we limit annotation errors or accidental inconsistencies (in the linking itself, but also in terms of NER labels).
To our knowledge, DWIE is the first dataset with this level of conceptual consistency over the considered information extraction tasks. We therefore expect that the dataset
will play a key role in advancing research exploring potential benefits of
\begin{enumerate*}[(i)]
    \item entity-level information extraction in terms of reducing potential inconsistent decisions (within EL across multiple mentions, as well as across multiple tasks), and
    \item using entity-centric information stored in a KB to complement the 
    otherwise exclusively
    text-dependent IE tasks such as NER, RE, and coreference resolution. 
\end{enumerate*}

Additionally, we use a bottom-up, data-driven annotation approach where we manually define our annotations (\eg in terms of the entity and relation types) to maximally reflect the information of the corpus at hand. 
Currently dominant datasets are driven by distant supervision and executed top-down, by which we mean that the selection of entity and relation types is a priori defined and limited in coverage (\ie the raw data potentially contains other types that thus remain un-annotated).
Conversely, we do not a priori limit the entity and relation types to annotate, but adopt a bottom-up approach driven by the data itself.
Our proposed bottom-up approach encompasses a three-pass annotation procedure where we use the first exploratory annotation pass to derive the main annotation types (annotation schema) from the corpus, and the next two passes to perform schema-driven annotations and refine them by carrying out an additional parallel annotation of the corpus for fixing errors inferred from inter-annotator inconsistencies.

Besides the dataset itself, we also contribute empirical modeling results 
to address the aforementioned IE tasks. 
Our goal is to study two important properties
that are inherent to \datasetname.
The first key property is the need for \textit{long-range contextual information sharing}
to make document-level predictions involving entities whose mentions are located in different parts of the document. 
The second key property 
involves the \textit{joint interaction between tasks} where the information obtained in one task can help to solve another task. For example, in \figref{fig:example} knowing the types of entities (which involves NER and coreference tasks) \textit{Britain} and \textit{Kensington Palace} can boost the performance of the relation extraction task by limiting the number of possible relation types between these two entities (\eg \textsf{ministry\_of} but not \textsf{citizen\_of}).
In order to study the impact of these two phenomena inherent to our {\datasetname} dataset on the final results, we experiment with neural graph-based models \citep{li2015gated,xu2018powerful,wu2020comprehensive}. These models allow message passing between local contextual encodings, making it possible to measure the impact of local contextual information sharing both on a more general document level and across the tasks.
Furthermore, previous work already has shown the positive effect of using graph-based information passing techniques on single tasks \citep{lee2018higher, kantor2019coreference}, and between tasks \citep{luan2019general, wadden2019entity, fei2020boundaries, fu2019graphrel} on mention-driven datasets. We expand this work even further by extending these models to be used on the entity-centric, document-level {\datasetname} dataset. More specifically, we experiment with both single-task (\secref{sec:single_task_models}) as well as joint (\secref{sec:joint_training}) models to study the effect of contextual information propagation in 
single task and
joint settings.
Additionally, for the NER and RE tasks, we propose a new entity-centric evaluation metric that not only considers the predictions on separate entity mentions (as is  done in related IE datasets), but also accounts for the impact of the predictions on entity cluster level.

In summary, the main objective that we address in the current paper is to introduce an entity-centric multi-task IE dataset that covers different related tasks on a document level as well as provides a connection with external structured knowledge (through the entity linking task). Furthermore, we aim to explore how neural graph-based models can boost the performance 
by enabling local contextual information propagation across the document (single-task models) and between different tasks (joint models). The results presented in this paper  
suggest that, while challenging, \datasetname~opens up new possibilities of research in the domain of 
joint entity-centric information extraction methods.
The main contributions of our work are that: 
\begin{enumerate}[(1)]
    \item We construct a self-contained dataset (\secref{sec:annotation}) with joint annotations for four basic information extraction tasks (NER, entity linking, coreference resolution, and 
    RE), 
    that provide entity-centric document-level annotations (as opposed to typical 
    mention-driven
    sentence-level annotations for, 
    \eg RE) connecting unstructured (text) and structured (KB) information sources. 
    \item We introduce a data-driven, bottom-up three-pass annotation approach complemented by context-based logical rules to build such dataset (\secref{sec:annotation}).
    \item We propose a new evaluation metric for the NER and RE tasks (\secref{sec:metrics}), in line with the entity-centric nature of \datasetname. 
    \item We extend the competitive graph-based neural IE model DyGIE \citep{luan2019general}  
    for the four IE tasks in {\datasetname}  
    (\secref{sec:models}) and provide source code for NER, coreference resolution, and 
    RE. Furthermore, we introduce a new latent attention-driven \propformat{AttProp} graph propagation method and show its advantages in both single and joint model settings.
    The experimental results (\secref{sec:results}) demonstrate the potential of such neural graph based models. 
\end{enumerate}

\section{Related Work}
\label{sec:related}

This section summarizes the overview of related datasets (\secref{sec:related_datasets}), and explores the differences between our newly created {\datasetname} and other similar datasets widely used by the scientific community.
The main qualitative differences are presented in \Tabref{tab:quali_compare}, while the quantitative comparison is provided in \Tabref{tab:quanti_compare}. 
Next, we describe the current trends 
in IE to solve the tasks included in \datasetname, and compare them to our proposed approach (\secref{sec:recent_advances}). Finally, we discuss currently used metrics to evaluate model performance on IE datasets and introduce some challenges in applying them to measuring the performance on \datasetname~(\secref{sec:metrics_and_evaluation}). 

\subsection{Related Datasets}
\label{sec:related_datasets}

Most of IE datasets have focused on a single task, making it very challenging to develop systems that  jointly train for different annotation subtasks on a single corpus. 
Well-known single-task datasets include
\begin{enumerate*}[(i)]
    \item \emph{for NER:} CoNLL-2003 \citep{sang2003introduction} and WNUT 2017 \citep{derczynski2017results},
    \item \emph{for relation extraction:} Semeval-2010 T8 \citep{hendrickx2010semeval}, TACRED \citep{zhang2017position} and FewRel \citep{han2018fewrel},
    \item \emph{for entity linking:} IITB \citep{kulkarni2009collective}, CoNLL-YAGO \citep{hoffart2011robust}, and WikilinksNED \citep{eshel2017named}, and
    \item \emph{for coreference resolution:} CoNLL-2012 \citep{pradhan2012conll} and GAP \citep{webster2018mind}.
\end{enumerate*}
Conversely, in this work we propose a multi-task dataset as a single corpus annotated with different information extraction layers: named entities, mention clustering in entities (\ie coreference), relations between entity clusters of mentions, and entity linking.
We further complement our dataset with additional tasks such as document classification and keyword extraction. It is worth noting that our \textit{coreference} annotations differ from the widely adopted CoNLL-2012 \citep{pradhan2012conll} scheme in two aspects:
\begin{enumerate*}[(i)]
    \item we retain singleton entities composed by only one mention as a valid entity cluster, 
    \item we only cluster proper nouns, leaving out nominal and anaphoric expressions. 
\end{enumerate*}
{\renewcommand\baselinestretch{1}
\begin{table}[!t]
\centering
\caption[test]{Qualitative comparison of the datasets. We divide our comparison in five groups: \protect\begin{enumerate*}[(i)] 
	 \item \textit{Core Tasks} represent the main subtasks covered in \datasetname, 
	 \item \textit{Doc-Based} indicates whether different subtasks are annotated on the document-level,
	 \item \textit{Entity-Centric} indicates which annotations are done with respect to entity clusters (\cmark) as opposed to individual mentions (\xmark), 
	 \item \textit{Unaided} specifies whether the annotation process was completely manual (\cmark) or with some form of distant supervision (\xmark), and 
	 \item \textit{Open} indicates whether the dataset is freely available. 
	 \protect\end{enumerate*}} 
\resizebox{1.0\textwidth}{!}{
\begin{tabular}{l cccc cccc cccc cccc c}
\toprule
& \multicolumn{4}{c}{\textbf{Core Tasks}} & \multicolumn{4}{c}{\textbf{Doc-Based}} & \multicolumn{4}{c}{\textbf{Entity-Centric}} & \multicolumn{4}{c}{\textbf{Unaided}} & \\
\cmidrule(lr){2-5} \cmidrule(lr){6-9}\cmidrule(lr){10-13}\cmidrule(lr){14-17}
\multicolumn{1}{c}{\textbf{Dataset}} & \rotatebox{90}{\textbf{NER}} & \rotatebox{90}{\textbf{Coreference}} & \rotatebox{90}{\textbf{Relations}} & \rotatebox{90}{\textbf{Linking}} & \rotatebox{90}{\textbf{Coreference}} & \rotatebox{90}{\textbf{Relations}} & \rotatebox{90}{\textbf{Multi-label Rel}} & \rotatebox{90}{\textbf{Keywords}} & \rotatebox{90}{\textbf{Classification}} & \rotatebox{90}{\textbf{Multi-label Ent}} & \rotatebox{90}{\textbf{Relations}} & \rotatebox{90}{\textbf{Linking}} & \rotatebox{90}{\textbf{NER}} & \rotatebox{90}{\textbf{Coreference}} & \rotatebox{90}{\textbf{Relations}} & \rotatebox{90}{\textbf{Linking}} & \rotatebox{90}{\textbf{Open}}\\
\midrule
\textbf{\datasetname} & \cellcolor{presentattr}\cmark & \cellcolor{presentattr}\cmark & \cellcolor{presentattr}\cmark & \cellcolor{presentattr}\cmark & \cellcolor{presentattr}\cmark & \cellcolor{presentattr}\cmark & \cellcolor{presentattr}\cmark & \cellcolor{presentattr}\cmark & \cellcolor{presentattr}\cmark & \cellcolor{presentattr}\cmark & \cellcolor{presentattr}\cmark & \cellcolor{presentattr}\cmark & \cellcolor{presentattr}\cmark & \cellcolor{presentattr}\cmark & \cellcolor{presentattr}\cmark & \cellcolor{presentattr}\cmark & \cellcolor{presentattr}\cmark\\
TAC-KBP~\citep{ji2010overview,ji2015overview,ji2017overview} & \cellcolor{presentattr}\cmark & \cellcolor{presentattr}\cmark & \cellcolor{presentattr}\cmark & \cellcolor{presentattr}\cmark & \cellcolor{presentattr}\cmark & \cellcolor{absentattr}\xmark & \cellcolor{absentattr}\xmark & \cellcolor{absentattr}\xmark & \cellcolor{presentattr}\cmark & \cellcolor{absentattr}\xmark & \cellcolor{absentattr}\xmark & \cellcolor{presentattr}\cmark & \cellcolor{presentattr}\cmark & \cellcolor{presentattr}\cmark & \cellcolor{presentattr}\cmark & \cellcolor{presentattr}\cmark & \cellcolor{absentattr}\xmark\\
BC5CDR~\citep{li2016biocreative,wei2015overview} & \cellcolor{presentattr}\cmark & \cellcolor{presentattr}\cmark & \cellcolor{presentattr}\cmark & \cellcolor{presentattr}\cmark & \cellcolor{presentattr}\cmark & \cellcolor{presentattr}\cmark & \cellcolor{absentattr}\xmark & \cellcolor{absentattr}\xmark & \cellcolor{absentattr}\xmark & \cellcolor{absentattr}\xmark & \cellcolor{presentattr}\cmark & \cellcolor{absentattr}\xmark & \cellcolor{presentattr}\cmark & \cellcolor{presentattr}\cmark & \cellcolor{absentattr}\xmark & \cellcolor{presentattr}\cmark & \cellcolor{presentattr}\cmark\\
MUC-7~\citep{chinchor1998muc} & \cellcolor{presentattr}\cmark & \cellcolor{presentattr}\cmark & \cellcolor{presentattr}\cmark & \cellcolor{absentattr}\xmark & \cellcolor{presentattr}\cmark & \cellcolor{presentattr}\cmark & \cellcolor{absentattr}\xmark & \cellcolor{absentattr}\xmark & \cellcolor{presentattr}\cmark & \cellcolor{absentattr}\xmark & \cellcolor{presentattr}\cmark & \cellcolor{absentattr}\xmark & \cellcolor{presentattr}\cmark & \cellcolor{presentattr}\cmark & \cellcolor{presentattr}\cmark & \cellcolor{absentattr}\xmark & \cellcolor{absentattr}\xmark\\
SciERC~\citep{luan2018multi} & \cellcolor{presentattr}\cmark & \cellcolor{presentattr}\cmark & \cellcolor{presentattr}\cmark & \cellcolor{absentattr}\xmark & \cellcolor{presentattr}\cmark & \cellcolor{absentattr}\xmark & \cellcolor{absentattr}\xmark & \cellcolor{absentattr}\xmark & \cellcolor{presentattr}\cmark & \cellcolor{absentattr}\xmark & \cellcolor{absentattr}\xmark & \cellcolor{absentattr}\xmark & \cellcolor{presentattr}\cmark & \cellcolor{presentattr}\cmark & \cellcolor{presentattr}\cmark & \cellcolor{presentattr}\cmark & \cellcolor{presentattr}\cmark\\
DocRED~\citep{yao2019docred} & \cellcolor{presentattr}\cmark & \cellcolor{presentattr}\cmark & \cellcolor{presentattr}\cmark & \cellcolor{absentattr}\xmark & \cellcolor{presentattr}\cmark & \cellcolor{presentattr}\cmark & \cellcolor{presentattr}\cmark & \cellcolor{absentattr}\xmark & \cellcolor{presentattr}\cmark & \cellcolor{absentattr}\xmark & \cellcolor{presentattr}\cmark & \cellcolor{absentattr}\xmark & \cellcolor{absentattr}\xmark & \cellcolor{absentattr}\xmark & \cellcolor{absentattr}\xmark & \cellcolor{absentattr}\xmark & \cellcolor{presentattr}\cmark\\
Rich ERE~\citep{song2015light,aguilar2014comparison} & \cellcolor{presentattr}\cmark & \cellcolor{presentattr}\cmark & \cellcolor{presentattr}\cmark & \cellcolor{absentattr}\xmark & \cellcolor{presentattr}\cmark & \cellcolor{absentattr}\xmark & \cellcolor{absentattr}\xmark & \cellcolor{absentattr}\xmark & \cellcolor{presentattr}\cmark & \cellcolor{absentattr}\xmark & \cellcolor{absentattr}\xmark & \cellcolor{absentattr}\xmark & \cellcolor{presentattr}\cmark & \cellcolor{presentattr}\cmark & \cellcolor{presentattr}\cmark & \cellcolor{absentattr}\xmark & \cellcolor{absentattr}\xmark\\
ACE 2005~\citep{walker2006ace} & \cellcolor{presentattr}\cmark & \cellcolor{presentattr}\cmark & \cellcolor{presentattr}\cmark & \cellcolor{absentattr}\xmark & \cellcolor{absentattr}\xmark & \cellcolor{absentattr}\xmark & \cellcolor{absentattr}\xmark & \cellcolor{absentattr}\xmark & \cellcolor{presentattr}\cmark & \cellcolor{absentattr}\xmark & \cellcolor{absentattr}\xmark & \cellcolor{absentattr}\xmark & \cellcolor{presentattr}\cmark & \cellcolor{presentattr}\cmark & \cellcolor{presentattr}\cmark & \cellcolor{absentattr}\xmark & \cellcolor{absentattr}\xmark\\
OntoNotes 5.0~\citep{hovy2006ontonotes,weischedel2013ontonotes} & \cellcolor{presentattr}\cmark & \cellcolor{presentattr}\cmark & \cellcolor{absentattr}\xmark & \cellcolor{absentattr}\xmark & \cellcolor{absentattr}\xmark & \cellcolor{absentattr}\xmark & \cellcolor{absentattr}\xmark & \cellcolor{absentattr}\xmark & \cellcolor{absentattr}\xmark & \cellcolor{absentattr}\xmark & \cellcolor{absentattr}\xmark & \cellcolor{absentattr}\xmark & \cellcolor{presentattr}\cmark & \cellcolor{presentattr}\cmark & \cellcolor{presentattr}\cmark & \cellcolor{presentattr}\cmark & \cellcolor{presentattr}\cmark\\
ScienceIE~\citep{augenstein2017semeval} & \cellcolor{presentattr}\cmark & \cellcolor{absentattr}\xmark & \cellcolor{presentattr}\cmark & \cellcolor{absentattr}\xmark & \cellcolor{absentattr}\xmark & \cellcolor{presentattr}\cmark & \cellcolor{absentattr}\xmark & \cellcolor{presentattr}\cmark & \cellcolor{absentattr}\xmark & \cellcolor{absentattr}\xmark & \cellcolor{absentattr}\xmark & \cellcolor{absentattr}\xmark & \cellcolor{presentattr}\cmark & \cellcolor{absentattr}\xmark & \cellcolor{presentattr}\cmark & \cellcolor{absentattr}\xmark & \cellcolor{presentattr}\cmark\\
FewRel~\citep{han2018fewrel} & \cellcolor{presentattr}\cmark & \cellcolor{presentattr}\cmark & \cellcolor{presentattr}\cmark & \cellcolor{presentattr}\cmark & \cellcolor{absentattr}\xmark & \cellcolor{absentattr}\xmark & \cellcolor{absentattr}\xmark & \cellcolor{absentattr}\xmark & \cellcolor{absentattr}\xmark & \cellcolor{absentattr}\xmark & \cellcolor{presentattr}\cmark & \cellcolor{presentattr}\cmark & \cellcolor{absentattr}\xmark & \cellcolor{absentattr}\xmark & \cellcolor{absentattr}\xmark & \cellcolor{absentattr}\xmark & \cellcolor{presentattr}\cmark\\
GENIA~\citep{kim2003genia} & \cellcolor{presentattr}\cmark & \cellcolor{presentattr}\cmark & \cellcolor{presentattr}\cmark & \cellcolor{absentattr}\xmark & \cellcolor{presentattr}\cmark & \cellcolor{absentattr}\xmark & \cellcolor{absentattr}\xmark & \cellcolor{absentattr}\xmark & \cellcolor{absentattr}\xmark & \cellcolor{absentattr}\xmark & \cellcolor{absentattr}\xmark & \cellcolor{absentattr}\xmark & \cellcolor{presentattr}\cmark & \cellcolor{absentattr}\xmark & \cellcolor{absentattr}\xmark & \cellcolor{absentattr}\xmark & \cellcolor{presentattr}\cmark\\
AIDA CoNLL-YAGO~\citep{hoffart2011robust} & \cellcolor{presentattr}\cmark & \cellcolor{absentattr}\xmark & \cellcolor{absentattr}\xmark & \cellcolor{presentattr}\cmark & \cellcolor{absentattr}\xmark & \cellcolor{absentattr}\xmark & \cellcolor{absentattr}\xmark & \cellcolor{absentattr}\xmark & \cellcolor{absentattr}\xmark & \cellcolor{absentattr}\xmark & \cellcolor{absentattr}\xmark & \cellcolor{absentattr}\xmark & \cellcolor{presentattr}\cmark & \cellcolor{absentattr}\xmark & \cellcolor{absentattr}\xmark & \cellcolor{presentattr}\cmark & \cellcolor{presentattr}\cmark\\
SemEval 2010 T8~\citep{hendrickx2010semeval} & \cellcolor{presentattr}\cmark & \cellcolor{absentattr}\xmark & \cellcolor{presentattr}\cmark & \cellcolor{absentattr}\xmark & \cellcolor{absentattr}\xmark & \cellcolor{absentattr}\xmark & \cellcolor{absentattr}\xmark & \cellcolor{absentattr}\xmark & \cellcolor{absentattr}\xmark & \cellcolor{absentattr}\xmark & \cellcolor{absentattr}\xmark & \cellcolor{absentattr}\xmark & \cellcolor{presentattr}\cmark & \cellcolor{absentattr}\xmark & \cellcolor{presentattr}\cmark & \cellcolor{absentattr}\xmark & \cellcolor{presentattr}\cmark\\
NYT~\citep{riedel2010modeling} & \cellcolor{presentattr}\cmark & \cellcolor{absentattr}\xmark & \cellcolor{presentattr}\cmark & \cellcolor{presentattr}\cmark & \cellcolor{absentattr}\xmark & \cellcolor{absentattr}\xmark & \cellcolor{absentattr}\xmark & \cellcolor{absentattr}\xmark & \cellcolor{absentattr}\xmark & \cellcolor{absentattr}\xmark & \cellcolor{absentattr}\xmark & \cellcolor{absentattr}\xmark & \cellcolor{absentattr}\xmark & \cellcolor{absentattr}\xmark & \cellcolor{absentattr}\xmark & \cellcolor{absentattr}\xmark & \cellcolor{presentattr}\cmark\\
ACEtoWiki~\citep{bentivogli2010extending} & \cellcolor{absentattr}\xmark & \cellcolor{absentattr}\xmark & \cellcolor{absentattr}\xmark & \cellcolor{presentattr}\cmark & \cellcolor{absentattr}\xmark & \cellcolor{absentattr}\xmark & \cellcolor{absentattr}\xmark & \cellcolor{absentattr}\xmark & \cellcolor{absentattr}\xmark & \cellcolor{absentattr}\xmark & \cellcolor{absentattr}\xmark & \cellcolor{absentattr}\xmark & \cellcolor{absentattr}\xmark & \cellcolor{absentattr}\xmark & \cellcolor{absentattr}\xmark & \cellcolor{presentattr}\cmark & \cellcolor{presentattr}\cmark\\
WNUT 2017~\citep{derczynski2017results} & \cellcolor{presentattr}\cmark & \cellcolor{absentattr}\xmark & \cellcolor{absentattr}\xmark & \cellcolor{absentattr}\xmark & \cellcolor{absentattr}\xmark & \cellcolor{absentattr}\xmark & \cellcolor{absentattr}\xmark & \cellcolor{absentattr}\xmark & \cellcolor{absentattr}\xmark & \cellcolor{absentattr}\xmark & \cellcolor{absentattr}\xmark & \cellcolor{absentattr}\xmark & \cellcolor{presentattr}\cmark & \cellcolor{absentattr}\xmark & \cellcolor{absentattr}\xmark & \cellcolor{absentattr}\xmark & \cellcolor{presentattr}\cmark\\
CoNLL-2003~\citep{sang2003introduction} & \cellcolor{presentattr}\cmark & \cellcolor{absentattr}\xmark & \cellcolor{absentattr}\xmark & \cellcolor{absentattr}\xmark & \cellcolor{absentattr}\xmark & \cellcolor{absentattr}\xmark & \cellcolor{absentattr}\xmark & \cellcolor{absentattr}\xmark & \cellcolor{absentattr}\xmark & \cellcolor{absentattr}\xmark & \cellcolor{absentattr}\xmark & \cellcolor{absentattr}\xmark & \cellcolor{presentattr}\cmark & \cellcolor{absentattr}\xmark & \cellcolor{absentattr}\xmark & \cellcolor{absentattr}\xmark & \cellcolor{presentattr}\cmark\\
TACRED~\citep{zhang2017position} & \cellcolor{absentattr}\xmark & \cellcolor{absentattr}\xmark & \cellcolor{presentattr}\cmark & \cellcolor{absentattr}\xmark & \cellcolor{absentattr}\xmark & \cellcolor{absentattr}\xmark & \cellcolor{absentattr}\xmark & \cellcolor{absentattr}\xmark & \cellcolor{absentattr}\xmark & \cellcolor{absentattr}\xmark & \cellcolor{absentattr}\xmark & \cellcolor{absentattr}\xmark & \cellcolor{absentattr}\xmark & \cellcolor{absentattr}\xmark & \cellcolor{presentattr}\cmark & \cellcolor{absentattr}\xmark & \cellcolor{absentattr}\xmark\\
\bottomrule
\end{tabular}
}
\label{tab:quali_compare}\end{table}}

Furthermore, most prominent efforts to produce jointly annotated datasets have focused on using a \emph{top-down} annotation approach. This method involves an a priori defined annotation schema that drives the process of selection and labeling of the corpus. 
The de facto datasets used in most of the joint learning baselines such as ACE 2005 
\citep{doddington2004automatic, walker2006ace}, 
TAC-KBPs \citep{ji2010overview,ellis2014overview,ji2015overview,ellis2015overview,ji2017overview} and Rich ERE \citep{song2015light} 
use this annotation approach. More specifically, during the creation of the ACE 2005 dataset, 
the annotators initially tagged candidate documents as ``good'' or ``bad'' depending on the estimated number and types of entities present in each one. In subsequent annotation stages, only ``good'' documents were fully annotated and included in the final dataset. Similarly, during the creation of the TAC-KBP 
datasets, the annotators focused on producing 
annotations evenly distributed among three entity types (PERs, ORGs, and GPEs) by annotating only the documents that contained a minimum number of entities related to event types. In the case of Rich ERE, 
the documents to tag were prioritized by the event trigger word density calculated per 1,000 tokens, thus focusing only on content with a high number of previously defined key event-related tokens. Furthermore, other IE-related datasets \citep{augenstein2017semeval,han2018fewrel,yao2019docred,zhang2017position,hendrickx2010semeval} use similar pre-filtering techniques in order to select the text to be annotated.
As a consequence, the corpus and annotations in these datasets tend to be biased and likely not representative of the language used in the different input domains. Conversely, we adopt a radically different \textit{bottom-up} approach where we derive the annotations (\eg entity classification types, relation types) from the data itself. 
This bottom-up data-driven procedure guarantees that the annotations in {\datasetname} are representative of the document corpus information and reflects the particularities of the language used in its journalistic domain. Furthermore, it better represents the properties that are inherently present in written corpora, \eg the long-tail distribution of different annotation types.
\setlength{\tabcolsep}{5pt}
\renewcommand{\arraystretch}{1.0}
{\renewcommand\baselinestretch{1}\begin{table}[t]
\centering
\caption{Numerical comparison of \datasetname~and well-known IE datasets. Note that some datasets (including \datasetname) use an entity-centric approach, organizing entity mentions in entity clusters, and annotating entities, relations, and linking on the cluster level. Hence, we provide both mention-level as well as cluster-level (if a particular dataset supports it) statistics.}
\resizebox{1.0\textwidth}{!}{
\begin{tabular}{llllllllll} 
\toprule
& & \multicolumn{3}{c}{\textbf{Entities}} & \multicolumn{3}{c}{\textbf{Relations}} & \multicolumn{2}{c}{\textbf{Linking}} \\ 
\cmidrule(lr){3-5} \cmidrule(lr){6-8}\cmidrule(lr){9-10} 
Dataset & \# Tokens & \# Mentions & \# Entity & \# Entity & \# Relation & \# Relation & \# Relation & \# Mention & \# Cluster\\
 &  &  & clusters & types & mentions & clusters & types & KB Links & KB Links\\
\midrule
NYT & 5,765,332 & 1,388,982 & - & - & 142,823 & - & 52 & 1,388,982 & -\\
TACRED & 3,866,863 & - & - & - & 21,784 & - & 42 & - & -\\
TAC-KBP\tablefootnote{The EDL track only of TAC-KBP 2010.} & 3,053,336 & 6,495 & 3,750 & - & - & - & - & 3,818 & 2,094\\
OntoNotes 5.0 & 2,088,832 & 161,783 & 136,037 & - & - & - & - & - & -\\
FewRel\tablefootnote{Numbers based on publicly available train and development sets.} & 1,397,333 & 114,213 & 112,000 & - & 58,267 & 56,000 & 80 & 114,213 & 112,000\\
DocRED & 1,018,297 & 132,392 & 98,610 & 6 & 155,535 & 50,503 & 96 & -  & -\\
MUC-4 & 717,798 & 14,196 & - & 13 & - & - & - & - & -\\
GENIA & 554,346 & 56,743 & 10,728 & 5 & 2,337 & - & 2 & - & -\\
\cellcolor{ourdataset}{\datasetname} & \cellcolor{ourdataset}501,095 & \cellcolor{ourdataset}43,373 & \cellcolor{ourdataset}23,130 & \cellcolor{ourdataset}311 & \cellcolor{ourdataset}317,204 & \cellcolor{ourdataset}21,749 & \cellcolor{ourdataset}65 & \cellcolor{ourdataset}28,482 & \cellcolor{ourdataset}13,086\\
BC5CDR & 343,175 & 29,271 & 10,326 & 2 & 47,813 & 3,116 & 1 & 29,562 & 10,326\\
CoNLL-2003 & 301,418 & 35,089 & - & 4 & - & - & - & - & -\\
CoNLL-YAGO & 301,418 & 34,929 & - & - & - & - & - & 34,929 & -\\
ACE 2005 & 259,889 & 54,824 & 37,622 & 51 & 8,419 & 7,786 & 18 & - & -\\
ACEtoWiki & 259,889 & - & - & - & - & - & - & 16,310 & -\\
SEval 2010 T8 & 207,307 & 21,434 & - & - & 6,674 & - & 9 & - & -\\
ACE 2004 & 185,696 & 29,949 & 12,507 & 43 & 5,976 & 5,525 & 24 & - & -\\
WNUT 2017 & 101,857 & 3,890 & - & 6 & - & - & - & - & -\\
ScienceIE & 99,580 & 9,946 & 9,536 & 3 & 638 & - & 1 & - & -\\
SciERC & 65,334 & 8,094 & 1,015 & 6 & 2,687 & - & 7 & - & -\\
\bottomrule
\end{tabular}
}
\label{tab:quanti_compare}
\end{table}}
\setlength{\tabcolsep}{5pt}

Finally, from the perspective of the necessary evidence to annotate a particular entity type or relation, we propose to make a distinction for the currently existing datasets between \textit{trigger-based} 
and 
\textit{document-based} 
annotations (see \textit{Doc-Based} comparison group in \Tabref{tab:quali_compare}).  
The \textit{trigger-based} datasets require that a particular relation or entity 
type should only be annotated if it is supported by an explicit reference in a text. For example, in \figref{fig:example} there is a concrete reference of the relation between ``Meghan" and ``Harry" in form of triggers such as ``gets engaged" in sentence~1 and ``The wedding" in sentence~2. Most of the traditionally used jointly annotated datasets such as ACE 2005 \citep{doddington2004automatic,walker2006ace}, TAC-KBPs \citep{ji2010overview,ellis2014overview,ji2015overview,ellis2015overview,ji2017overview} and Rich ERE \citep{song2015light}, as well as others, including FewRel \citep{han2018fewrel}, OntoNotes \citep{hovy2006ontonotes,weischedel2013ontonotes}, TACRED \citep{zhang2017position}, SemEval 2010 Task 8 \citep{hendrickx2010semeval} and SciERC \citep{luan2018multi}, are \emph{trigger-based}. The disadvantage of such an approach is that it only captures the most simple cases of relations and 
entity
types that are explicitly mentioned in the text. As a general rule, this also limits 
the datasets to cover only the relations between entity mentions (\ie the annotation process is mention-driven) that appear within a single or at most few adjacent sentences where the relation trigger occurs (see \figref{fig:rel_span} in \secref{sec:annotation} for a more detailed illustration of this phenomenon).
However, as we move to a broader \textit{document-based} interpretation, it is common to find relations that are not explicitly mentioned in text. Thus, in our example of \figref{fig:example} the relation between ``Ministry of Defense'' and ``Britain'' is not explicitly indicated in the text. However, after reading the whole article we can infer relations such as \textsf{ministry\_of}, \textsf{agency\_of} and \textsf{based\_in} between these two entities. This document-level reasoning makes it essential to adopt an entity-centric approach (see \textit{Entity-Centric} comparison group in \Tabref{tab:quali_compare}) where each \textit{entity}
comprises one or more entity \textit{mentions}, and the annotations (\ie \textit{relations}, \textit{entity tags} and \textit{entity linking} in \datasetname) are made on the entity level, thus abstracting from specific 
\textit{mention-driven} triggers. 

\subsection{Recent advances in Information Extraction}
\label{sec:recent_advances}

In the last couple of years, 
the advances in joint modeling have been accompanied by an ever increasing interest in the use of graph-based neural networks \citep{li2015gated,xu2018powerful,wu2020comprehensive}. Initially, this approach has been applied to improve the performance of the single coreference resolution task by transferring document-level contextual information between coreferenced entity mention spans \citep{lee2018higher, kantor2019coreference}. Most recently, these graph propagation techniques have been successfully used in a joint setting \citep{luan2019general, wadden2019entity, fei2020boundaries, fu2019graphrel} 
by performing graph message passing updates between the shared spans across different tasks. However, while successful on mention-driven datasets such as ACE 2005 \citep{walker2006ace} and 
NYT \citep{riedel2010modeling}, 
as far as we are aware, the advantages of these techniques have not yet been investigated in an entity-centric document-level setting.  
We fill this gap by extending the neural graph-based model initially proposed by \cite{luan2019general} to be used on \datasetname~(see \secref{sec:models}). More specifically, we explore the effect of performing document-level coreference (\propformat{CorefProp})~\citep{lee2018higher,luan2019general} and relation-driven (\propformat{RelProp})~\citep{luan2019general} graph message passing updates between the spans. Additionally, we introduce a new latent attention-based graph propagation method (\propformat{AttProp}) and compare it to 
previously proposed task-driven graph propagation methods (\propformat{CorefProp} and \propformat{RelProp}).  

\subsection{Metrics and evaluation}
\label{sec:metrics_and_evaluation}

Current dominant IE systems consider \emph{mention-level} scoring of NER as well as 
RE
components when reporting on datasets such as CoNLL-2003 \citep{lample2016neural,chiu2016named,baevski2019cloze,akbik2019pooled,akbik2018contextual}, OntoNotes \citep{clark2018semi,chiu2016named, strubell2017fast}, ACE 2004 \citep{bekoulis2018adversarial,li2014incremental,zhang2017end}, ACE 2005 \citep{zhang2017end,luan2019general,fei2020boundaries}, TACRED \citep{zhang2017position,soares2019matching,zhang2018graph}, and SelEval 2010-Task 8 \citep{hu2020cross,peters2019knowledge, guo2019attention} among others.
In contrast, the \datasetname~dataset is entity-centric where all the annotations are done on the entity cluster level. Consequently, adopting a purely mention-based evaluation approach can lead to a dominance of the score by predictions on entities composed by many mentions as opposed to entities composed by only few ones. 
Conversely, a purely cluster-level evaluation would be overly strict, requiring correct prediction of relation/entity types as well as an exact match of the predicted entity clusters.
To tackle this problem, we propose a new scoring method that combines entity mention-level and cluster-level evaluation, while avoiding the pitfalls of either method alone (see \secref{sec:metrics}).

\section{Annotation process}
\label{sec:annotation}

\renewcommand\baselinestretch{1}\small
\begin{table}
    \centering
    \caption{\textit{Descriptions} and \textit{Examples} (with entity mentions underlined) of each of the most granular entity classes in \datasetname~(\textit{ENTITY}, \textit{VALUE} and \textit{OTHER}) in the \textit{type} tag hierarchy. Additionally, for the type \textit{ENTITY}, we describe and give examples of each of its direct subtypes (\textit{location}, \textit{organization}, \textit{person}, \textit{misc}, \textit{event} and \textit{ethnicity}).}
    \label{tab:top_entities_detail}

    \small
    \renewcommand{\arraystretch}{1.5}
    \begin{tabular}{lp{6.0cm}p{6.0cm}}
    \toprule
    \textbf{Entity Tag} &
    \multicolumn{1}{c}{\textbf{Description}} &
    \multicolumn{1}{c}{\textbf{Example}}
    \\
    \midrule
    ENTITY & 
    All nominal named entities. &
    ``\underline{UK} court rules \underline{WikiLeaks}' \underline{Assange} should be extradited to \underline{Sweden}''
    \\    
    \ \ \ \ location & 
    Entities referring to a particular geographical location. &
    ``\underline{Libya} is one of \underline{Germany}'s strongest trading partners in northern \underline{Africa}.''
    \\
    \ \ \ \ organization &
    Organizations such as companies, governmental organizations, etc. & 
    ``According to the report, \underline{Amazon} would pay the same level of royalty fees as \underline{Apple}.''
    \\ 
    \ \ \ \ person &
    Entities referring to people in general such as politicians, artists, sport players, etc. & 
    ``With \underline{Ramires} out, \underline{Drogba} could start as striker, with \underline{Torres} moving to the wing.''
    \\
    \ \ \ \ misc &
    Miscellaneous entity types such as names of work of arts, treaties, product names, etc.  & 
    ``According to the director's own words, \underline{The Post} is a `patriotic film'.''
    \\
    \ \ \ \ event &
    Events such as sport competitions, summits, etc. & 
    ``Last year's \underline{Champions League} final drew a crowd of just 14,303.''
    \\
    \ \ \ \ ethnicity &
    Entity type used to identify different ethnic groups. & 
    ``Attempt to assimilate \underline{Uyghurs} into dominant \underline{Han Chinese} culture.''
    \\
    VALUE &
    Values in general such as time, money, etc.  &
    ``It ended the \underline{2014} fiscal year \underline{45 million} \underline{euros} (\underline{\$51 million}) in the red.''
    \\
    OTHER & Includes the nominal variations of entity types (\eg 
    includes variations of country names such as ``German'', which is a variation of ``Germany''). 
     & ``\underline{Franco}-\underline{German} `war child' granted \underline{German} citizenship.'' \\ 
    \bottomrule
    \end{tabular}
\end{table}
\renewcommand\baselinestretch{1.0}\normalsize
In this work we introduce our \textit{bottom-up} data-driven annotation approach.
Our main goal is to get an annotation schema that reflects
the types of entities and relations that are effectively mentioned throughout the corpus
to maximally capture the information it contains. 
Therefore, we derive the annotation schema from the corpus itself, adopting three annotation passes that are detailed next:
\begin{enumerate*}[(i)]
  \item \textit{exploratory pass}, 
  \item \textit{schema-driven pass}, and 
  \item \textit{inter-annotator refinement}.
\end{enumerate*}
Each pass encompasses substeps to cover all IE subtasks:
\begin{enumerate*}[(i)]
\item mention annotation (\ie the entities and their types),
\item coreference resolution,
\item relation extraction on the entity level (\ie clustering all mentions referring to the same entity), and
\item entity linking (again, on the entity level, providing the same link for all clustered mentions).
\end{enumerate*}

\subsection{Exploratory pass} 
The first annotation pass aims to discover the annotation structure (\ie annotation schema) to be used on the corpus, in particular the types to use for named entity recognition (NER) and relation extraction (RE) tasks. Three annotators are involved in this step to provide annotations on the mention level: one expert annotator and two paid students. However, no parallel annotation is done and the role of the expert annotator is to annotate part of the corpus, as well as instructing and supervising the paid annotators. No a priori fixed schema is followed, but we ask the annotators to be as consistent as possible during the process. More specifically, the annotators are free to define their own entity and relation types for the NER and RE tasks that reflect the contents of the articles as long as they comply with the following generic guidelines: 
\begin{enumerate}
    \item \textbf{Named Entities:} any physical or abstract object (\eg ``Washington'',``Jeff Davis'',``Nobel Prize'', ``Lisbon Treaty'', etc.) that can be denoted with a proper noun. Entities are usually upper-cased in the text, although values such as money and time can also be included. Use short and specific entity types (\eg person, organization, etc.) to classify entities, the types can be overlapping (a single entity can have multiple types). 
    \item \textbf{Relations:} identify meaningful relations between entities. The type of a relation should be specific and reflect the type of the connected entities as well as the semantic meaning of the relation. For example, instead of using a generic ``located in'' relation for entities located in a particular country, we can divide it in ``based in country'' for organizations that are based in a country, ``city located in country'' for cities located in the country, etc. The types of the relations should have short names, ideally not exceeding 15 characters. 
\end{enumerate}
By not constraining the annotation process to specific entity and relation types, we ensure that our annotations are representative of the actual information contained in the annotated corpus.

\begin{figure}[!t]
\centering
\includegraphics[width=1.0\columnwidth, trim={0cm 0cm 0cm 0cm},clip]{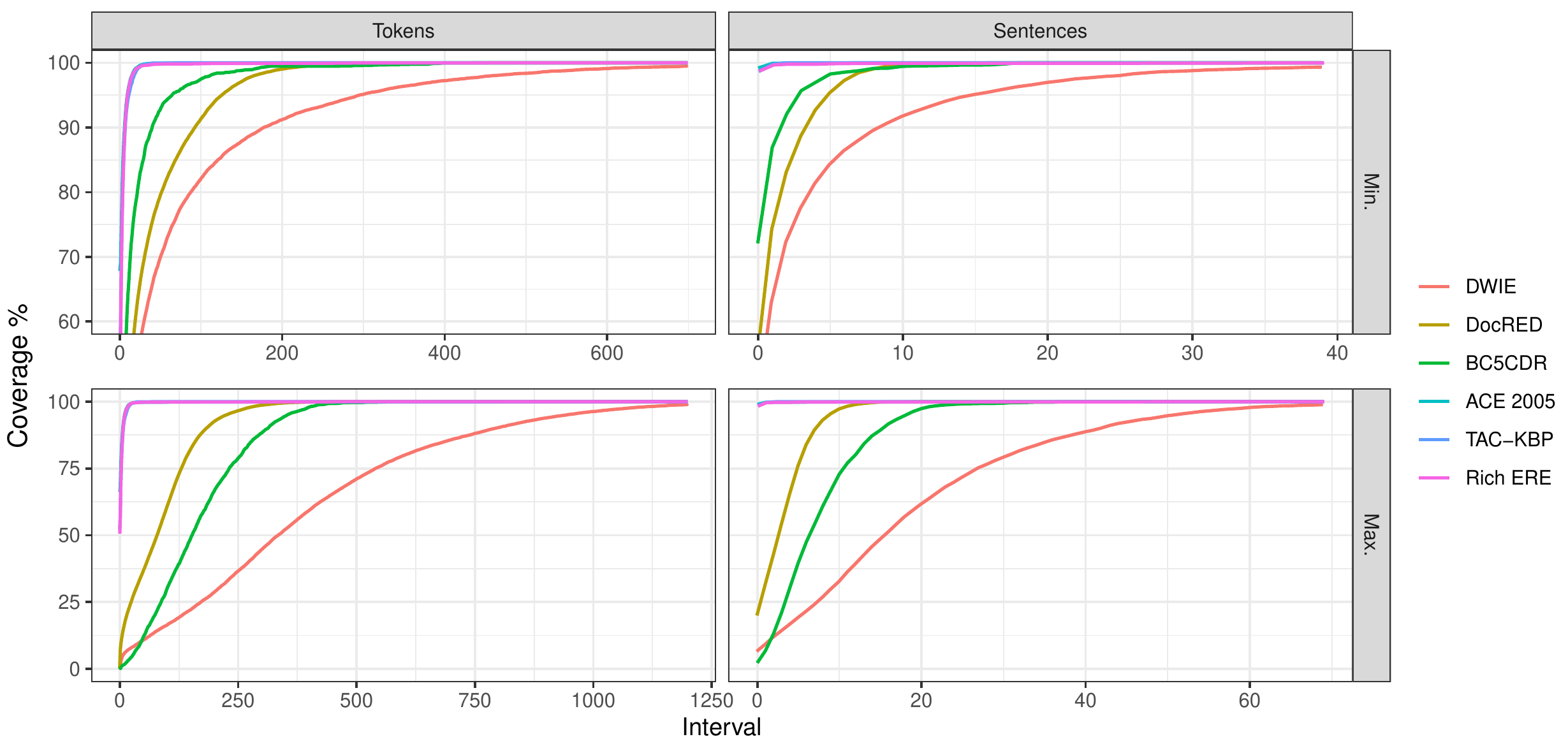}
\caption[test]{Comparison of the coverage of the \% of relations with increasing  
interval in tokens
(left) and 
interval in sentences
(right). The graph at the top illustrates the relations coverage measuring the minimum distance between entities (closest mentions). Conversely, the graph at the bottom shows the coverage measuring the maximum distance between entity mentions.
In both 
graphs, we note that the distance between the related mentions in our dataset is higher than in other widely used datasets. }
\label{fig:rel_span}
\end{figure}

\subsection{Schema-driven pass}
\label{sec:schema_driven_pass}

\setlength{\tabcolsep}{5pt}
The main goal of this step is to create a consistent annotation schema for 
\begin{enumerate*}[(i)]
\item named entity types and
\item relation types
\end{enumerate*} based on the annotations made in the \textit{exploratory pass}.
As a first step, we identify the 
classification \textit{tags} to be assigned to \textbf{entities}. 
We divide these tags in five main categories: \textit{type}, \textit{topic}, \textit{iptc},
\textit{slot}, and \textit{gender} (see \tabref{tab:main_entity_types}). 
Our \textit{type} tag is organized in a hierarchical structure (see \tabref{tab:entities_stats} in  \appref{app:dataset_insights}), making it easier to extend our annotations to more granular subtypes. \tabref{tab:top_entities_detail} defines and provides examples 
of
each of the top \textit{type} tags in the entity type hierarchy (\textit{ENTITY}, \textit{VALUE} and \textit{OTHER}) as well as the direct subtypes of \textit{ENTITY}. The \textit{topic} tag allows to assign topics (\eg politics, culture, education, etc.)\ to the entities and it complements the \textit{type} tag (see \tabref{tab:ner_labels_matrix}). The \textit{iptc} tag is used for the universally defined IPTC news categories based on a media taxonomy (\url{https://iptc.org/standards/subject-codes/}). The \textit{slot} tag is used for additional categorization that is transversal to different entity types. One example of this is the slot \textit{interviewee} that can be assigned to any person (entity of type \textit{person}) interviewed in a particular article.\footnote{Other possible slot values are: \textit{keyword}, \textit{head}, \textit{death}, \textit{interviewer} and \textit{expert}.} Finally, the \textit{gender} tag is used to indicate the gender of the entities that refer to people. 
By defining these multiple overlapping tag types, we 
realize 
that the entity classification is multi-label by nature and thus allows different complementary entity tags to be assigned to a particular entity.\footnote{The average number of labels per entity is 4.0 in our \datasetname~dataset.} 
This contrasts with prevailing single-label multi-class datasets such as ACE 2005~\citep{doddington2004automatic,walker2006ace}, TAC-KBPs~\citep{ellis2014overview,ji2015overview,ellis2015overview,ji2017overview}, Rich ERE~\citep{song2015light}, WNUT 2017~\citep{derczynski2017results} and CoNLL-2003~\citep{sang2003introduction}. 
\setlength{\tabcolsep}{5pt}
\renewcommand{\arraystretch}{1.0}
\renewcommand\baselinestretch{1}\small
\begin{table}
    \centering
    \caption{\textit{Descriptions} and \textit{Examples} of the top 5 most occurring relation types in \datasetname. The entity mentions involved in the relations are underlined.}
    \label{tab:top_relations_detail}

    \small
    \renewcommand{\arraystretch}{1.5}
    \begin{tabular}{lp{6.2cm}p{6.2cm}}
    \toprule
    \renewcommand{\arraystretch}{1.0}
    \begin{tabular}{c@{}c@{}}\textbf{Relation} \\ \textbf{Type}\end{tabular}
    &
    \multicolumn{1}{c}{\textbf{Description}} &
    \multicolumn{1}{c}{\textbf{Example}}
    \\
    \midrule
    based\_in0 & 
    Relations between organizations and the countries they are based in, ex: \relation{based\_in0}{University of Cologne}{Germany} &
    ``Now he's back in \underline{Germany} carrying on with his cancer research at the \underline{University of Cologne}.''
    \\
    in0 &
    Relations between geographic locations and the countries they are located in, ex: \relation{in0}{Athens}{Greece} & 
    ``The murder of a left-wing activist in \underline{Athens} has shaken up \underline{Greece} and inspired a backlash.''
    \\ 
    citizen\_of &
    Relations between people and the country they are citizens of, ex: \relation{citizen\_of}{Guerrero}{Peru} & 
    ``Even as a teenager, \underline{Guerrero} played for the national side in his native \underline{Peru}.''
    \\
    based\_in0-x &
    Relations between organizations and the nominal variations of the countries they are based in, ex: \relation{based\_in0-x}{SPD}{German} & 
    ``\underline{SPD} denies `green light' for new \underline{German} government, but keeps options open''
    \\
    citizen\_of-x &
    Relations between people and the nominal variations of the countries they are citizens of, ex: \relation{citizen\_of-x}{Assange}{Australian} & 
    ``\underline{Australian} national \underline{Assange} said the accusations were politically motivated.''
    \\
    \bottomrule
    \end{tabular}
\end{table}
\renewcommand\baselinestretch{1.0}\normalsize

For our \textbf{relation} annotations, we focus on annotating relations between entities themselves (\cf\textit{document-based entity-centric} approach). 
Our adopted approach allows us to think concept-wise and come up not only with relations that are explicitly stated, but also those that can be implicitly inferred from the text. As a result, our dataset includes relations whose connected mentions are located further apart in the document. 
This can be seen in \figref{fig:rel_span}, where we compare the \textit{minimum} 
(\textit{Min.}) 
 and \textit{maximum} 
 (\textit{Max.}) 
 distances between the mentions of the two entities connected by a relation for various mention-driven (Rich ERE\footnote{We use the Rich ERE dataset from the LDC2015E29 and LDC2015E68 catalogs.}, TAC-KBP\footnote{We use the TAK-KBP 2017 dataset from the LDC2017E54 and LDC2017E55 catalogs.}, and ACE 2005) and entity-centric (DocRED, BC5CDR, and the final version of our \datasetname~dataset) 
RE datasets. 
We note how other datasets that define the relation in terms of entities (BC5CDR and DocRED) require a higher number of token and sentence spans to cover all the relations in the respective dataset:
entity-centric relations very often involve mentions located in different sentences in the document that refer to those entities. This is not the case for 
mention-driven trigger-based
relations as in the TAC-KBP, Rich ERE and ACE 2005 datasets, where the annotation bias is towards finding explicitly mentioned relations, often involving entity mentions in a single sentence. 

Similarly as with entity tags, we organize our relation annotations using multi-label types (see 
\tabref{tab:relations_multilabel_stats}
for details). \tabref{tab:top_relations_detail} gives some examples from the \datasetname~corpus for the top 5 most occurring relation types (a detailed list can be consulted in \tabref{tab:relations_stats}). For reasons of space, the examples only involve relations between entities whose mentions occur in a single sentence; for an example involving document-level relations we refer to \figref{fig:example}.

Additionally, we define logical rules to automatically guarantee the consistency of the relations and their types. The following is an example, 
\begin{align}
\relationalign{based\_in2}{X}{Z} \land \relationalign{in0}{Z}{Y} \implies \relationalign{based\_in0}{X}{Y} \label{eq:complex_logic}
\end{align}
reflecting the knowledge that if an organization \textit{X} is based in a city \textit{Z} (relation \textsf{based\_in2}), and that this city \textit{Z} is located in the country \textit{Y} (relation \textsf{in0}), the fact that company \textit{X} is also located in that country (relation \textsf{based\_in0}) is valid as well.
The goal of this step is mainly consistency of the annotations, but it implies that an effective predictor would need to perform some form of reasoning to correctly predict all relations in the dataset. A complete list of logical rules is provided in \appref{app:rel_rules_list}. 

\subsection{Inter-Annotator Refinement} 
\label{sec:interannotator}
\renewcommand{\thefootnote}{\fnsymbol{footnote}}  
{\renewcommand\baselinestretch{1}\begin{table}[t]
    \centering
\caption[test]{The inter-annotation agreement Cohen's kappa scores for all the different annotation tasks \textit{before} and \textit{after} the dataset 
refinement
used to analyze and correct
the discrepancies between the parallel annotations.
}
\begin{minipage}{1.0\textwidth}
\renewcommand{\thefootnote}{\fnsymbol{footnote}}  
\renewcommand*\footnoterule{}
    \centering
    \begin{tabular}{lcc}
    \toprule
    {\textbf{Task}} & \textbf{Before Refinement} & \textbf{After Refinement} \\
    \midrule
    Named Entity & 0.8497 & 0.8703 \\
    \ \ \ Named Entity Detection & 0.9665 & 0.9673 \\ 
    \ \ \ Named Entity Classification & 0.8812 & 0.9026 \\ 
    Coreference & 0.9302 & 0.9324 \\ 
    Entity Linking & 0.9280 & 0.9320 \\ 
    Relation & 0.6594 & 0.8729 \\
    \ \ \ Relation Detection & 0.7686 & 0.8727 \\ 
    \ \ \ Relation Classification & 0.8118 & 0.9666 \\ 
   \bottomrule
    \end{tabular}
\end{minipage}
    \label{tab:kappas_pass}
\end{table}}
\renewcommand{\thefootnote}{\arabic{footnote}}
In order to assess and further improve the quality of our dataset we re-annotate a 100 randomly selected news articles (12.5\% of the articles used in the previous annotation rounds) 
from scratch. This work is done by a second independent expert annotator. The annotations in this pass are performed by following the already defined annotation schema based on the annotation process in the \textit{exploratory} and \textit{schema-driven} passes. We use this second annotated subset to measure the inter-annotator agreement and subsequently determine the parts of the dataset that still need to be improved. \tabref{tab:kappas_pass} compares the kappa scores before and after this refinement pass for each of the tasks (see \appref{app:kappa_agreement_details} for details on how the kappa score is calculated). We observe that, after the refinement, all of the kappa scores are above 0.85, which is considered a `strong' \citep{mchugh2012interrater} to `almost perfect' \citep{landis1977measurement} agreement.

Note that the revisions were seeded by and evaluated on the subset of 100 re-annotated articles. However, we argue that the 
inter-annotator refinement
improved the annotation consistency of the entire dataset, given that the reviewed entity and relation types are used in more than 99.4\% of all annotations in \datasetname. 
  
\section{Model Architecture}
\label{sec:models}

\begin{figure}[!t]
\centering
\includegraphics[width=0.8\columnwidth, trim={0cm 2.8cm 8.5cm 0.2cm},clip]{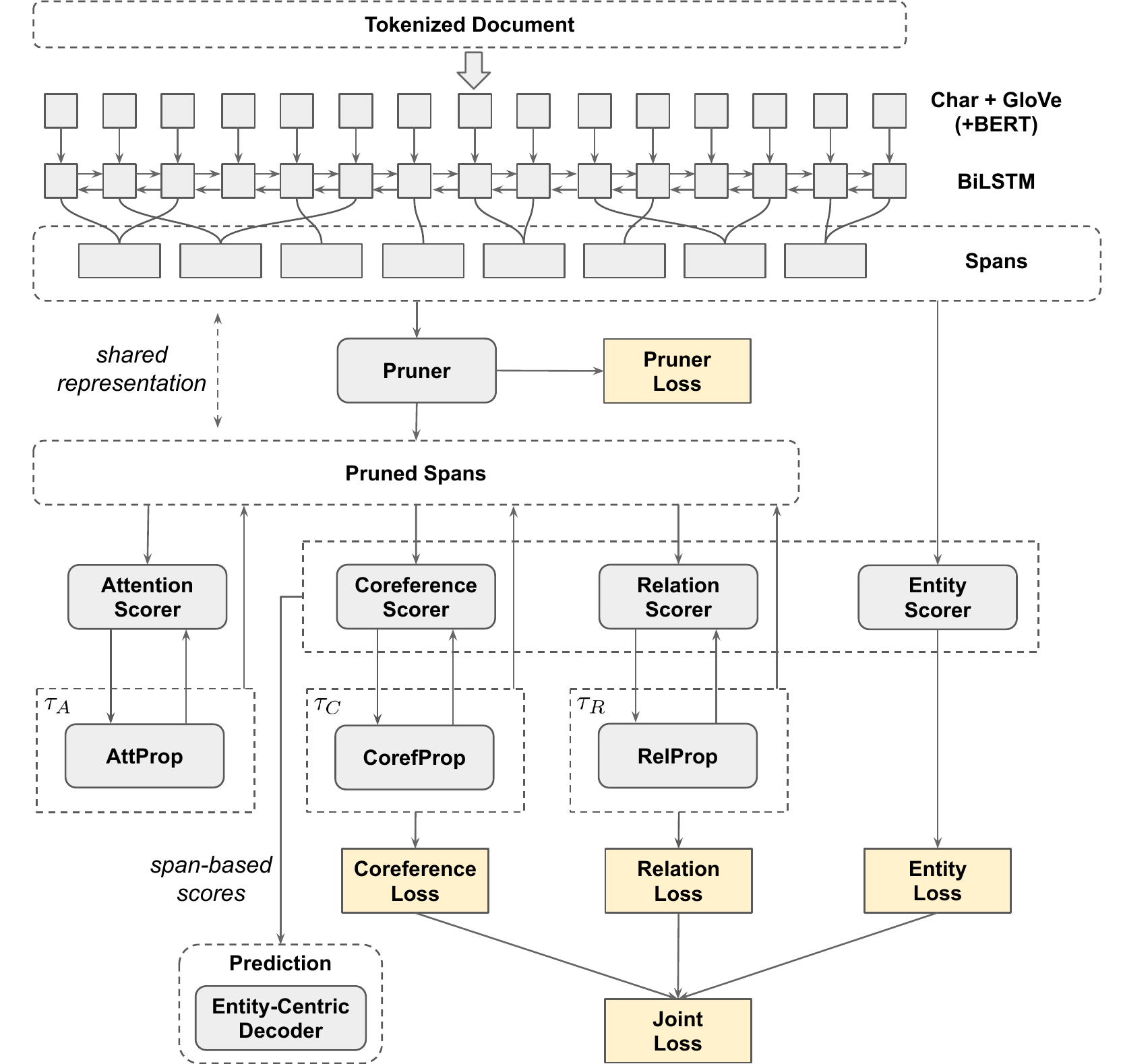}
\captionsetup{singlelinecheck=off}
\caption[test]{Architecture of our model; the span-oriented approach makes it possible to execute \textit{coreference} (\secref{subsubsec:coref}) and \textit{relation} (\secref{subsubsec:relation}) scorers independently from \textit{entity scorer}  (\secref{subsubsec:entity_module}). However, a pruning step (described in \secref{sec:shared_model}) is needed in order to 
limit
the memory required to perform matrix operations on span representations involved in graph propagation (\propformat{AttProp}, \propformat{CorefProp}, \propformat{RelProp})(\secref{sec:graph_prop_model}) as well as in the attention, coreference and relation scorer modules. The \textit{pruned spans} share the same representation with the rest of the \textit{spans} (\textit{shared representation}). This way, the update in span representations caused by the graph propagation modules also affects the \textit{entity scorer}. Our \propformat{AttProp} graph propagation method runs independently from coreference, relation, and entity scorers, enabling its use in combination with any task. Finally, the \textit{entity-centric decoder} (\secref{sec:decoding}) uses the entity clusters predicted by the \textit{coreference scorer} to convert the span-based predictions from the \textit{relation} and \textit{entity} scorers to entity-centric ones.}
\label{fig:model_architecture}
\end{figure}

In this section we introduce the end-to-end architecture used to compare the performance of models trained on the separate tasks  with the models that are trained jointly for multiple tasks  
on the \datasetname~dataset. 
The main component of our approach is the use of Graph Neural Networks~\citep{scarselli2008graph,li2015gated,xu2018powerful,wu2020comprehensive}, relying on propagation techniques in both single-task and joint setups. More specifically, we implement span-based 
graph message passing
on coreference (\propformat{CorefProp})~\citep{lee2018higher,luan2019general} and relation levels (\propformat{RelProp})~\citep{luan2019general}. Additionally, 
we introduce a latent attentive propagation method (\propformat{AttProp}) which 
is not driven by annotations of any task in particular
and, 
as a result, 
can be freely applied to any task or joint combination of tasks. The interconnection between the different components of our model architecture is depicted in \figref{fig:model_architecture}. 
It is based on the 
\textit{span-based architecture} introduced in \cite{lee2017end}, 
which supports training on the space of all entity spans simultaneously, dynamically updating span representations by using the graph propagation approach (further detailed in \secref{sec:graph_prop_model}). Recent works have shown that this idea has the potential for improved effectiveness (albeit at a higher computational cost)~\citep{lee2018higher,luan2019general,dixit2019span,fei2020boundaries}, compared to 
more traditional sequence-labeling approaches 
\citep{lample2016neural,ma2016end,luan2017scientific,katiyar2018nested}.
More concretely, the use of a span-based approach where all the spans are shared between the individual task modules avoids the cascading of errors from the entity mention identification module (\textit{entity scorer} in \figref{fig:model_architecture}) to the rest of the tasks. 

The most similar architecture to ours in using joint span-based neural graph IE is DyGIE~\citep{luan2019general} and its successor DyGIE$++$~\citep{wadden2019entity}. Our model is described in detail below, but here we already list the aspects in which it differs from these models:
\begin{enumerate}
    \item We introduce the graph propagation technique \propformat{AttProp} (see \secref{sec:graph_prop_model}), which is not directly conditioned on a particular task and can be used in  single-task (for each of the tasks) as well as joint settings. 
    \item We define a coreference architecture that, unlike previous work in span-based coreference resolution~\citep{lee2017end,lee2018higher}, allows to also account for singleton entities in the \datasetname~dataset (see Sections~\ref{subsubsec:coref} and \ref{subsubsec:single_coref}) by using an additional \textit{pruner loss}, which turns out essential for the single model focusing on end-to-end coreference resolution. 
    \item Due to the document-level nature of \datasetname, we run graph propagations on the whole document. This contrasts with a sentence-based approach adopted initially in the DyGIE/DyGIE$++$ architectures. It also drives some changes such as the use of a single \textit{pruner} (see \secref{sec:shared_model}) to extract spans used in coreference and 
    RE
    modules. Similarly, instead of applying the shared BiLSTM sentence by sentence as in \cite{luan2019general} and \cite{wadden2019entity}, we do it on the entire document, in order to allow capturing cross-sentence dependencies for document-level relations and entity clusters in \datasetname.
    \item We add an additional decoding step (see \secref{sec:decoding}) needed to transform mention-based predictions for 
    RE
    and NER tasks into entity-based ones, as required by the entity-centric nature of \datasetname, and propose corresponding evaluation metrics (see \secref{sec:metrics}).
    \item Finally, we make changes in the loss and prediction components to support multi-label classification (in NER and RE) as required in \datasetname. 
\end{enumerate}

\subsection{Span-Based Representation}
\label{sec:shared_model}
The input to our model consists of document-level annotation instances. Each document $D$ from the considered document collection $\mathcal{D}$ is represented by its sequence $T$ of tokens. 
These tokens are represented internally as a concatenation of GloVe~\citep{pennington2014glove} and character embeddings~\citep{ma2016end}. We also experiment with additionally concatenating BERT~\citep{devlin2019bert} contextualized embeddings. Since BERT is run on a sub-token level, to the representation of each token we only concatenate the BERT-based representation of the first sub-token, as originally proposed by \cite{devlin2019bert}. This input is fed into a BiLSTM layer in order to obtain the output token representations by concatenating the forward and backward LSTM hidden states. The BiLSTM outputs for the considered document $D$ are written on the token level as $\textbf{e}_i\in\mathbb{R}^m\;(i=1,\ldots,\vert T\vert)$.
These are converted into \textit{span representations}. The set of all possible spans for $D$, up to maximum span width $w_{\mathrm{max}}$ (which is a hyperparameter of the model), is written as
$S= \{s_1, \dotsc, s_{|S|}\}$.
The number of spans 
can be calculated as follows, 
\begin{equation}
|S| = \sum\limits_{k=1}^{w_{\mathrm{max}}} |T| - k + 1 \;={w_{\mathrm{max}}}\left(|T| - \frac{w_{\mathrm{max}}-1}{2}\right)\label{eq:spans}\end{equation}
We obtain the representation $\textbf{g}^0_i$ for span $s_i$, ranging from token $l$ to token $r$, by concatenating their respective BiLSTM states $\textbf{e}_{l}$ and $\textbf{e}_{r}$ with an embedding $\boldsymbol{\psi}_{r-l}$ for the span width $w_i = r-l$
\begin{equation}
\textbf{g}^0_i = [\textbf{e}_{l}; \textbf{e}_{r}; \boldsymbol{\psi}_{r-l}] \label{eq:spanrepr}
\end{equation}

As seen from \equref{eq:spans}, the number of possible spans scales approximately linearly with the maximum span width $w_{\mathrm{max}}$, as well as the document length $\vert T\vert$ (assuming $w_{\mathrm{max}}\ll\vert T\vert$). This leads to a strongly increased set of spans, as compared to previous works where $\vert S\vert$ scales with the length of individual sentences rather than entire documents~\citep{luan2019general,wadden2019entity}.
In order to mitigate the required memory of our model, 
we use a shared \textit{pruner} to reduce $S$ to a smaller set $P$ of candidate spans
to be used by the coreference and 
RE
scorers and in the graph propagation modules (see further). The choice of using a single pruner contrasts with similar work in \cite{luan2019general} and \cite{wadden2019entity} where two separate pruners are used, one for the relation task, and another for coreference. Our design choice is based on the fact that both of these tasks 
use the same document-level entity mentions. This contrasts with datasets used in \cite{luan2019general} and \cite{wadden2019entity} where, while the coreference is defined on the document-level, the relations are sentence-based. 

Finally, we use graph propagation to iteratively refine the pruned spans representations. Three graph propagation mechanisms are compared in the experiments. Our own contribution is the attention-based graph propagation method \propformat{AttProp}, where the span representations are updated in $\tau_A$ iterations. Alternatively, $\tau_C$ iterations of \propformat{CorefProp}~\citep{lee2018higher,luan2019general} can be performed, or $\tau_R$ iterations of \propformat{RelProp}~\citep{luan2019general}.  

The span representation of a particular span $s_i$ after iteration $t$ is denoted as
$\textbf{g}_i^t$ in our notation. The details of graph propagation are explained in \secref{sec:graph_prop_model}. Note that in theory several of these graph propagation techniques could be accumulated, but in our setting the benefits thereof in terms of model effectiveness were minor, at a significantly higher computational cost. Therefore, in our experiments, we only compare models without graph propagation with models applying a single form of graph propagation. 
To keep the sections introducing the models clear, we will write $\tau$ to denote the number of propagations in general (which could be 0, or any of ${\tau_A, \tau_C \mathrm{\ or\ } \tau_R}$, depending on the chosen experiments and considered model components).

\subsection{Joint Model for Entity Recognition, Coreference Resolution, and Relation Extraction}
\label{sec:joint_training}

In this section, we present the joint model including recognition of entity mentions as belonging to $L_T$ types (introduced as NER), the clustering of the entity mentions into entities (coreference resolution), and identifying relations between entities, all on the document level. The building blocks responsible for the three subtasks are discussed next, as well as the total loss of the joint model. 
The details of the graph propagation mechanisms are then provided further on (\secref{sec:graph_prop_model}). 

\subsubsection{Entity Mention Module} \label{subsubsec:entity_module}
All spans $s_i$ (up to width $w_\mathrm{max}$) of the considered document\footnote{For convenience, the subscript $D$ indicating the current document is left out in the equations of this section.} are scored by feeding their representation (starting from \equref{eq:spanrepr} and potentially updated after $\tau$ graph propagation iterations) into the feed-forward neural network (FFNN) written
as
$\mathbfcal{F}_{\text{mention}}$, with as many outputs as there are entity types:
\begin{align}
    \boldsymbol{\Phi}^{\tau}_{\mathrm{mention}}(s_i) = \mathbfcal{F}_{\mathrm{mention}}(\textbf{g}^{\tau}_i). \label{eq:ner1}
\end{align}
Throughout this section, we will maintain the same notation of $\mathbfcal{F}(\mathbf{x})$ to denote a FFNN that takes as input a vector $\mathbf{x}$ and produces a vector of scores, and $\mathcal{F}(\mathbf{x})$ 
 to refer to a FFNN with a scalar output.

The probability of each label being valid for the considered span is modeled by component-wise application of a sigmoid ($\sigma(x)=1/(1+e^{-x})$) to these scores $\boldsymbol{\Phi}^\tau_{\mathrm{mention}}(s_i)\in \mathbb{R}^{L_T}$ (with $L_T$  the number of entity 
tags). 
The log probability of the ground truth mention labels for all spans of document $D$ is given by
\begin{equation}
    \log P_{\mathrm{mention}}\left(E^*\vert G^\tau\right) = \sum_{i=1}^{\vert S\vert}\sum_{l=1}^{L_T} I_{i,l} \log\sigma(\boldsymbol{\Phi}^{\tau}_{\mathrm{mention}}(s_i)_l)
    + (1-I_{i,l}) \log\big( 1 - \sigma\left(\boldsymbol{\Phi}^{\tau}_{\mathrm{mention}}(s_i)_l\right)\big),
    \label{eq:ner2}
\end{equation}
in which $E^*$ represents the set of ground truth mention labels for all spans in the document, and $I_{i,l}\in\{0,1\}$ is the ground truth indicator label for mention 
tag
$l$ of span $s_i$. $G^\tau$ denotes the set of all considered span representations for the current document. The superscript $\tau$ reflects the fact that, in case graph propagation is applied, the subset of $\vert P\vert$ representations (for the spans retained after pruning) have been updated over $\tau$ iterations.  By summing over all entity types ($l=1,\ldots,L_T$), we account for the fact that a particular span can have multiple associated entity tags (i.e., the considered NER task is multi-label). 
At inference time, spans get assigned those entity types for which the corresponding score $\boldsymbol{\Phi}^{\tau}_{\mathrm{mention}}(s_i)>0$. Note that not all valid entity mentions necessarily get an entity type assigned: if the relation extractor determines that a span is part of a relation, it effectively becomes an entity mention, even if none of the pre-defined types is considered applicable by the entity scorer.

\subsubsection{Coreference Module}\label{subsubsec:coref}
While the entity scoring is performed on all span representations $S$, this is not possible for the coreference and relation scorers, due to memory limitations. The latter scorers predict on \textit{pruned spans}, as shown in \figref{fig:model_architecture}. How the pruner is trained jointly with the model, is described in \secref{subsubsec:pruner}. 
In order to avoid confusion by introducing additional notations, we list the spans in the pruned set $P$ as $s_1,\ldots,s_{\vert P\vert}$, according to their original order in the text. 

The module for coreference resolution is based on pairwise scoring of the pruned spans from~$P$. Following ideas from \cite{lee2017end,lee2018higher,luan2018multi,luan2019general}, 
for any span $s_j$, scores with respect to each of the preceding (also referred to as `antecedent') spans $s_i\;(i\leq j)$ in the document are calculated with a neural network $\mathcal{F}_{\mathrm{coref}}$:
\begin{equation}
\Phi^\tau_{\mathrm{coref}}(s_i,s_j) = \mathcal{F}_{\mathrm{coref}} \left([\textbf{g}^\tau_i; \textbf{g}^\tau_j; \textbf{g}^\tau_i \odot \textbf{g}^\tau_j; \boldsymbol{\varphi}_{i,j}]\right).
\label{eq:coref1} 
\end{equation}
This expression scores the compatibility between spans $s_i$ and $s_j$, taking as input the concatenation of their respective span representations (after $\tau$ propagation iterations), their component-wise product, and an embedding $\boldsymbol{\varphi}_{i,j}$ representing their distance in terms of the number of ordered candidate spans from $s_i$ to $s_j$. 

In order to deal with non-coreferent or incorrect spans, previous work in span-based coreference~\citep{lee2017end,lee2018higher} defines a dummy antecedent $\epsilon$ to which all non-coreferent or invalid spans point.  
While this approach is effective in datasets that do not contain singleton entity clusters, such as OntoNotes-based CoNLL-2012~\citep{pradhan2012conll}, it does not allow to distinguish between valid singleton entity mentions and invalid mention spans. 
This makes it unsuitable to use on \datasetname, since it contains singleton entity clusters, consisting of a single mention. In fact, 66.4\% of the entity clusters in {\datasetname} are singletons. Furthermore, the current official CoNLL-2012 evaluation script\footnote{\url{https://github.com/conll/reference-coreference-scorers}} based on \cite{pradhan2014scoring} accounts for scenarios where either the dataset or the predicted mentions are singletons, which 
has a direct impact on the established
B-CUBED~\citep{bagga1998algorithms} and CEAF$_{\text{e}}$~\citep{luo2005coreference} coreference scores. 
In order to tackle the singleton entity cluster detection in our coreference model, we 
propose to start from $\Phi^\tau_{\mathrm{coref}}(s_j,s_j)$\footnote{This would be replaced with $\Phi^\tau_{\mathrm{coref}}(\epsilon, s_j)$ in the \textit{dummy-based} formulation defined in \cite{lee2017end}.} as a self-coreference span score. By applying the correct target in the coreference loss, it allows indicating that either the span $s_j$ is not a valid mention, or that it is a valid mention that is not co-referenced with any antecedent span. 

The log probability of the ground truth coreference labels of document $D$ is given by
\begin{equation}
\log P_{\mathrm{coref}}\left(C^*\vert G^\tau\right) = \sum_{j=1}^{\vert P\vert} \log \frac{ \sum\limits_{s^*\in S_j^*}\exp\left(\Phi^\tau_{\mathrm{coref}}(s^*,s_j)\right)  }{ \sum\limits_{i=1}^j\exp\left(\Phi^\tau_{\mathrm{coref}}(s_i,s_j)\right)}.
\label{eq:coref2}
\end{equation}
The set of ground truth coreference labels is indicated as $C^*$. The summation over $j$ represents the contribution to the log likelihood of the correct antecedent labels for each span $s_j$ in the pruned set $P$. The individual terms in the right-hand side correspond to the log probability of the correct antecedent labels for a particular span $s_j$.  In the denominator, the summation ranges from the first span, up to span $s_j$ itself (i.e., for the self-coreference score), but not beyond it (given that only \emph{antecedents} in the sorted sequence of pruned spans are considered). The numerator contains the contributions from the potentially multiple ground truth antecedents for span $s_j$. This stems from the fact that multiple antecedent mentions may belong to the same cluster as $s_j$, which all contribute to the probability of the correct antecedent labels. The set of ground truth antecedents corresponding to span $s_j$ is written $S_j^*$.

At inference time, the highest scoring antecedent for span $s_j$ (including $s_j$ itself) is picked. Due to the idea of only predicting \emph{antecedents}, picking any of the ground truth antecedents leads to the correct mention clusters
\citep{durrett2013easy, wiseman2015learning,lee2017end,lee2018higher}.

\subsubsection{Relation Module}
\label{subsubsec:relation}
Similar to the coreference module (\equref{eq:coref1}), we score span pairs using an FFNN 
\begin{align}
\boldsymbol{\Phi}^\tau_{\mathrm{relation}}(s_i,s_j) = \mathbfcal{F}_{\mathrm{relation}} \left([\textbf{g}^\tau_i; \textbf{g}^\tau_j; \textbf{g}^\tau_i \odot \textbf{g}^\tau_j; \boldsymbol{\varphi}_{i,j}]\right),
\label{eq:rel1} 
\end{align}
where
$\boldsymbol{\varphi}_{i,j}$ is again the distance embedding as introduced in \secref{subsubsec:coref}. 
$\boldsymbol{\Phi}^\tau_{\mathrm{relation}}(i,j) \in \mathbb{R}^{L_R}$ is a vector representing relation span pair scores
for each of the $L_R$ possible relation types between spans $s_i$ and $s_j$.

The log probability of the ground truth relation labels of document $D$ is given by
\begin{align}
\log P_{\mathrm{relation}} \left(R^* \vert G^\tau \right) = \sum\limits_{{i,j=1}}^{|P|} \sum\limits_{l=1}^{L_R}  {I_{i,j,l} \log \sigma(\boldsymbol{\Phi}^\tau_{\mathrm{relation}}(s_i,s_j)_l) + (1-I_{i,j,l}) \log\left(1 - \sigma(\boldsymbol{\Phi}^\tau_{\mathrm{relation}}(s_i,s_j)_l)\right)},
\label{eq:rel2} 
\end{align}
in which $R^*$ represents the set of ground truth relation labels for all combination of pruned span pairs in the document, and $I_{i,j,l} \in \{0,1\}$ is the ground truth indicator label for relation type $l$ of the span pair $(s_i, s_j)$. Note that all $\vert P\vert^2$ pruned span pairs are considered, since the order of the spans in the relation matters (unlike the coreference case).
By summing over all possible relation types $L^R$, we account for the fact that a particular relation between two spans can be multi-label (which is the case for more than 30\% of relations, as shown in \Tabref{tab:relations_multilabel_stats}).

Since this model is run in parallel with the coreference module, it is used to predict relations only between entity mentions and not entity clusters. During inference, candidate relations are accepted when $\boldsymbol{\Phi}^\tau_{\mathrm{relation}}(s_i, s_j)_l > 0$.

\subsubsection{Span Pruner}\label{subsubsec:pruner}
The span pruner is an FFNN, denoted $\mathcal{F}_{\text{pruner}}$, that scores all spans $s_i$ based on their initial representation $\mathbf{g}_i^0$, after which only the highest scoring spans are retained in the pruned span set $P$.  In our experiments $P$ contains the top $0.2\,\vert T\vert$ highest scoring spans, which covers more than $98\%$ of all the ground truth mention spans in the \datasetname~dataset. We represent the pruner score for span $s_i$ as 
\begin{align}
     \Phi_{\mathrm{pruner}}\left(s_i\right) = \mathcal{F}_{\mathrm{pruner}} \left(\textbf{g}^0_i\right).
     \label{eq:pruner1} 
\end{align}
Several strategies can be used to train the pruner. One option is to directly optimize the probability of the pruner to detect the spans of correct entity mentions. With $S^*$ the set of spans with at least one ground truth entity type, and $I_i\in\{0,1\}$ an indicator for whether $s_i\in S^*$, the corresponding log likelihood can be written as 
\begin{equation}
\log P_{\mathrm{pruner}} \left(S^*\vert G^0 \right) = \sum_{i=1}^{\vert S\vert} I_i \log\sigma\left({\Phi}_{\mathrm{pruner}}(s_i)\right)
    + (1-I_i) \log\left( 1 - \sigma\left({\Phi}_{\mathrm{pruner}}(s_i)\right)\right),
    \label{eq:pruner_loss}  
\end{equation}
leading to a separate pruner loss term. 
Alternatively, the pruner can be trained indirectly by adapting the mention score from \equref{eq:ner1}, the coreference score from \equref{eq:coref1} or the relation score from \equref{eq:rel1} as follows:
\begin{align}
     \tilde\Phi_{\mathrm{mention}}^\tau\left(s_i\right) &= \Phi_{\mathrm{mention}}^\tau\left(s_i\right) + \Phi_{\mathrm{pruner}}\left(s_i\right)
     \label{eq:pruner2}
     \\
     \tilde\Phi_{\mathrm{coref}}^\tau\left(s_i, s_j\right) &= \Phi^\tau_{\mathrm{coref}}\left(s_i, s_j\right) + \Phi_{\mathrm{pruner}}\left(s_i\right)
     \label{eq:pruner3}
     \\
    \tilde\Phi_{\mathrm{relation}}^\tau\left(s_i, s_j\right) &= \Phi^\tau_{\mathrm{relation}}\left(s_i, s_j\right) + \Phi_{\mathrm{pruner}}\left(s_i\right)
    \label{eq:pruner4}      
\end{align}
for use in the expressions \equref{eq:ner2}, \equref{eq:coref2} and \equref{eq:rel2}, respectively. As such, higher pruner scores would directly correspond to higher mention or coreference scores, and lead to a meaningful ranking of spans according to pruner scores. 
All three strategies seem to work on a similar level, but for the presented joint model experiments, we use the indirect training through the coreference module, as in \equref{eq:pruner3}. Note that we did not experiment with training the pruner through the relation module, because it would be trained only on those spans involved in relations, which is a mere subset of all valid mentions.

\subsubsection{Joint Model}\label{subsubsec:joint_model}
We perform joint training in order to explore the degree to which the graph propagation techniques (see  \secref{sec:graph_prop_model}) affect related tasks in \datasetname. For instance, we expect that performing a coreference propagation can have a positive impact on the NER task. We hypothesize that enriching the entity spans with broader contextual information coming from other mention spans in the cluster, can improve the effectiveness of the entity module.
Furthermore, given the entity-centric nature of \datasetname, the mention-based predictions for NER and RE
have to be grouped in coreference clusters (see section \ref{sec:decoding} for details), which makes it necessary to execute these tasks jointly with the coreference task. 

The joint loss for each document $D$ is a weighted sum of the individual loss functions of the subtasks: 
\begin{equation}
    \mathcal{L}^{\mathrm{joint}}_D = \sum_{(E^*, C^*, R^*)} 
    \lambda_E \log P_{\mathrm{mention}}\left(E^*\vert G^\tau\right) 
    +\lambda_C \log P_{\mathrm{coref}}\left(C^*\vert G^\tau\right)
    +\lambda_R \log P_{\mathrm{relation}} \left(R^* \vert G^\tau \right),
    \label{eq:jointloss}
\end{equation}
in which $\lambda_E$, $\lambda_C$, and $\lambda_R$ are hyperparameters of the joint model.

\subsection{Decoding and Prediction}
\label{sec:decoding}
Unlike previous datasets used in span-based predictions~\citep{luan2018multi,kulkarni2018annotated,walker2006ace,doddington2004automatic} where the relation and entity extraction 
are done on the mention-level, \datasetname~is an entity-centric dataset. During inference, this requires an additional decoding step to cluster the mention-based span-dependent predictions into entity-centric ones. The component responsible for this decoding in the proposed architecture is the \textit{entity-centric decoder} (see \figref{fig:model_architecture}). The pseudo-code in \Algref{alg:entity_centric} summarizes the steps performed by this component. First, the decoder receives as \textit{input} 
the predicted span clusters ($p\_cl$), entity mentions ($p\_men$) and relations between spans ($p\_rel$) obtained from the scores calculated in 
\equref{eq:pruner3},
\equref{eq:ner1} and \equref{eq:rel1}, respectively. Next, the predicted entity mentions are 
connected with
the respective clusters by using the dictionary $C$ that maps mention spans to cluster ids (lines 3--12 in \Algref{alg:entity_centric}). 
Specifically, 
each of the entity clusters is assigned the union of the entity types predicted for any of the 
mention spans 
inside the cluster (line 11 in \Algref{alg:entity_centric}). If the predicted entity mention can not be located inside the predicted clusters, a new singleton cluster is added (lines 5--6 in \Algref{alg:entity_centric}). Finally, all the pairwise predicted relations on the mention level ($p\_rel$) between members of two different clusters are assigned as predicted relations between the (cluster-level) entities (lines 13--20 in \Algref{alg:entity_centric}). Similarly as with entity mentions, the dictionary $C$ is used to map the 
mention spans ($span\_h$ and $span\_t$) of a particular relation type $rel\_type$ to the corresponding cluster ids. 
Furthermore, the relations added between two clusters are the union of all the relations predicted between 
any pair of
mentions 
inside these clusters (line~18 in \Algref{alg:entity_centric}). 

\renewcommand\baselinestretch{1}\small
\begin{algorithm}
\textbf{Input:} predicted clusters ($p\_cl$), entity mentions ($p\_men)$ and relations between mentions ($p\_rel$): \vspace*{-2mm}
\begin{enumerate}
    \item $p\_cl$ is a dictionary (map) that maps cluster ids to mention spans \vspace{-3mm}
    \item $p\_men$ is list of tuples $\langle$predicted span, predicted tag$\rangle$ \vspace{-3mm}
    \item $p\_rel$ is list of tuples $\langle$predicted head span, predicted relation, predicted tail span$\rangle$ 
\end{enumerate}\vspace*{-1mm}
\textbf{Output:} clusters ($p\_cl$), decoded entities ($d\_ent$) and relations between entities ($d\_rel$)
  \begin{algorithmic}[1]
\State Initialize $d\_ent, d\_rel\gets\mathrm{empty\ dictionary\ (map)}$ 
\State $C \gets$ transformed $p\_cl$ that maps spans to cluster ids \vspace*{1mm}
\LeftComment{Decode entity mentions ($p\_men$) to entities ($d\_ent$) (lines 3--12)}
\For{$span,tag$ \textbf{in} $p\_men$}
    \If{$span$ \textbf{not in} $C.keys()$}
        \State $C[span] \gets$ new concept id
        \State $p\_cl[C[span]] \gets \mathrm{list}([span])$
    \EndIf
    \If{$C[span]$ \textbf{not in} $d\_ent.keys()$}
        \State $d\_ent[C[span]] \gets$ empty set
    \EndIf
    \State $d\_ent[C[span]].\mathrm{add}(tag)$
\EndFor
\vspace*{1mm}
\LeftComment{Decode relations between mentions ($p\_rel$) to relations between entities ($d\_rel$) (lines 13--20)}
\For {$span\_h, rel\_type, span\_t$ \textbf{in} $p\_rel$}
    \If{$(span\_h$ \textbf{in} $C.keys())$ \textbf{and} $(span\_t$ \textbf{in} $C.keys())$}
        \If{$\langle C[span\_h],C[span\_t] \rangle$ \textbf{not in} $d\_rel.keys()$}
            \State $d\_rel[\langle C[span\_h],C[span\_t] \rangle] \gets$ empty set
        \EndIf
        \State $d\_rel[\langle C[span\_h],C[span\_t] \rangle].\mathrm{add}(rel\_type)$
    \EndIf
\EndFor
 \caption{Entity-centric decoder for the \textit{Joint} model.}\label{alg:entity_centric}
\end{algorithmic}
\end{algorithm}
\renewcommand\baselinestretch{1.0}\normalsize

\subsection{Graph Propagation Mechanisms}\label{sec:graph_prop_model}
In order to evaluate the impact of graph-based propagation of contextual information between the spans, 
we propose \propformat{AttProp}, and reimplement the \propformat{CorefProp} and \propformat{RelProp} graph propagation algorithms. 
\cite{lee2018higher} proposed the gated graph propagation update function for use on coreference resolution, which was then successfully applied in a joint multi-task setting by \cite{luan2019general,wadden2019entity}.
The graph propagation equations are written as:
\begin{align}
    \textbf{f}^t_x(s_i) &= \sigma\left(\mathbfcal{F}_{x} ([\textbf{g}^t_i; \textbf{u}^t_x(s_i)])\right), \label{eq:update_spans1}
    \\ 
    \textbf{g}_i^{t+1} &= \textbf{f}^t_x(s_i) \odot \textbf{g}^t_i + \left(1 - \textbf{f}^t_x(s_i)\right) \odot \textbf{u}^t_x(s_i),
    \label{eq:update_spans2}
\end{align}
where in our case $x \in \{A, C, R\}$ denotes \propformat{AttProp}, \propformat{CorefProp}, and \propformat{RelProp}, respectively.
The $n$-dimentional vector $\textbf{f}^t_x(s_i)$, produced by the single-layer FFNN $\mathbfcal{F}_{x}$ can be interpreted as 
a gating vector that acts as a switch between the current span representations $\textbf{g}^t_i \in \mathbb{R}^n$, and the update span vector $\textbf{u}^t_x(s_i) \in \mathbb{R}^n$. The various graph propagation methods differ in how $\textbf{u}^t_x(s_i)$ is calculated.

\boldpartitle{\propformat{CorefProp}} The coreference confidence score between span $s_i$ and $s_j$ for propagation iteration \textit{t} is denoted as $P^t_C(s_i,s_j)$ and calculated as follows,
\begin{align}
    P^t_C(s_i,s_j) = \dfrac{\text{exp}\left(\tilde\Phi^t_{\mathrm{coref}}(s_i,s_j)\right)}{\sum\limits_{i'=1}^{j} \text{exp}\left(\tilde\Phi^t_{\mathrm{coref}}(s_{i'},s_{j})\right)},
    \label{eq:coref_propagation} 
\end{align}
in which $i' \in \{1, \dotsc, j\}$ refers to all antecedent spans $s_{i'}$ to span $s_j$ in the pruned span set. 
Note that the coreference scores according to \equref{eq:pruner3} are used. This means the confidence scores not only reflect whether the considered spans are compatible, but also whether the individual spans are likely to be retained by the pruner as potential entity mentions.
In order to perform a \propformat{CorefProp} graph iteration, the span update vector $\textbf{u}^t_C(i) \in \mathbb{R}^n$ is first calculated as a weighted average of the current representation of span $s_j$ and all of its antecedents 
\begin{align}
    \textbf{u}^t_C(s_j) = \sum\limits_{i=1}^{j}P^t_C(s_i,s_j) \> \textbf{g}^t_i,
    \label{eq:coref_update_vector} 
\end{align}
in which the weighting coefficients quantify the coreference compatibility of the corresponding span with $s_j$.
After that, the update equations \equref{eq:update_spans1} and \equref{eq:update_spans2} are applied.

\boldpartitle{\propformat{RelProp}} Similarly as with \propformat{CorefProp},
a relation span update vector is calculated as formalized next,
\begin{align}
    \textbf{u}^t_R(s_j) = \sum\limits_{i=1}^{|P|} \left(\textbf{A}_R\; f\left(\boldsymbol{\Phi}^t_{\mathrm{relation}}(s_i,s_j)\right) \right)   \>  \odot \textbf{g}^t_i, \label{eq:relation_update_vector}    
\end{align}
where $\textbf{A}_R \in \mathbb{R}^{n\times L_R}$ is a trainable projection tensor, and $f$ is a non-linear activation function (ReLU). Similarly as in \equref{eq:coref_update_vector}, the update vector can be interpreted as a weighted sum of all span representations, with the additional expressiveness stemming from the projection matrix $\textbf{A}_R$ in accounting for the relation scores.

\boldpartitle{\propformat{AttProp}} In order to measure the impact of the `supervised' \propformat{CorefProp} and \propformat{RelProp} propagation techniques described by equations (\ref{eq:coref_propagation})-(\ref{eq:relation_update_vector}) above, we introduce a latent attentive propagation. Unlike \propformat{CorefProp} and \propformat{RelProp} that are driven by the task-specific confidence propagation scores $\Phi_{\mathrm{coref}}^t(s_i,s_j)$ and $\boldsymbol{\Phi}_{\mathrm{relation}}^t(s_i,s_j)$, \propformat{AttProp} is influenced only by latent attention weights between all the pruned spans $P$ calculated as follows,  
\begin{align}
\Phi^t_{\mathrm{att}}(s_i,s_j) = \mathcal{F}_{\mathrm{att}} \left([\textbf{g}^t_i; \textbf{g}^t_j; \textbf{g}^t_i \odot \textbf{g}^t_j; \boldsymbol{\varphi}_{i,j}]\right), \label{eq:att_scores} 
\end{align}
where $\boldsymbol{\varphi}_{i,j}$ is the distance feature embedding function between spans $s_i$ and $s_j$, and $\Phi^t_{\mathrm{att}}(s_i,s_j)$ is the attention score between these spans.
This score is normalized with a softmax to get the $P^t_{A}(s_i,s_j)$ confidence score
\begin{align}
    P^t_{A}(s_i,s_j) = \dfrac{\text{exp}\left(\Phi^t_{\mathrm{att}}(s_i,s_j)\right)}{\sum\limits_{j'=1}^{|P|} \text{exp}\left(\Phi^t_{\mathrm{att}}(s_i,s_{j'})\right)}.
    \label{eq:att_propagation} 
\end{align}
The span update vector $\textbf{u}^t_A(s_i) \in \mathbb{R}^n$ 
is calculated as a weighted sum of all the $P$ span representations as opposed to only antecedents in \propformat{CorefProp}
\begin{align}
    \textbf{u}^t_A(s_i) = \sum\limits_{j=1}^{|P|}P^t_A(s_i,s_j) \> \textbf{g}^t_j.
    \label{eq:att_update_vector} 
\end{align}

\subsection{Single Task Models}
\label{sec:single_task_models}

In this section we shortly describe independent baseline models for the three individual core tasks under study in this paper, as training these models not entirely corresponds to merely minimizing the corresponding loss term from the total loss \equref{eq:jointloss}.

\subsubsection{Single Entity Recognition Model}
\label{sec:single_ner_model}
The single-task NER model is designed for detecting and correctly labeling the individual entity spans, and is based on \equref{eq:ner2}. However, even for the single models, the graph propagation mechanism \propformat{AttProp} may be useful, but for that the pruner needs to be jointly trained with the model. This is obtained by augmenting the mention loss $-\log P_{\mathrm{mention}}(E^*\vert G^\tau)$ with the pruner loss $-\log P_{\mathrm{pruner}} \left(S^*\vert G^0 \right)$ according to \equref{eq:pruner_loss}. 

\subsubsection{Single Coreference Resolution Model}
\label{subsubsec:single_coref}
The single-task end-to-end coreference model needs to detect mentions and correctly cluster them. Here again, the standard coreference loss $-\log P_{\mathrm{coref}}\left(C^*\vert G^\tau\right)$ according to 
\equref{eq:pruner3} and
\equref{eq:coref2} is extended with the  pruner loss $-\log P_{\mathrm{pruner}} (S^*\vert G^0 )$.
This turned out essential for correctly predicting the singleton clusters. 

\subsubsection{Single Relation Extraction Model}
The single relation extraction model is trained to detect mentions as well as the correct pairwise relations between mentions (i.e., without the coreference step). 
In order to train the pruner as well, the standard relation score is extended as described in \equref{eq:pruner4}
before calculating the loss $- \log P_{\mathrm{relation}} \left(R^* \vert G^\tau \right)$ based on \equref{eq:rel2}. 

\section{Entity-Centric Metrics}
\label{sec:metrics}

\begin{figure}
\centering
\includegraphics[width=0.9\columnwidth]{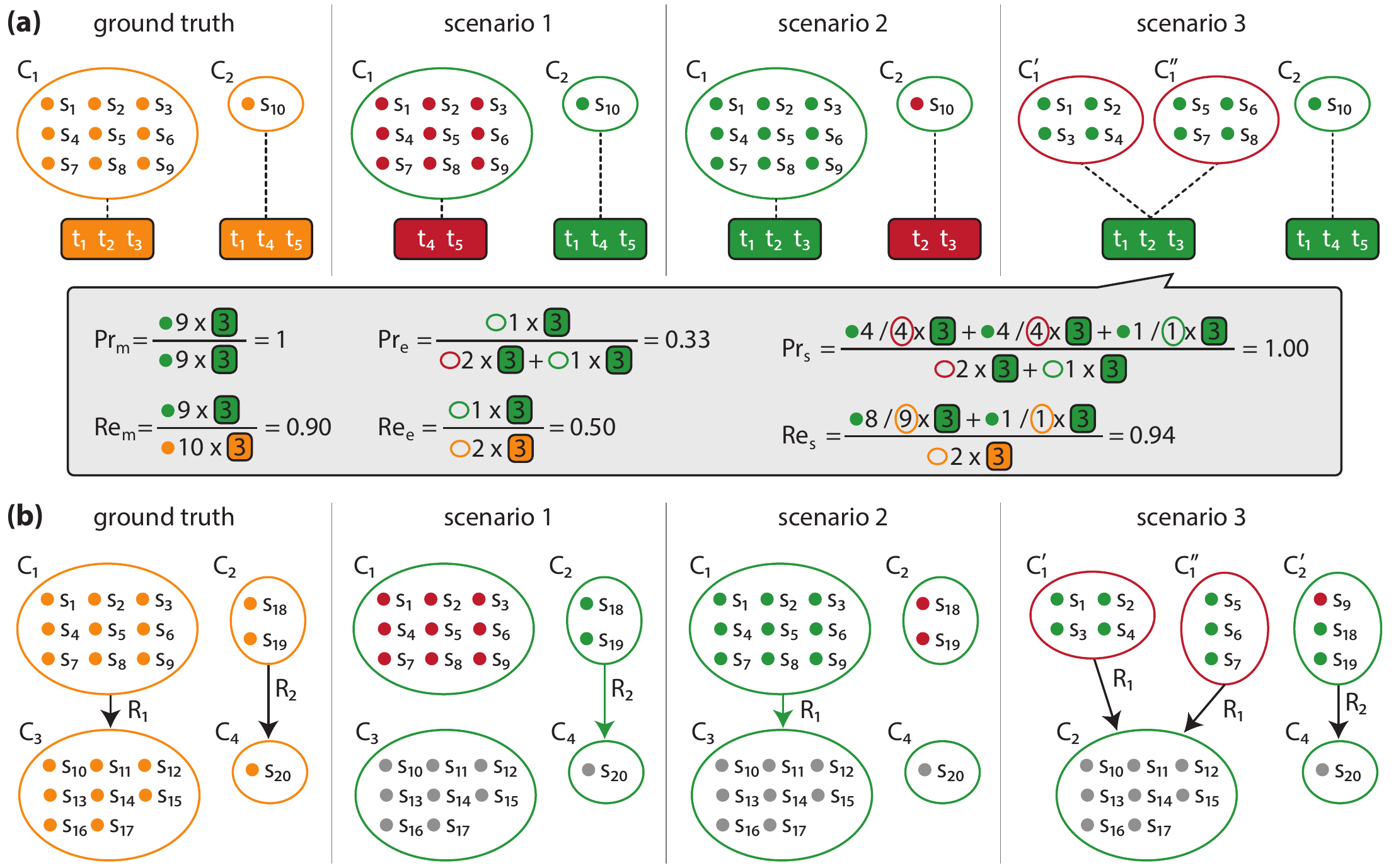}
\caption{Illustration of entity prediction scenarios for \textbf{(a)}~NER and \textbf{(b)}~relation extraction, with large clusters ($C_1, C_3$) and smaller ones ($C_2, C_4$). \emph{Scenario 1} erroneously labels the large one, \emph{scenario 2} incorrectly labels the small one, \emph{scenario 3} incorrectly splits up the large one and makes a mistake for one of its mentions, $s_9$. The formulas in the grey box illustrate the calculation of mention-level 
($\mathrm{Pr_m}$, $\mathrm{{Re}_m}$),
hard entity-level ($\mathrm{{Pr}_e}$, $\mathrm{{Re}_e}$) and soft entity-level ($\mathrm{{Pr}_s}$, $\mathrm{{Re}_s}$) precision and recall for NER in \emph{scenario 3}. Note that in~(b), the mention dots are colored for correct (green) and incorrect (red) relation heads only.}
\label{fig:metric_motivation}
\end{figure}

Unlike the currently widespread datasets that use a mention-driven approach to annotate named entities \citep{sang2003introduction,derczynski2017results,weischedel2011ontonotes,bekoulis2017reconstructing}, relations \citep{augenstein2017semeval,song2015light,doddington2004automatic,ji2017overview,kim2003genia,luan2018multi, bekoulis2017reconstructing} and entity linking \citep{bentivogli2010extending,riedel2010modeling,hoffart2011robust}, \datasetname~is entirely entity-centric.
As explained before, we group entity mentions $s_i$ referring to the same entity into clusters $C_k$. While we can, and will, adopt the traditional coreference measures as defined by \citet{pradhan2014scoring} to judge this cluster formation, the NER and relation extraction (RE) evaluation (using precision, recall and $\mathrm{F_1}$) can be done either on
\begin{enumerate*}[(i)]
    \item mention level, or
    \item entity (cluster) level.
\end{enumerate*}
The first option however would have the metrics being dominated by the more frequently occurring entities, while the second would penalize mistakes in the clustering (since partially correctly identified clusters would be seen as completely incorrect). This is illustrated in \figref{fig:metric_motivation} and the corresponding performance metrics in \tabref{tab:metric_motivation}, where scenarios~1 and 2 highlight the effect of making labeling mistakes on the cluster level for different sizes, and scenario~3 highlights the pessimistic view of hard entity-level metrics in case of clustering mistakes. Note that we indicate the mention-level metrics with subscript $m$, while the (hard) entity-level metrics will have subscript with $e$.

Because the (hard) entity-level metrics in our opinion overly penalize clustering mistakes 
(cf.\ scenario~3), 
we propose a variant of entity-level evaluation which we term \emph{soft} entity-level metrics (denoted by subscript~$s$). 
Basically, instead of adopting a binary count of 1 (all mentions correct) or 0 (as soon as a single mention is missed) on an entity cluster level, we rather count the fraction of its mentions that are correctly labeled. This is illustrated in the formula part of \figref{fig:metric_motivation}(a) for NER, and below we present the adopted formulas in detail. Note that in case clusters are completely predicted correctly, the soft entity-level metrics are the same as hard entity-level metrics (and thus avoid the metric being dominated by frequent mentions, as in the mention-level case).

The formal definition of the metrics depends on counting true positives $\textit{tp}_p(l)$ and $\textit{tp}_g(l)$, false positives $\textit{fp}(l)$, and false negatives $\textit{fn}(l)$ for a particular NER tag/relation type $l$, which are specified in \eqsref{eq:soft_m_tp_generic}{eq:soft_m_negatives_generic}. These and other notation definitions are summarized in \tabref{tab:metric_symbols}. 
Further, note that we define two true positives for a particular label $l$, because of the potential difference between predicted and ground truth clusters: $\textit{tp}_p(l)$ 
sums
fractions of \emph{predicted} clusters and is used to calculate the precision $\mathrm{Pr}_\mathrm{s}$ in \equref{eq:soft_metrics}, while $\textit{tp}_g(l)$ considers \emph{ground truth} clusters and is used for the recall $\mathrm{Re}_\mathrm{s}$ in \equref{eq:soft_metrics}.
This allows us to preserve the \textit{cluster-based} relationships between true positives, false positives and false negatives as described for expressions $\textit{tp}_p(l) + \textit{fp}(l)$ and $\textit{tp}_g(l) + \textit{fn}(l)$ in \Tabref{tab:metric_cluster_sum}. Thus our soft entity-level metrics are still cluster-based, while accounting for the mention-level predictions.

{\renewcommand\baselinestretch{1}\begin{table}[]
\centering
\caption{Comparison of different metrics for the example scenarios depicted in \figref{fig:metric_motivation}, for \textbf{(a)}~NER and \textbf{(b)}~relation extraction.}
\label{tab:metric_motivation}
\begin{tabular}{l l ccc ccc ccc}
  \toprule 
& & \multicolumn{3}{c}{\textbf{Mention-Level}} &  \multicolumn{3}{c}{\textbf{Hard Entity-Level}} & \multicolumn{3}{c}{\textbf{Soft Entity-Level}} \\
\cmidrule(lr){3-5} \cmidrule(lr){6-8} \cmidrule(lr){9-11} 
 & & $\mathbf{Pr_m}$ & $\mathbf{Re_m}$ & $\mathbf{F_{1,m}}$
 & $\mathbf{Pr_s}$ & $\mathbf{Re_s}$ & $\mathbf{F_{1,s}}$
 & $\mathbf{Pr_e}$ & $\mathbf{Re_e}$ & $\mathbf{F_{1,e}}$
\\
\midrule
\multirow{4}{*}{\textbf{(a)~NER}} & Ground Truth & 1.000 & 1.000 & 1.000 & 1.000 & 1.000 & 1.000 & 1.000 & 1.000 & 1.000 \\
& Scenario 1 & 0.143 & 0.100 & 0.118 & 0.600 & 0.500 & 0.545 &  0.600 & 0.500 & 0.545 \\
& Scenario 2 & 0.931 & 0.900 & 0.915 & 0.600 & 0.500 & 0.545 & 0.600 & 0.500 & 0.545 \\
& Scenario 3 & 1.000 & 0.900 & 0.947 & 0.333 & 0.500 & 0.400 & 1.000 & 0.944 & 0.971 \\
\midrule
\multirow{4}{*}{\textbf{(b)~RE}} & Ground Truth & 1.000 & 1.000 & 1.000 & 1.000 & 1.000 & 1.000 & 1.000 & 1.000 & 1.000 \\
&Scenario 1 & 1.000 & 0.027 & 0.053 & 1.000 & 0.500 & 0.667 & 1.000 & 0.500 & 0.667 \\
&Scenario 2 & 1.000 & 0.973 & 0.986 & 1.000 & 0.500 & 0.667 & 1.000 & 0.500 & 0.667 \\
&Scenario 3 & 0.983 & 0.783 & 0.872 & 0.000 & 0.000 & 0.000 & 0.889 & 0.889 & 0.889 \\
\bottomrule
\end{tabular}
\end{table}}

\begin{align}
    \textit{tp}_p(l) &= \sum\limits_{C_p \in P_C(l)}{\frac{|C_p \cap G_M(l) |}{|C_p|}}, & 
    \textit{tp}_g(l) &= \sum\limits_{C_g \in G_C(l)}{\frac{|C_g \cap P_M(l) |}{|C_g|}} \label{eq:soft_m_tp_generic} \\ 
    \textit{fp}(l) &= |P_C(l)| - \textit{tp}_p(l), & 
    \textit{fn}(l) &= |G_C(l)| - \textit{tp}_g(l) \label{eq:soft_m_negatives_generic}
\end{align}

Our soft entity-level precision, recall and F$_1$ metrics are 
formally defined as follows,
where $L$ refers to either the number of all possible tags
for NER or the number of all possible relation types
for RE:

\begin{equation}
    \mathrm{Pr}_\mathrm{s} = \frac{\sum\limits_{l=1}^L\textit{tp}_p(l)}{\sum\limits_{l=1}^L\textit{tp}_p(l) + \textit{fp}(l)},\qquad 
    \mathrm{Re}_\mathrm{s} = \frac{\sum\limits_{l=1}^L\textit{tp}_g(l)}{\sum\limits_{l=1}^L\textit{tp}_g(l) + \textit{fn}(l)}, 
    \qquad
    \mathrm{F}_\mathrm{1,s} = \frac{2 \cdot  \mathrm{Pr}_\mathrm{s} \cdot \mathrm{Re}_\mathrm{s}}{\mathrm{Pr}_\mathrm{s} + \mathrm{Re}_\mathrm{s}} \label{eq:soft_metrics} 
\end{equation}

\begin{table}
    \centering
    \caption{The relations between the weighted true positives by the size of predicted ($\textit{tp}_p(l)$) and ground truth ($\textit{tp}_g(l)$) entity clusters allows us to achieve the constraints needed for the denominators of precision ($\textit{tp}_p(l) + \textit{fp}(l)$) and recall ($\textit{tp}_p(l) + \textit{fn}(l)$) functions (\equref{eq:soft_metrics}) in terms of the number of entity clusters.}
    \label{tab:metric_cluster_sum}

    \small
    \begin{tabular}{lm{6.5cm}m{6.5cm}}
    \toprule
    \textbf{Expression} &
    \multicolumn{1}{c}{\textbf{(a)~Meaning for NER}} &
    \multicolumn{1}{c}{\textbf{(b)~Meaning for RE
    }}
    \\
    \midrule
    $\textit{tp}_p(l) + \textit{fp}(l)$ & 
    Number of \emph{predicted} entity clusters with tag $l$. &
    Number of \emph{predicted} relations of type $l$ between entity clusters.
    \\
    $\textit{tp}_g(l) + \textit{fn}(l)$ &
    Number of \emph{ground truth} entity clusters with tag $l$. & 
    Number of \emph{ground truth} relations of type $l$ between entity clusters.
    \\    
    \bottomrule
    \end{tabular}
\end{table}

\begin{table}
    \centering
    
    \caption{Short definition of the symbols and expressions involved in our \textit{soft-entity level} metric formulation in  \eqsref{eq:soft_m_tp_generic}{eq:soft_metrics} for both NER and RE tasks.}
    \label{tab:metric_symbols}
    
    \small
    \begin{tabular}{lm{6.5cm}m{6.5cm}}
    \toprule
    \multicolumn{1}{c}{\textbf{Symbol}} &    
    \multicolumn{1}{c}{\textbf{(a)~Meaning for NER}} &
    \multicolumn{1}{c}{\textbf{(b)~Meaning for RE}} \\
    \midrule
    $P_C(l)$ & Set of predicted entity clusters with tag $l$. & Set of predicted relations of type $l$ between the predicted entity clusters. \\
    $C_p \in P_C(l)$ & Set of predicted entity mentions for a particular entity cluster in $P_C(l)$. & Set of relations between the predicted entity mentions for a particular pair of related entity clusters in $P_C(l)$. \\
    $G_C(l)$ & Set of ground truth entity clusters annotated with tag $l$. & Set of ground truth relations of type $l$ between the ground truth entity clusters. \\
    $C_g \in G_C(l)$ & Set of ground truth entity mentions for a particular entity cluster in $G_C(l)$. & Set of relations between the ground truth entity mentions for a particular pair of related entity clusters in $G_C(l)$ \\
    $P_M(l)$ & Set of predicted entity mentions with tag $l$. & Set of predicted relations of type $l$ between the predicted entity mentions. \\ 
    $G_M(l)$ & Set of ground truth entity mentions annotated with tag $l$. & Set of ground truth relations of type $l$ between the ground truth entity mentions. \\    
    \midrule 
    $\textit{tp}_p(l)$ & Number of true positive predictions of tag $l$ on mentions re-weighted by predicted cluster sizes. & Number of true positive predictions of relation type $l$ between mentions re-weighted by the number of mention level relations between the connected pairs of predicted clusters. \\
    $\textit{tp}_g(l)$ & Number of true positive mention level predictions of tag $l$ re-weighted by ground truth cluster sizes. & Number of true positive predictions of relation type $l$ between mentions re-weighted by the number of mention level relations between the connected pairs of ground truth clusters. \\
    $\textit{fp}(l)$ & Number of false positive mention level predictions of tag $l$ re-weighted by predicted cluster sizes. & Number of false positive predictions of relation type $l$ between mentions re-weighted by the number of mention level relations between the connected pairs of predicted clusters. \\
    $\textit{fn}(l)$ & Number of false negative mentions with ground truth tag $l$ re-weighted by ground truth cluster sizes. & Number of false negative relations of type $l$ between mentions re-weighted by the number of mention level relations between the connected pairs of ground truth clusters. \\
    \bottomrule
    \end{tabular}
\end{table}

\section{Experimental results}
\label{sec:results}
\subsection{Experimental Setup}
\label{sec:exp_setup}

We 
train and evaluate
our model 
as described in \secref{sec:models}
on three tasks: NER, coreference, and 
 relation extraction (RE)
independently and jointly. We experiment with three main model 
variations:
\begin{enumerate}
    \item \textbf{Single}: Experiments 
    on individual tasks by training with the respective loss functions as described in \secref{sec:single_task_models}. 
    \item \textbf{Joint}: Experiments 
    jointly on all three tasks using pre-trained \emph{GloVe representations}\footnote{\url{http://nlp.stanford.edu/data/glove.840B.300d.zip}} concatenated to character embeddings in the shared input layer (see \figref{fig:model_architecture}). For training we use the joint loss defined in  \secref{sec:joint_training}. 
    \item \textbf{Joint+BERT}: 
    as in the \emph{Joint} setting, experiments jointly on all three tasks, but using pre-trained BERT$_{\mathrm{BASE}}$ embeddings\footnote{\url{https://storage.googleapis.com/bert_models/2018_10_18/cased_L-12_H-768_A-12.zip}} 
    concatenated to the GloVe and character embeddings. 
    We use an input window size of 250 tokens and concatenate the last 2 hidden layers of BERT to get token representations. 
\end{enumerate}
Additionally, for each of the three model setups we experiment with the graph propagation techniques defined in \secref{sec:graph_prop_model}. 
To maximize result consistency, we train each model 5 times and report the average of these 5 results for each of the experiments. 

We use a single-layer BiLSTM with forward and backward hidden states of 200 dimensions each.
All our FFNNs used to obtain confidence scores ($\mathcal{F}_{\mathrm{pruner}}$, $\mathcal{F}_{\mathrm{coref}}$, $\mathbfcal{F}_{\mathrm{mention}}$, $\mathbfcal{F}_{\mathrm{relation}}$, and $\mathcal{F}_{\mathrm{att}}$) have two 150-dimensional hidden layers trained with a dropout of 0.4. We set the maximum span width ${w_{\mathrm{max}}}$ to 5 and the pruner ratio to 0.2 of the total number of tokens in a document. For training, we use Adam with a learning rate of 
\num{1e-3} for 100 epochs with a linear decay of 0.1 starting at epoch 15. 

{\renewcommand\baselinestretch{1}\begin{table}
	\centering
\caption[test]{Main results of the experiments grouped in three model setups: \begin{enumerate*}[(i)]
	\item \textit{Single} models trained individually,
	\item \textit{Joint} model trained using as input GloVe and character embeddings, and 
	\item \textit{Joint+BERT} model trained on BERT$_{\mathrm{BASE}}$ embeddings.
	\end{enumerate*} 
	 To report the results, we use MUC, CEAF$_\text{e}$, B$^\text{3}$ as well as the average (Avg.) of these three metrics for \textit{coreference resolution}. For NER and RE we use 
	 mention-level (F$_\text{1,m}$), hard entity-level (F$_\text{1,e}$), and soft entity-level (F$_\text{1,s}$) metrics described in \secref{sec:metrics}. In bold we mark the best results for each model setup, the best overall results are underlined. Note that the metrics are expressed in percentage points.}
	\label{tab:main_results}
	\setlength{\tabcolsep}{4pt}
	\renewcommand{\arraystretch}{1.0}
	\begin{tabular}{l cccc c ccc c ccc}
	\toprule
	&	\multicolumn{4}{c}{\textbf{Coreference} $\mathbf{F_1}$} && \multicolumn{3}{c}{\textbf{NER} $\mathbf{F_1}$} && \multicolumn{3}{c}{\textbf{RE} $\mathbf{F_1}$} \\ 
\cmidrule(lr){2-5} \cmidrule(lr){7-9} \cmidrule(lr){11-13}
	 \textbf{Model Setup} & 
	 \textbf{MUC} & 
	 \textbf{CEAF}$\mathbf{_{e}}$ & 
	 $\mathbf{B^3}$ & 
	 \textbf{Avg.} & 
	 \hspace{.5em} & 
	 $\mathbf{F_{1,m}}$ & 
	 $\mathbf{F_{1,e}}$ & 
	 $\mathbf{F_{1,s}}$ & 
	 \hspace{.5em} & 
	 $\mathbf{F_{1,m}}$ & 
	 $\mathbf{F_{1,e}}$ & 
	 $\mathbf{F_{1,s}}$ \\ 
	 \midrule 
	Single & 92.8 & 90.9 & 88.2 & 90.6 && 85.7 & - & - && 68.2 & - & - \\ 
 	\ \ \propformat{+AttProp} & \textbf{93.2} & \textbf{91.5} & \textbf{88.7} & \textbf{91.1} && \textbf{87.1} & - & - && \textbf{71.3} & - & - \\ 
 	\ \ \propformat{+CorefProp} & 92.8 & 90.9 & 88.3 & 90.7 && - & - & - && - & - & - \\ 
 	\ \ \propformat{+RelProp} & - & - & - & - && - & - & - && 68.2 & - & - \\ 
 	 \midrule
	Joint & 92.5 & \textbf{90.5} & \textbf{87.3} & \textbf{90.1} && 85.4 & 71.7 & 84.4 && 68.1 & 46.8 & 66.5 \\ 
 	\ \ \propformat{+AttProp} & 92.3 & 90.4 & 87.3 & 90.0 && 87.1 & 72.9 & \textbf{86.1} && \textbf{72.1} & \textbf{50.4} & \textbf{72.1} \\ 
 	\ \ \propformat{+CorefProp} & 92.3 & 90.3 & 87.2 & 89.9 && \textbf{87.2} & \textbf{73.2} & 86.0 && 71.6 & 50.2 & 71.0 \\ 
 	\ \ \propformat{+RelProp} & \textbf{92.6} & 90.2 & 86.8 & 89.9 && 86.7 & 72.4 & 85.2 && 69.5 & 48.2 & 68.8 \\ 
 	 \midrule
	Joint+BERT & \underline{\textbf{93.8}} & \underline{\textbf{92.1}} & \underline{\textbf{89.0}} & \underline{\textbf{91.6}} && 87.6 & 74.2 & 86.4 && 70.6 & 48.7 & 68.9 \\ 
 	\ \ \propformat{+AttProp} & 93.2 & 91.4 & 88.6 & 91.1 && \underline{\textbf{88.8}} & 74.2 & \underline{\textbf{87.7}} && 72.3 & \underline{\textbf{50.4}} & \underline{\textbf{73.0}} \\ 
 	\ \ \propformat{+CorefProp} & 93.5 & 91.8 & 88.7 & 91.3 && 88.7 & 74.4 & 87.4 && \underline{\textbf{72.7}} & 50.0 & 71.9 \\ 
 	\ \ \propformat{+RelProp} & 93.7 & 91.8 & 88.7 & 91.4 && 88.4 & \underline{\textbf{74.8}} & 87.0 && 72.0 & 49.9 & 71.4 \\ 
 	\bottomrule 
	\end{tabular}
\end{table}}

\subsection{Results and Analyses}
\label{sec:results_and_analysis}
\begin{figure}
    \centering
    \includegraphics[width=1.0\columnwidth]{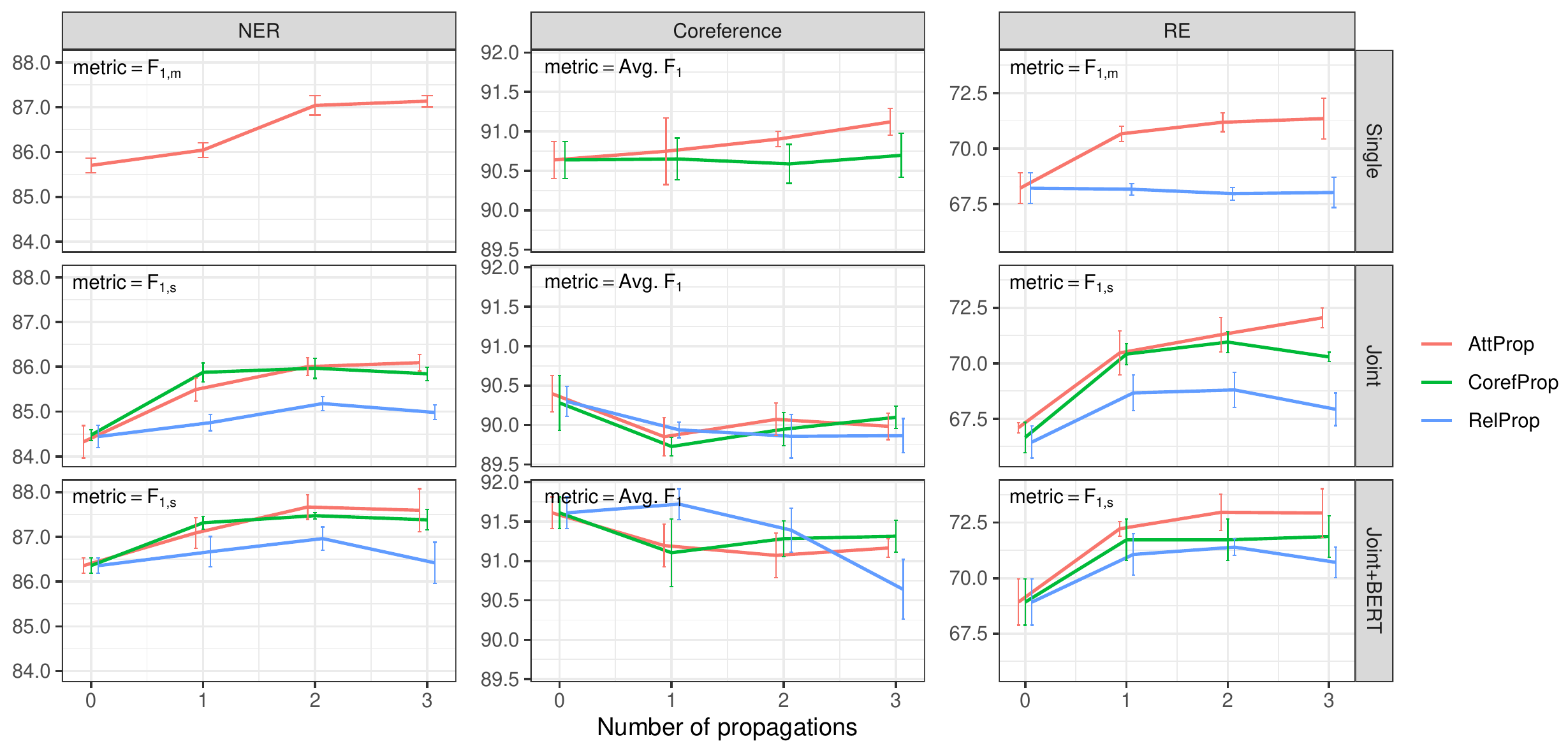}
    \caption{Impact of \propformat{AttProp}, \propformat{CorefProp} and \propformat{RelProp} graph propagations on performance metrics for each of the \emph{Single}, \emph{Joint} and \emph{Joint+BERT} model setups. Note the different Y-axis scales.}
    \label{fig:res_prop}
\end{figure}

 \Tabref{tab:main_results} gives an overview of the results achieved in \textit{Single} as well as \textit{Joint} and \textit{Joint + BERT} setups. Additionally, 
 \figref{fig:res_prop}
 illustrates the impact of the number of graph propagation iterations for each of the span graph propagation methods on the final results.
 
 First, we observe a general improvement in all our \emph{Single} tasks when using graph propagation techniques. More specifically, 
 our proposed latent 
 \propformat{AttProp}
 achieves superior results compared to the relation (\propformat{RelProp}) and coreference (\propformat{CorefProp}) propagations when added to the \textit{Single} setup. The biggest 
 improvement across iterations (see \figref{fig:res_prop}) is for the single 
 RE
 task 
 mention-level $\mathrm{F_{1,m}}$ score with a boost of ${\sim}3$ percentage points when incorporating \propformat{AttProp}. We also observe an improvement of ${\sim}1.5$ percentage points in $\mathrm{F_{1,m}}$ for the NER task and a consistent but smaller improvement of $0.5$ $\mathrm{F_1}$ percentage points for the coreference task. These results illustrate the effectiveness of \propformat{AttProp} when applied to single task models. 

A further improvement in results is achieved by training our model \emph{jointly} (see the {Joint} setup in \tabref{tab:main_results} and graphs in \figref{fig:res_prop}) for NER and 
RE
tasks. This illustrates that, besides the positive effect of neural graph propagation on single task models, training our model jointly has an additional benefit by exploiting the interaction between tasks. In particular, this effect can be seen for RE, where our \textit{Joint} model achieves a boost in performance of $0.8$ percentage points for the mention-level $\mathrm{F_{1,m}}$ metric compared to the best result for the \textit{Single} setup. Furthermore, our \propformat{AttProp} graph propagation method achieves the best performance on all the metrics for the RE task in the \textit{Joint} setting with up to $\sim{5.5}$ percentage points improvement in our newly proposed $\mathrm{F_{1,s}}$ metric. 
Additionally, we observe a beneficial effect of graph propagation for the
NER
task in the \textit{Joint} setup with slightly better results for the $\mathrm{F_{1,m}}$ metric compared to the \textit{Single} setting. Our \propformat{AttProp} technique performs on par with \propformat{CorefProp}, outperforming the latter by a small margin in terms of $\mathrm{F_{1,s}}$ metric. 

Similarly to the \textit{Joint} model variation, we observe benefits when using graph propagation techniques in the \textit{Joint+BERT} models.
\tabref{tab:analysis_deltas_props} illustrates the deltas in performance for the NER and 
relation extraction 
tasks. 
This way, we can see more clearly the difference in impact of our neural message passing methods grouped by the model setup and metric type.  
First, we observe that the general performance boost from using graph propagation techniques is lower in \textit{Joint+BERT} than in the \textit{Joint} setup. We hypothesize that this effect is due to the fact that BERT itself has a better long-range context extraction due to the attention-based mechanism, which spans the input window as opposed to purely local (non-contextualized) GloVe embeddings used in the \textit{Joint} setting. This is in line with the findings in \citet{han2020novel}, \citet{wadden2019entity}, and \citet{wu2019enriching} that show the advantage of using 
large BERT input window sizes 
to produce better IE results.
Second, we observe that our \propformat{AttProp} method achieves consistently superior performance on our proposed soft entity-level metric $\mathrm{F_{1,s}}$, capturing thus better the mention-based predictions as weighted by their cluster sizes. Finally, from \tabref{tab:analysis_deltas_props}(b) we notice that adding BERT to our joint model does not affect the boost in performance caused by the \propformat{RelProp} method for 
relation extraction. 
We hypothesize that this is due to the fact that \propformat{RelProp} propagation can capture relational semantics that 
goes beyond
BERT's contextual span representation similarity (which mainly drives the positive impact of \textit{Joint+BERT}). 

{\renewcommand\baselinestretch{1}\begin{table}
\centering
\caption{Deltas of improvement in performance for each of the graph propagation methods (\propformat{AttProp}, \propformat{CorefProp}, \propformat{RelProp}) in $\mathrm{F_1}$ scores for \textbf{(a)}~NER and \textbf{(b)}~relation extraction tasks.}
\label{tab:analysis_deltas_props}
\begin{tabular}{c l ccc c ccc}
\toprule
& & 
\multicolumn{3}{c}{\textbf{Joint}} & & 
\multicolumn{3}{c}{\textbf{Joint+BERT}} \\ 
\cmidrule(lr){3-5} \cmidrule(lr){7-9} 
& & 
$\mathbf{F_{1,m}}$ & 
$\mathbf{F_{1,e}}$ & 
$\mathbf{F_{1,s}}$ & 
&
$\mathbf{F_{1,m}}$ & 
$\mathbf{F_{1,e}}$ & 
$\mathbf{F_{1,s}}$ \\
\midrule 
\multirow{3}{*}{\textbf{(a)~NER}} &
$\Delta\mathrm{\ }$\propformat{AttProp} & 1.69 & 1.18 & \textbf{1.67} &  & \textbf{1.16} & $\mathrm{-0.02}$ & \textbf{1.31} \\ 
& $\Delta\mathrm{\ }$\propformat{CorefProp} & \textbf{1.78} & \textbf{1.50} & 1.54 &  & 1.05 & 0.20 & 1.02 \\ 
& $\Delta\mathrm{\ }$\propformat{RelProp} & 1.33 & 0.70 & 0.75 &  & 0.78 & \textbf{0.56} & 0.60 \\ 
\midrule 
\multirow{3}{*}{\textbf{(b)~RE}} &
$\Delta\mathrm{\ }$\propformat{AttProp} & \textbf{3.97} & \textbf{3.62} & \textbf{5.56} &  & 1.66 & \textbf{1.69} & \textbf{4.05} \\ 
& $\Delta\mathrm{\ }$\propformat{CorefProp} & 3.48 & 3.45 & 4.47 &  & \textbf{2.02} & 1.29 & 2.95 \\ 
& $\Delta\mathrm{\ }$\propformat{RelProp} & 1.35 & 1.47 & 2.32 &  & 1.37 & 1.20 & 2.48 \\ 
\bottomrule
\end{tabular}
\end{table}}

Unlike for the NER and 
RE
tasks, where we observe a consistent positive impact of span graph propagation and joint modeling across all our experiments, the impact on the \emph{coreference} task is not clear. Our experiments on \textit{Single} setup show small, but constant improvement of the Avg.-$\mathrm{F_1}$ score with the number of \propformat{AttProp} propagation iterations (see \figref{fig:res_prop}).
However, in our \textit{Joint} and \textit{Joint+BERT} setups the graph propagation appears to not have any positive impact on Avg.-$\mathrm{F_1}$ coreference scores.
We hypothesize that the main reason for this phenomenon lies in the coreference annotations in \datasetname: since we only annotate clusters of proper nouns, leaving out the nominal (\eg ``the prime minister") and anaphoric expressions (\eg ``he", ``she", ``they", etc), there might be little to no additional benefit in propagating 
information between co-referenced entity mentions, 
since the representation of proper nouns likely is not much influenced by textual context
(\eg the span ``Merkel" can have very similar span representation to ``Angela Merkel", gaining nothing in adding contextual graph propagation). 

Additionally, we explore in more detail the effect of the number of \propformat{AttProp}, \propformat{CorefProp}, \propformat{RelProp} graph propagation 
iterations on the final $\mathrm{F_1}$ score of all the tasks in 
\figref{fig:res_prop}.
We observe that the number of iterations have a decreasing effect on the improvement of performance for the NER and 
RE
tasks.
Furthermore, the positive effect of \propformat{CorefProp} and \propformat{RelProp} tends to saturate or even become negative after 1 or 2 iterations. This is in line with findings of \cite{luan2019general} on other datasets, where the performance peak is usually achieved at 2 graph propagation iterations.
For our \propformat{AttProp} however, we observe that the positive effect of additional iterations tends to persist longer, particularly on the \textit{Joint} setup where the positive effect of \propformat{AttProp} seems to be still growing after the last iteration~(3) in our experiments.

\section{Conclusions and Future Work}
\label{sec:conclusion}

In this work we introduced \datasetname, a manually annotated multi-task dataset that comprises Named Entity Recognition, Coreference, Relation Extraction and Entity Linking as main tasks. We highlight how \datasetname~is different from the mainstream datasets by focusing on document-level and entity-centric annotations. This also makes the predictions on this dataset more challenging by having not only to consider explicit, but also implicit document-level interactions between entities. 
Furthermore, we showed how Graph Neural Networks can help to tackle this issue by propagating local contextual mention span information on a document level for a single task as well as across the tasks on the \datasetname~dataset. 
We experiment with known graph propagation techniques driven by the scores of the coreference resolution (\propformat{CorefProp}) and relation extraction (\propformat{RelProp}) components, as well as introduced a new latent task-independent attention-based graph propagation method (\propformat{AttProp}). We demonstrated that, without relying on the task-specific scorers, \propformat{AttProp} can boost the performance of single-task as well as joint models, performing on par and even outperforming significantly in some scenarios the \propformat{RelProp} and \propformat{CorefProp} graph propagations.  

In future work we will aim to integrate an entity linking component into our joint architecture. As a consequence, we expect to obtain a further boost in performance of different tasks included in \datasetname~by taking advantage of the information coming from Wikipedia 2018, the reference knowledge base for the entity linking annotations. Conversely, we conjecture that the results of the entity linking component can be improved when training it jointly with other tasks, such as NER and coreference resolution. Finally, we plan extending the coreference annotations to include nominal and anaphoric expressions. We expect that including these diverse mention types, whose initial span embedding representation can be different from coreferenced named entities, will make our coreference resolution task more challenging, allowing to investigate further the potential benefits of using graph-based neural networks. 

\section*{Acknowledgements}
\noindent Part of the research leading to these results has received funding from
\begin{enumerate*}[(i)]
\item the European Union's Horizon 2020 research and innovation programme under grant agreement no.\ 761488 for the CPN project,\footnote{\url{https://www.projectcpn.eu/}} and
\item the Flemish Government under the ``Onderzoeksprogramma Artifici\"{e}le Intelligentie (AI) Vlaanderen'' programme.
\end{enumerate*}

\appendix
\renewcommand{\thetable}{\Alph{section}.\arabic{table}}
\renewcommand{\thefigure}{\Alph{section}.\arabic{figure}}

\renewcommand\baselinestretch{1}\small
\section{Dataset Insights}
\setcounter{table}{0}
\setcounter{figure}{0}
\setlength{\intextsep}{1pt plus 1pt minus 1pt}
\label{app:dataset_insights}
\begin{table}[H]
\centering
\caption{Statistics depicting the hierarchical structure of entity types described in \secref{sec:schema_driven_pass}. Only the most frequent entity types/subtypes are shown (\% Mentions $>$ 0.5\%)}
\resizebox{0.8\textwidth}{!}{
\begin{tabular}{lllll} 
\toprule
\textbf{Entity Type} & \textbf{\# Entities} & \textbf{\% Entities} & \textbf{\# Mentions} & \textbf{\% Mentions} \\ 
\midrule
	\textit{\hspace{0.00cm}ENTITY} & \textit{13,151} & \textit{56.9}\% & \textit{30,719} & \textit{70.8}\% \\ 
	\hspace{0.30cm}location & 4,957 & 21.4\% & 11,548 & 26.6\% \\ 
	\hspace{0.60cm}gpe & 3,965 & 17.1\% & 9,830 & 22.7\% \\ 
	\hspace{0.90cm}gpe0 & 2,225 & 9.6\% & 6,559 & 15.1\% \\ 
	\hspace{0.90cm}gpe2 & 1,497 & 6.5\% & 2,873 & 6.6\% \\ 
	\hspace{0.90cm}gpe1 & 244 & 1.1\% & 406 & 0.9\% \\ 
	\hspace{0.60cm}regio & 479 & 2.1\% & 916 & 2.1\% \\ 
	\hspace{0.60cm}facility & 259 & 1.1\% & 385 & 0.9\% \\ 
	\hspace{0.30cm}organization & 3,434 & 14.8\% & 8,165 & 18.8\% \\ 
	\hspace{0.60cm}media & 659 & 2.8\% & 984 & 2.3\% \\ 
	\hspace{0.60cm}igo & 547 & 2.4\% & 1,992 & 4.6\% \\ 
	\hspace{0.90cm}so & 171 & 0.7\% & 912 & 2.1\% \\ 
	\hspace{0.60cm}party & 381 & 1.6\% & 949 & 2.2\% \\ 
	\hspace{0.60cm}company & 368 & 1.6\% & 932 & 2.1\% \\ 
	\hspace{0.60cm}sport\_team & 367 & 1.6\% & 1,106 & 2.5\% \\ 
	\hspace{0.60cm}governmental\_organization & 342 & 1.5\% & 636 & 1.5\% \\ 
	\hspace{0.90cm}agency & 228 & 1.0\% & 444 & 1.0\% \\ 
	\hspace{0.60cm}armed\_movement & 108 & 0.5\% & 374 & 0.9\% \\ 
	\hspace{0.30cm}person & 3,390 & 14.7\% & 8,259 & 19.0\% \\ 
	\hspace{0.60cm}politician & 1,184 & 5.1\% & 3,326 & 7.7\% \\ 
	\hspace{0.90cm}head\_of\_state & 380 & 1.6\% & 1,271 & 2.9\% \\ 
	\hspace{0.90cm}head\_of\_gov & 247 & 1.1\% & 673 & 1.6\% \\ 
	\hspace{0.90cm}minister & 217 & 0.9\% & 458 & 1.1\% \\ 
	\hspace{0.60cm}sport\_player & 405 & 1.8\% & 844 & 1.9\% \\ 
	\hspace{0.60cm}artist & 260 & 1.1\% & 586 & 1.4\% \\ 
	\hspace{0.60cm}politics\_per & 209 & 0.9\% & 457 & 1.1\% \\ 
	\hspace{0.60cm}manager & 104 & 0.4\% & 297 & 0.7\% \\ 
	\hspace{0.60cm}offender & 75 & 0.3\% & 347 & 0.8\% \\ 
	\hspace{0.30cm}misc & 823 & 3.6\% & 1,646 & 3.8\% \\ 
	\hspace{0.60cm}work\_of\_art & 174 & 0.8\% & 247 & 0.6\% \\ 
	\hspace{0.30cm}event & 354 & 1.5\% & 701 & 1.6\% \\ 
	\hspace{0.60cm}sport\_competition & 183 & 0.8\% & 410 & 0.9\% \\ 
	\hspace{0.30cm}ethnicity & 84 & 0.4\% & 242 & 0.6\% \\ 
	\textit{\hspace{0.00cm}VALUE} & \textit{5,903} & \textit{25.5}\% & \textit{7,104} & \textit{16.4}\% \\ 
	\hspace{0.30cm}time & 2,907 & 12.6\% & 3,608 & 8.3\% \\ 
	\hspace{0.30cm}role & 2,390 & 10.3\% & 2,865 & 6.6\% \\ 
	\hspace{0.30cm}money & 606 & 2.6\% & 631 & 1.5\% \\ 
	\textit{\hspace{0.00cm}OTHER} & \textit{2,724} & \textit{11.8}\% & \textit{5,482} & \textit{12.6}\% \\ 
	\hspace{0.30cm}gpe0-x & 1,596 & 6.9\% & 3,827 & 8.8\% \\ 
	\hspace{0.30cm}footer & 413 & 1.8\% & 413 & 1.0\% \\ 
	\hspace{0.30cm}loc-x & 353 & 1.5\% & 585 & 1.3\% \\ 
	\hspace{0.30cm}religion-x & 235 & 1.0\% & 486 & 1.1\% \\ 
\midrule
\textbf{TOTAL} & 23,130 & 100.0\% & 43,373 & 100.0\% \\ 
\bottomrule
\end{tabular}
}
\label{tab:entities_stats}
\end{table}

\newgeometry{margin=1.5cm}
\begin{landscape}

\begin{table}[h]
    \centering
\caption{Illustration of NER entity types in \datasetname. Each cells contains possible entity subtypes (of different hierarchy levels) corresponding to the respective parent entity type (column) and topic (row).}
\label{tab:ner_labels_matrix}
\footnotesize
    \begin{tabular}{l>{\raggedright\arraybackslash}p{5.4cm}>{\raggedright\arraybackslash}p{5.4cm}>{\raggedright\arraybackslash}p{2.4cm}>{\raggedright\arraybackslash}p{2.4cm}>{\raggedright\arraybackslash}p{7.4cm}}
\toprule
         & \multicolumn{1}{c}{\entityname{person}}  & \multicolumn{1}{c}{\entityname{organization}}  & \multicolumn{1}{c}{\entityname{event}}  & \multicolumn{1}{c}{\entityname{location}}  & \multicolumn{1}{c}{\entityname{misc}} \\
         \midrule
         \topicname{politics} & head\_of\_gov, head\_of\_state, minister, politician\_regional, politician\_local, politician\_national, candidate, politician, politics\_per, activist, gov\_per & politics\_institution, politics\_org, party, ngo, igo, so, policy\_institute, movement, agency, ministry, military\_alliance & summit\_meeting, scandal, politics\_event & politics\_facility & politics\_misc, project, treaty, report \\ 
         \midrule
         \topicname{culture} & character, culture\_per, artist, writer, actor, filmmaker, musician, photographer & music\_band, culture\_org, theatre\_org, dance\_org & festival, film\_festival & culture\_facility & art\_title, culture\_title, exhibition\_title, culture\_misc, work\_of\_art, book\_title, film\_title, tv\_title, music\_title, theatre\_title, musical\_title, film\_award, book\_award, music\_award, tv\_award, column\_title, game, comic, radio\_title, dance\_title, opera \\ 
         \midrule
         \topicname{education} & teacher, education\_per, education\_student & education\_org &  & education\_facility & education\_study \\ 
         \midrule
         \topicname{religion} & deity, clergy & religion\_org & religious\_event & religion\_facility & religion, religion\_misc \\ 
         \midrule
         \topicname{human} & royalty &  &  &  & film\_award, book\_award, award, music\_award, tv\_award, sport\_award \\ 
         \hline 
\topicname{conflict} & military\_personnel, military\_rebel & army, military\_alliance, armed\_movement & war, protest & military\_facility & military\_equipment, military\_mission \\ 
         \midrule 
\topicname{media} & journalist & media &  &  &  \\ 
         \midrule 
\topicname{science} & researcher, science\_per & research\_center &  &  & species, research\_journal, technology \\ 
         \midrule 
\topicname{sport} & sport\_player, sport\_coach, sport\_head, sport\_referee, sport\_person & sport\_team, sport\_org & sport\_competition & sport\_facility & sport\_award \\ 
         \midrule 
\topicname{labor} & union\_head, union\_member, union\_rep, union\_per & union &  &  &  \\ 
         \midrule 
\topicname{business} & manager, employee, business\_per & company, business\_org, brand, trade\_fair, market\_exchange, advocacy &  & business\_facility & product, market\_index, business\_misc \\ 
         \midrule 
\topicname{health} & health\_per & health\_org &  & health\_facility & health\_disease, health\_drug \\ 
         \midrule 
\topicname{justice} & offender, advisor, victim, judge, police\_per, justice\_per & court, criminal\_org, police\_org, justice\_org &  & prison & justice\_misc, case \\ 
         \midrule 
\topicname{weather} &  &  & storm &  & \\ 
         \bottomrule
       \end{tabular}
\end{table}
\end{landscape}
\restoregeometry
\clearpage

\renewcommand\baselinestretch{1}\small \Tabref{tab:entity_linking_stats} describes the statistics of linked entities with respect to the total number of entities in each of the \textit{Entity} subtypes. The columns \textit{\% Linked Entities} and \textit{\% Linked Mentions} indicate the percentage of annotated linked entities and mentions with respect to the total number of annotated entities/mentions in a particular \textit{Entity} type category. Furthermore, we calculate two accuracies on test split when linking the entity mention with the most frequent entity link used either in \datasetname: \begin{enumerate*}[(i)]
    \item training set of \datasetname~dataset (``Acc. Prior Train''), or
    \item Wikipedia corpus (``Acc. Prior Wiki'')
\end{enumerate*}. 
Overall, using prior linking annotations from Wikipedia gives 9 percentage points better performance (79.0\%) than when using train set (70.0\%). This difference is explained by the fact that Wikipedia has much larger corpus to calculate the prior linking information from. Nevertheless, we still observe that for some entity types such as \textit{sport\_team} and  \textit{media} the accuracy based on \datasetname~training set prior is higher. This suggests the use of domain-specific language to refer to some entities in \datasetname~not used in a more general Wikipedia domain.
\setlength{\intextsep}{12.0pt plus 2.0pt minus 2.0pt}
\begin{table}[!htb]
\centering
\caption{Entity linking statistics, only the top 5 types and subtypes with largest number of linked entities are showed. The \textit{total} is calculated on all the entity types. The accuracy (both for most likely prior links on train and Wiki corpora) is computed on test set. }
\resizebox{0.9\textwidth}{!}{\begin{tabular}{lllllll} 
\toprule
\begin{tabular}{@{}c@{}}\textbf{\ } \\ \textbf{Entity Type} \end{tabular}& \begin{tabular}{@{}l@{}} \textbf{\# Linked} \\ \textbf{Entities} \end{tabular} & \begin{tabular}{@{}l@{}} \textbf{\% Linked} \\ \textbf{Entities} \end{tabular} & \begin{tabular}{@{}l@{}} \textbf{\# Linked} \\  \textbf{Mentions} \end{tabular} & \begin{tabular}{@{}l@{}} \textbf{\% Linked} \\  \textbf{Mentions} \end{tabular} & \begin{tabular}{@{}l@{}} \textbf{Acc. Prior} \\ \textbf{Train} \end{tabular} & \begin{tabular}{@{}l@{}} \textbf{Acc. Prior} \\ \textbf{Wiki} \end{tabular} \\ 
\midrule
	\textit{\hspace{0.00cm}LOCATION} & \textit{4,863} & \textit{98.1}\% & \textit{11,496} & \textit{99.5}\% & \textit{85.7\%} & \textit{92.9\%} \\ 
	\hspace{0.30cm}gpe & 3,938 & 99.3\% & 9,810 & 99.8\% & 89.8\% & 95.6\% \\ 
	\hspace{0.30cm}regio & 456 & 95.2\% & 889 & 97.1\% & 83.3\% & 76.3\% \\ 
	\hspace{0.30cm}facility & 229 & 88.4\% & 381 & 99.0\% & 19.7\% & 73.8\% \\ 
	\hspace{0.30cm}waterbody & 90 & 98.9\% & 145 & 100.0\% & 83.3\% & 91.7\% \\ 
	\hspace{0.30cm}district & 37 & 94.9\% & 45 & 100.0\% & 33.3\% & 33.3\% \\ 
	\textit{\hspace{0.00cm}ORGANIZATION} & \textit{3,145} & \textit{91.6}\% & \textit{8,029} & \textit{98.3}\% & \textit{69.8\%} & \textit{70.8\%} \\ 
	\hspace{0.30cm}media & 622 & 94.4\% & 979 & 99.5\% & 81.8\% & 59.5\% \\ 
	\hspace{0.30cm}igo & 525 & 96.0\% & 1,952 & 98.0\% & 76.4\% & 78.8\% \\ 
	\hspace{0.30cm}party & 358 & 94.0\% & 897 & 94.5\% & 77.5\% & 66.7\% \\ 
	\hspace{0.30cm}company & 320 & 87.0\% & 923 & 99.0\% & 67.6\% & 89.7\% \\ 
	\hspace{0.30cm}sport\_team & 366 & 99.7\% & 1,105 & 99.9\% & 71.0\% & 47.5\% \\ 
	\textit{\hspace{0.00cm}PERSON} & \textit{2,627} & \textit{77.5}\% & \textit{8,217} & \textit{99.5}\% & \textit{45.7\%} & \textit{69.4\%} \\ 
	\hspace{0.30cm}politician & 1,162 & 98.1\% & 3,324 & 99.9\% & 66.0\% & 78.1\% \\ 
	\hspace{0.30cm}sport\_player & 404 & 99.8\% & 843 & 99.9\% & 34.4\% & 71.3\% \\ 
	\hspace{0.30cm}artist & 246 & 94.6\% & 567 & 96.8\% & 0.0\% & 29.4\% \\ 
	\hspace{0.30cm}politics\_per & 126 & 60.3\% & 456 & 99.8\% & 23.7\% & 42.1\% \\ 
	\hspace{0.30cm}manager & 58 & 55.8\% & 296 & 99.7\% & 22.2\% & 33.3\% \\ 

	\textit{\hspace{0.00cm}MISC} & \textit{607} & \textit{73.8}\% & \textit{1,532} & \textit{93.1}\% & \textit{58.4\%} & \textit{73.4\%} \\ 
	\hspace{0.30cm}work\_of\_art & 142 & 81.6\% & 246 & 99.6\% & 0.0\% & 100.0\% \\ 
	\hspace{0.30cm}award & 72 & 80.0\% & 186 & 94.9\% & 63.6\% & 81.8\% \\ 
	\hspace{0.30cm}treaty & 60 & 74.1\% & 149 & 99.3\% & 66.7\% & 50.0\% \\ 
	\hspace{0.30cm}product & 50 & 76.9\% & 146 & 98.6\% & 52.0\% & 92.0\% \\ 
	\hspace{0.30cm}species & 10 & 25.0\% & 14 & 18.4\% & 0.0\% & 100.0\% \\ 
	\textit{\hspace{0.00cm}EVENT} & \textit{320} & \textit{90.4}\% & \textit{683} & \textit{97.4}\% & \textit{49.4\%} & \textit{67.1\%} \\ 
	\hspace{0.30cm}sport\_competition & 163 & 89.1\% & 397 & 96.8\% & 64.6\% & 87.5\% \\ 
	\hspace{0.30cm}summit\_meeting & 15 & 68.2\% & 37 & 92.5\% & 100.0\% & 100.0\% \\ 
	\hspace{0.30cm}holiday & 21 & 95.5\% & 39 & 97.5\% & 100.0\% & 100.0\% \\ 
	\hspace{0.30cm}history & 17 & 89.5\% & 30 & 100.0\% & 100.0\% & 100.0\% \\ 
	\hspace{0.30cm}protest & 14 & 100.0\% & 22 & 100.0\% & 80.0\% & 100.0\% \\ 
\midrule
\textbf{TOTAL} & \textbf{13,086} & \textbf{56.6\%} & \textbf{28,482} & \textbf{65.7\%} & \textbf{70.0\%} & \textbf{79.0\%} \\ 
\bottomrule
\end{tabular}}
\label{tab:entity_linking_stats}
\end{table}

\clearpage

\begin{table}[t]
\centering
\caption{Main named entity tag categories with statistics of the number and \% of covered entities and mentions as well as the number of classes in each and average number of labels per entity cluster.}
\resizebox{0.9\textwidth}{!}{
\begin{tabular}{llllllll}
\toprule
\begin{tabular}{@{}l@{}}\textbf{Entity Tag} \\ \textbf{Category}\end{tabular} & \textbf{\# Entities} & \textbf{\% Entities} & \textbf{\# Mentions} & \textbf{\% Mentions} & \textbf{\# Classes} & \begin{tabular}{@{}l@{}}\textbf{Labels per} \\ \textbf{Entity}\end{tabular} \\\midrule
	type & 21,745 & 94.0\% & 43,122 & 99.4\% & 174 & 2.9 \\ 
	topic & 7,843 & 33.9\% & 18,359 & 42.3\% & 14 & 1.0 \\ 
	iptc & 7,059 & 30.5\% & 17,195 & 39.6\% & 114 & 1.3 \\ 
	gender & 3,352 & 14.5\% & 8,200 & 18.9\% & 2 & 1.0 \\ 
	slot & 3,232 & 14.0\% & 14,983 & 34.5\% & 7 & 1.2 \\ 
\midrule
	\textbf{TOTAL} & 23,130 & 100.0\% & 43,373 & 100.0\% & 311 & 4.0 \\ 
\bottomrule
\end{tabular}
}
\label{tab:main_entity_types}
\end{table}

\setlength{\tabcolsep}{5pt}
\Tabref{tab:main_entity_types} illustrates the number of annotated entities and mentions per each tag category (type, topic, iptc, gender and slot). It also showcases the multi-label nature of entity classification task in \datasetname, with the average number of labels per entity of 4.0. 

\Tabref{tab:relations_multilabel_stats} illustrates the number and percentage of related entities and mentions of our dataset grouped by the number of relation labels. 
It also compares with other entity-centric 
RE
datasets, namely BC5CDR~\citep{li2016biocreative,wei2015overview} and DocRED~\citep{yao2019docred} datasets. 
\setlength{\tabcolsep}{5pt}
\renewcommand{\arraystretch}{1.0}
\begin{table}[h]
\centering
\caption{This table groups the number of related pairs in \datasetname~by the number of assigned relation labels to each of these pairs. We compare with other two entity-centric datasets: BC5CDR and DocRED.}
\resizebox{0.90\textwidth}{!}{
\begin{tabular}{lllllll} 
\toprule
 & \multicolumn{4}{c}{\textbf{\datasetname}} & \multicolumn{1}{c}{\textbf{BC5CDR}} & \multicolumn{1}{c}{\textbf{DocRED}} \\
\cmidrule(lr){2-5} \cmidrule(lr){6-6} \cmidrule(lr){7-7}

\textbf{\# Relation} & \textbf{\# Related} & \textbf{\% Related} & \textbf{\# Related} & \textbf{\% Related} & \textbf{\% Related} & \textbf{\% Related} \\
\textbf{labels} & \textbf{ent.\ pairs} & \textbf{ent.\ pairs} & \textbf{mention pairs} & \textbf{mention pairs} & \textbf{ent.\ pairs} & \textbf{ent.\ pairs} \\
\midrule
	1 & 12,856 & 76.32\% & 112,708 & 69.40\% & 100\% & 92.89\%\\ 
	2 & 3,101 & 18.41\% & 34,948 & 21.52\% & 0\% & 6.82\% \\ 
	3 & 884 & 5.25\% & 14,650 & 9.02\% & 0\% & 0.26\% \\ 
	4 & 3 & 0.02\% & 100 & 0.06\% & 0\% & 0.03\% \\ 
\midrule
	\textbf{TOTAL} & 16,844 & 100.0\% & 162,406 & 100.0\% & 100.0\% & 100.0\% \\ 
\bottomrule
\end{tabular}
}
\label{tab:relations_multilabel_stats}
\end{table}

\setlength{\tabcolsep}{5pt}
\renewcommand{\arraystretch}{1.0}

\begin{table}[!htb]
\centering
\caption{Relation type statistics. We compare the number of related entity and mention pairs per relation type. Only the most frequent relation types are shown (\% Related Mention Pairs $>$ \num{0.1}\%)}
\footnotesize
\resizebox{0.7\textwidth}{!}{
\begin{tabular}{lllll} 
\toprule
\begin{tabular}{@{}l@{}}\textbf{Relation} \\ \textbf{Type}\end{tabular} & \begin{tabular}{@{}l@{}}\textbf{\# Related} \\ \textbf{Ent. Pairs}\end{tabular}& \begin{tabular}{@{}l@{}}\textbf{\% Related} \\ \textbf{Ent. Pairs}\end{tabular}& \begin{tabular}{@{}l@{}}\textbf{\# Related} \\ \textbf{Men. Pairs}\end{tabular}& \begin{tabular}{@{}l@{}}\textbf{\% Related} \\ \textbf{Men. Pairs}\end{tabular} \\\midrule
	based\_in0 & 2,361 & 14.0\% & 18,771 & 11.6\% \\ 
	in0 & 2,120 & 12.6\% & 15,810 & 9.7\% \\ 
	citizen\_of & 1,969 & 11.7\% & 25,752 & 15.9\% \\ 
	based\_in0-x & 1,882 & 11.2\% & 12,211 & 7.5\% \\ 
	citizen\_of-x & 1,844 & 10.9\% & 17,049 & 10.5\% \\ 
	member\_of & 1,616 & 9.6\% & 19,953 & 12.3\% \\ 
	gpe0 & 1,569 & 9.3\% & 18,110 & 11.2\% \\ 
	in0-x & 1,474 & 8.8\% & 8,784 & 5.4\% \\ 
	agent\_of & 954 & 5.7\% & 15,776 & 9.7\% \\ 
	head\_of & 564 & 3.3\% & 7,710 & 4.7\% \\ 
	agency\_of & 435 & 2.6\% & 4,775 & 2.9\% \\ 
	player\_of & 401 & 2.4\% & 5,692 & 3.5\% \\ 
	agency\_of-x & 382 & 2.3\% & 2,108 & 1.3\% \\ 
	head\_of\_state & 380 & 2.3\% & 7,986 & 4.9\% \\ 
	head\_of\_state-x & 343 & 2.0\% & 3,853 & 2.4\% \\ 
	appears\_in & 294 & 1.7\% & 4,555 & 2.8\% \\ 
	vs & 281 & 1.7\% & 7,187 & 4.4\% \\ 
	head\_of\_gov & 273 & 1.6\% & 4,015 & 2.5\% \\ 
	head\_of\_gov-x & 247 & 1.5\% & 2,383 & 1.5\% \\ 
	minister\_of & 234 & 1.4\% & 2,280 & 1.4\% \\ 
	minister\_of-x & 213 & 1.3\% & 1,629 & 1.0\% \\ 
	based\_in2 & 185 & 1.1\% & 971 & 0.6\% \\ 
	event\_in0 & 181 & 1.1\% & 843 & 0.5\% \\ 
	part\_of & 164 & 1.0\% & 2,858 & 1.8\% \\ 
	in2 & 157 & 0.9\% & 1,055 & 0.6\% \\ 
	created\_by & 134 & 0.8\% & 945 & 0.6\% \\ 
	agent\_of-x & 125 & 0.7\% & 897 & 0.6\% \\ 
	award\_received & 111 & 0.7\% & 969 & 0.6\% \\ 
	institution\_of & 105 & 0.6\% & 2,113 & 1.3\% \\ 
	ministry\_of & 81 & 0.5\% & 666 & 0.4\% \\ 
	coach\_of & 65 & 0.4\% & 1,211 & 0.7\% \\ 
	won\_vs & 61 & 0.4\% & 1,531 & 0.9\% \\ 
	spouse\_of & 55 & 0.3\% & 599 & 0.4\% \\ 
	directed\_by & 44 & 0.3\% & 318 & 0.2\% \\ 
	is\_meeting & 41 & 0.2\% & 968 & 0.6\% \\ 
	event\_in2 & 40 & 0.2\% & 259 & 0.2\% \\ 
	spokesperson\_of & 39 & 0.2\% & 177 & 0.1\% \\ 
	plays\_in & 38 & 0.2\% & 330 & 0.2\% \\ 
	gpe1 & 35 & 0.2\% & 135 & 0.1\% \\ 
	product\_of & 31 & 0.2\% & 334 & 0.2\% \\ 
	parent\_of & 22 & 0.1\% & 281 & 0.2\% \\ 
	child\_of & 22 & 0.1\% & 281 & 0.2\% \\ 
	based\_in1 & 22 & 0.1\% & 376 & 0.2\% \\ 
	signed\_by & 20 & 0.1\% & 521 & 0.3\% \\ 
	law\_of & 16 & 0.1\% & 286 & 0.2\% \\
\midrule
	\textbf{TOTAL} & 16,844 & 100.0\% & 162,406 & 100.0\% \\ 
\bottomrule
\end{tabular}
}
\label{tab:relations_stats}
\end{table}
\clearpage

\section{Inter-annotator agreement calculations}
\label{app:kappa_agreement_details}
 In order to measure the agreement we use Cohen's kappa coefficient \citep{cohen1960coefficient}, defined as
\begin{align}
\kappa = \dfrac{p_o - p_e}{1 - p_e} \label{eq:kappa_generic} 
\end{align}
where $p_o$ represents the observed agreement between the two annotators and $p_e$ is the expected agreement between the annotators (\ie agreement by chance). 
More specifically, in our case we calculate the observed probability $p_o$ as in \equref{eq:kappa_observed} where $N$ is the number of annotated items, $A_{i,j}$ is the annotation made by annotator $i$ for item $j$, and 
$\mathbbm{1}\{ A_{1,j} = A_{2,j} \}$
returns $1$ if $A_{1,j}$ is equal to $A_{2,j}$ and $0$ otherwise. Thus, $p_o$ can be interpreted as the fraction of the labels two annotators agree, also called \textit{percent agreement} \citep{mchugh2012interrater,scott1955reliability}.
\begin{align}
p_o = \dfrac{\sum\limits_{j=1}^N{\mathbbm{1}\{ A_{1,j} = A_{2,j} \}}}{N} \label{eq:kappa_observed} 
\end{align}
To calculate the expected agreement probability we use the formulation in \equref{eq:kappa_expected}. It can be interpreted as
the probability that both annotators, when randomly distributing all of their label annotations among the items to be annotated, assign the same label to a given item.
\begin{align}
p_e = \sum\limits_{l=1}^L{\frac{n_{1,l}}{N} \> \frac{n_{2,l}}{N}} 
\label{eq:kappa_expected} 
\end{align}
In this context, $n_{i,l}$ is the number of items the annotator $i$ annotated with label $l$ and $L$ is the total number of labels. For multi-label annotations where it is possible to assign multiple classes for a particular annotation item (\ie named entity and relation types), we report a weighted kappa score. 
\clearpage
\section{Relation consistency rules}
\label{app:rel_rules_list}
\setcounter{table}{0}
\setcounter{figure}{0}
This appendix enumerates the logical predicates used as a consistency check in our dataset. 
\allowdisplaybreaks
\setcounter{equation}{0}
\begin{align}
	\relationalign{spouse\_of}{Y}{X} \implies \relationalign{spouse\_of}{X}{Y} \label{eq:rel_cons1} \\ 
	\relationalign{vs}{Y}{X} \implies \relationalign{vs}{X}{Y} \label{eq:rel_cons2} \\ 
	\relationalign{won\_vs}{X}{Y} \implies \relationalign{vs}{X}{Y} \label{eq:rel_cons3} \\ 
 	\relationalign{won\_vs}{X}{Y} \implies \relationalign{vs}{Y}{X} \label{eq:rel_cons4} \\ 
 	\relationalign{child\_of}{Y}{X} \implies \relationalign{parent\_of}{X}{Y} \label{eq:rel_cons5} \\ 
 	\relationalign{parent\_of}{Y}{X} \implies \relationalign{child\_of}{X}{Y} \label{eq:rel_cons6} \\ 
 	\relationalign{ministry\_of}{X}{Y} \implies \relationalign{agency\_of}{X}{Y} \label{eq:rel_cons7} \\ 
 	\relationalign{agency\_of\text{-}x}{X}{Z} \land \relationalign{gpe0}{Z}{Y} \implies \relationalign{agency\_of}{X}{Y} \label{eq:rel_cons8} \\ 
 	\relationalign{agency\_of}{X}{Y} \land \relationalign{gpe0}{Z}{Y} \implies \relationalign{agency\_of\text{-}x}{X}{Z} \label{eq:rel_cons9} \\ 
 	\relationalign{agent\_of\text{-}x}{X}{Z} \land \relationalign{gpe0}{Z}{Y} \implies \relationalign{agent\_of}{X}{Y} \label{eq:rel_cons10} \\ 
 	\relationalign{agent\_of}{X}{Y} \land \relationalign{gpe0}{Z}{Y} \implies \relationalign{agent\_of\text{-}x}{X}{Z} \label{eq:rel_cons11} \\ 
 	\relationalign{minister\_of}{X}{Y} \implies \relationalign{agent\_of}{X}{Y} \label{eq:rel_cons12} \\ 
 	\relationalign{head\_of\_gov}{X}{Y} \implies \relationalign{agent\_of}{X}{Y} \label{eq:rel_cons13} \\ 
 	\relationalign{head\_of\_state}{X}{Y} \implies \relationalign{agent\_of}{X}{Y} \label{eq:rel_cons14} \\ 
 	\relationalign{citizen\_of\text{-}x}{X}{Z} \land \relationalign{gpe0}{Z}{Y} \implies \relationalign{citizen\_of}{X}{Y} \label{eq:rel_cons15} \\ 
 	\relationalign{citizen\_of}{X}{Y} \land \relationalign{gpe0}{Z}{Y} \implies \relationalign{citizen\_of\text{-}x}{X}{Z} \label{eq:rel_cons16} \\ 
 	\relationalign{minister\_of\text{-}x}{X}{Z} \land \relationalign{gpe0}{Z}{Y} \implies \relationalign{minister\_of}{X}{Y} \label{eq:rel_cons17} \\ 
 	\relationalign{minister\_of}{X}{Y} \land \relationalign{gpe0}{Z}{Y} \implies \relationalign{minister\_of\text{-}x}{X}{Z} \label{eq:rel_cons18} \\ 
 	\relationalign{head\_of\_state\text{-}x}{X}{Z} \land \relationalign{gpe0}{Z}{Y} \implies \relationalign{head\_of\_state}{X}{Y} \label{eq:rel_cons19} \\ 
 	\relationalign{head\_of\_state}{X}{Y} \land \relationalign{gpe0}{Z}{Y} \implies \relationalign{head\_of\_state\text{-}x}{X}{Z} \label{eq:rel_cons20} \\ 
 	\relationalign{head\_of\_gov\text{-}x}{X}{Z} \land \relationalign{gpe0}{Z}{Y} \implies \relationalign{head\_of\_gov}{X}{Y} \label{eq:rel_cons21} \\ 
 	\relationalign{head\_of\_gov}{X}{Y} \land \relationalign{gpe0}{Z}{Y} \implies \relationalign{head\_of\_gov\text{-}x}{X}{Z} \label{eq:rel_cons22} \\ 
 	\relationalign{in0\text{-}x}{X}{Z} \land \relationalign{gpe0}{Z}{Y} \implies \relationalign{in0}{X}{Y} \label{eq:rel_cons23} \\ 
 	\relationalign{in0}{X}{Y} \land \relationalign{gpe0}{Z}{Y} \implies \relationalign{in0\text{-}x}{X}{Z} \label{eq:rel_cons24} \\ 
 	\relationalign{in2}{X}{Z} \land \relationalign{in0}{Z}{Y} \implies \relationalign{in0}{X}{Y} \label{eq:rel_cons25} \\ 
 	\relationalign{in1}{X}{Z} \land \relationalign{in0}{Z}{Y} \implies \relationalign{in0}{X}{Y} \label{eq:rel_cons26} \\ 
 	\relationalign{based\_in2}{X}{Z} \land \relationalign{in0}{Z}{Y} \implies \relationalign{based\_in0}{X}{Y} \label{eq:rel_cons27} \\ 
 	\relationalign{based\_in1}{X}{Z} \land \relationalign{in0}{Z}{Y} \implies \relationalign{based\_in0}{X}{Y} \label{eq:rel_cons28} \\ 
  	\relationalign{agency\_of}{X}{Y} \land \relationalignsingle{gpe0}{Y} \implies \relationalign{based\_in0}{X}{Y} \label{eq:rel_cons29} \\ 
 	\relationalign{event\_in2}{X}{Z} \land \relationalign{in0}{Z}{Y} \implies \relationalign{event\_in0}{X}{Y} \label{eq:rel_cons30} \\ 
 	\relationalign{event\_in1}{X}{Z} \land \relationalign{in0}{Z}{Y} \implies \relationalign{event\_in0}{X}{Y} \label{eq:rel_cons31} \\ 
 	\relationalign{head\_of}{X}{Y} \implies \relationalign{member\_of}{X}{Y} \label{eq:rel_cons32} \\ 
 	\relationalign{coach\_of}{X}{Y} \implies \relationalign{member\_of}{X}{Y} \label{eq:rel_cons33} \\ 
 	\relationalign{spokesperson\_of}{X}{Y} \implies \relationalign{member\_of}{X}{Y} \label{eq:rel_cons34} \\ 
 	\relationalign{member\_of}{X}{Y} \land \relationalignsingle{sport\_player}{X} \implies \relationalign{player\_of}{X}{Y} \label{eq:rel_cons35} \\ 
 	\relationalign{mayor\_of}{X}{Y} \implies \relationalign{head\_of\_gov}{X}{Y} \label{eq:rel_cons36} \\ 
 	\relationalign{directed\_by}{X}{Y} \implies \relationalign{created\_by}{X}{Y} \label{eq:rel_cons37} \\ 
 	\relationalign{character\_in}{X}{Y} \land \relationalign{played\_by}{X}{Z} \implies \relationalign{plays\_in}{Z}{Y} \label{eq:rel_cons38} \\ 
 	\relationalign{institution\_of}{X}{Y} \implies \relationalign{part\_of}{X}{Y} \label{eq:rel_cons39} \\ 
 	\relationalign{based\_in0\text{-}x}{X}{Z} \land \relationalign{gpe0}{Z}{Y} \implies \relationalign{based\_in0}{X}{Y} \label{eq:rel_cons40} \\ 
 	\relationalign{based\_in0}{X}{Y} \land \relationalign{gpe0}{Z}{Y} \implies \relationalign{based\_in0\text{-}x}{X}{Z} \label{eq:rel_cons41} 
\end{align}

\bibliography{references}

\begin{thebibliography}{95}
\expandafter\ifx\csname natexlab\endcsname\relax\def\natexlab#1{#1}\fi
\providecommand{\url}[1]{\texttt{#1}}
\providecommand{\href}[2]{#2}
\providecommand{\path}[1]{#1}
\providecommand{\DOIprefix}{doi:}
\providecommand{\ArXivprefix}{arXiv:}
\providecommand{\URLprefix}{URL: }
\providecommand{\Pubmedprefix}{pmid:}
\providecommand{\doi}[1]{\href{http://dx.doi.org/#1}{\path{#1}}}
\providecommand{\Pubmed}[1]{\href{pmid:#1}{\path{#1}}}
\providecommand{\bibinfo}[2]{#2}
\ifx\xfnm\relax \def\xfnm[#1]{\unskip,\space#1}\fi
\bibitem[{Aguilar et~al.(2014)Aguilar, Beller, McNamee, Van~Durme, Strassel,
  Song \& Ellis}]{aguilar2014comparison}
\bibinfo{author}{Aguilar, J.}, \bibinfo{author}{Beller, C.},
  \bibinfo{author}{McNamee, P.}, \bibinfo{author}{Van~Durme, B.},
  \bibinfo{author}{Strassel, S.}, \bibinfo{author}{Song, Z.}, \&
  \bibinfo{author}{Ellis, J.} (\bibinfo{year}{2014}).
\newblock \bibinfo{title}{A comparison of the events and relations across ace,
  ere, tac-kbp, and framenet annotation standards}.
\newblock In {\it \bibinfo{booktitle}{Proceedings of the 2nd Workshop on
  EVENTS: Definition, Detection, Coreference, and Representation}\/} (pp.
  \bibinfo{pages}{45--53}).
\bibitem[{Akbik et~al.(2019)Akbik, Bergmann \& Vollgraf}]{akbik2019pooled}
\bibinfo{author}{Akbik, A.}, \bibinfo{author}{Bergmann, T.}, \&
  \bibinfo{author}{Vollgraf, R.} (\bibinfo{year}{2019}).
\newblock \bibinfo{title}{Pooled contextualized embeddings for named entity
  recognition}.
\newblock In {\it \bibinfo{booktitle}{Proceedings of the 2019 Conference of the
  North American Chapter of the Association for Computational Linguistics:
  Human Language Technologies}\/} (pp. \bibinfo{pages}{724--728}).
\bibitem[{Akbik et~al.(2018)Akbik, Blythe \& Vollgraf}]{akbik2018contextual}
\bibinfo{author}{Akbik, A.}, \bibinfo{author}{Blythe, D.}, \&
  \bibinfo{author}{Vollgraf, R.} (\bibinfo{year}{2018}).
\newblock \bibinfo{title}{Contextual string embeddings for sequence labeling}.
\newblock In {\it \bibinfo{booktitle}{Proceedings of the 2018 International
  Conference on Computational Linguistics}\/} (pp.
  \bibinfo{pages}{1638--1649}).
\bibitem[{Augenstein et~al.(2017)Augenstein, Das, Riedel, Vikraman \&
  McCallum}]{augenstein2017semeval}
\bibinfo{author}{Augenstein, I.}, \bibinfo{author}{Das, M.},
  \bibinfo{author}{Riedel, S.}, \bibinfo{author}{Vikraman, L.}, \&
  \bibinfo{author}{McCallum, A.} (\bibinfo{year}{2017}).
\newblock \bibinfo{title}{Semeval 2017 task 10: Scienceie-extracting keyphrases
  and relations from scientific publications}.
\newblock In {\it \bibinfo{booktitle}{Proceedings of the 11th International
  Workshop on Semantic Evaluation (SemEval-2017)}\/} (pp.
  \bibinfo{pages}{546--555}).
\bibitem[{Baevski et~al.(2019)Baevski, Edunov, Liu, Zettlemoyer \&
  Auli}]{baevski2019cloze}
\bibinfo{author}{Baevski, A.}, \bibinfo{author}{Edunov, S.},
  \bibinfo{author}{Liu, Y.}, \bibinfo{author}{Zettlemoyer, L.}, \&
  \bibinfo{author}{Auli, M.} (\bibinfo{year}{2019}).
\newblock \bibinfo{title}{Cloze-driven pretraining of self-attention networks}.
\newblock In {\it \bibinfo{booktitle}{Proceedings of the 2019 Conference on
  Empirical Methods in Natural Language Processing and International Joint
  Conference on Natural Language Processing}\/} (pp.
  \bibinfo{pages}{5363--5372}).
\bibitem[{Bagga \& Baldwin(1998)}]{bagga1998algorithms}
\bibinfo{author}{Bagga, A.}, \& \bibinfo{author}{Baldwin, B.}
  (\bibinfo{year}{1998}).
\newblock \bibinfo{title}{Algorithms for scoring coreference chains}.
\newblock In {\it \bibinfo{booktitle}{Proceedings of the 1998 International
  Conference on Language Resources and Evaluation Workshop on Linguistics
  Coreference}\/} (pp. \bibinfo{pages}{563--566}).
\bibitem[{Bekoulis et~al.(2017)Bekoulis, Deleu, Demeester \&
  Develder}]{bekoulis2017reconstructing}
\bibinfo{author}{Bekoulis, G.}, \bibinfo{author}{Deleu, J.},
  \bibinfo{author}{Demeester, T.}, \& \bibinfo{author}{Develder, C.}
  (\bibinfo{year}{2017}).
\newblock \bibinfo{title}{Reconstructing the house from the ad: Structured
  prediction on real estate classifieds}.
\newblock In {\it \bibinfo{booktitle}{Proceedings of the 15th Conference of the
  European Chapter of the Association for Computational Linguistics: Volume 2,
  Short Papers}\/} (pp. \bibinfo{pages}{274--279}).
\bibitem[{Bekoulis et~al.(2018{\natexlab{a}})Bekoulis, Deleu, Demeester \&
  Develder}]{bekoulis2018adversarial}
\bibinfo{author}{Bekoulis, G.}, \bibinfo{author}{Deleu, J.},
  \bibinfo{author}{Demeester, T.}, \& \bibinfo{author}{Develder, C.}
  (\bibinfo{year}{2018}{\natexlab{a}}).
\newblock \bibinfo{title}{Adversarial training for multi-context joint entity
  and relation extraction}.
\newblock In {\it \bibinfo{booktitle}{Proceedings of the 2018 Conference on
  Empirical Methods in Natural Language Processing}\/} (pp.
  \bibinfo{pages}{2830--2836}).
\bibitem[{Bekoulis et~al.(2018{\natexlab{b}})Bekoulis, Deleu, Demeester \&
  Develder}]{bekoulis2018joint}
\bibinfo{author}{Bekoulis, G.}, \bibinfo{author}{Deleu, J.},
  \bibinfo{author}{Demeester, T.}, \& \bibinfo{author}{Develder, C.}
  (\bibinfo{year}{2018}{\natexlab{b}}).
\newblock \bibinfo{title}{Joint entity recognition and relation extraction as a
  multi-head selection problem}.
\newblock {\it \bibinfo{journal}{Expert Systems with Applications}\/},  {\it
  \bibinfo{volume}{114}\/}, \bibinfo{pages}{34--45}.
\bibitem[{Bentivogli et~al.(2010)Bentivogli, Forner, Giuliano, Marchetti,
  Pianta \& Tymoshenko}]{bentivogli2010extending}
\bibinfo{author}{Bentivogli, L.}, \bibinfo{author}{Forner, P.},
  \bibinfo{author}{Giuliano, C.}, \bibinfo{author}{Marchetti, A.},
  \bibinfo{author}{Pianta, E.}, \& \bibinfo{author}{Tymoshenko, K.}
  (\bibinfo{year}{2010}).
\newblock \bibinfo{title}{Extending english ace 2005 corpus annotation with
  ground-truth links to wikipedia}.
\newblock In {\it \bibinfo{booktitle}{Proceedings of the 2nd Workshop on The
  People's Web Meets NLP: Collaboratively Constructed Semantic Resources}\/}
  (pp. \bibinfo{pages}{19--27}).
\bibitem[{Bhattacharjee et~al.(2020)Bhattacharjee, Haque, de~Buy~Wenniger \&
  Way}]{bhattacharjee2020investigating}
\bibinfo{author}{Bhattacharjee, S.}, \bibinfo{author}{Haque, R.},
  \bibinfo{author}{de~Buy~Wenniger, G.~M.}, \& \bibinfo{author}{Way, A.}
  (\bibinfo{year}{2020}).
\newblock \bibinfo{title}{Investigating query expansion and coreference
  resolution in question answering on bert}.
\newblock In {\it \bibinfo{booktitle}{International Conference on Applications
  of Natural Language to Information Systems}\/} (pp. \bibinfo{pages}{47--59}).
\newblock \bibinfo{organization}{Springer}.
\bibitem[{Broscheit(2019)}]{broscheit2019investigating}
\bibinfo{author}{Broscheit, S.} (\bibinfo{year}{2019}).
\newblock \bibinfo{title}{Investigating entity knowledge in bert with simple
  neural end-to-end entity linking}.
\newblock In {\it \bibinfo{booktitle}{Proceedings of the 23rd Conference on
  Computational Natural Language Learning (CoNLL)}\/} (pp.
  \bibinfo{pages}{677--685}).
\bibitem[{Chen et~al.(2017)Chen, Fisch, Weston \& Bordes}]{chen2017reading}
\bibinfo{author}{Chen, D.}, \bibinfo{author}{Fisch, A.},
  \bibinfo{author}{Weston, J.}, \& \bibinfo{author}{Bordes, A.}
  (\bibinfo{year}{2017}).
\newblock \bibinfo{title}{Reading wikipedia to answer open-domain questions}.
\newblock In {\it \bibinfo{booktitle}{Proceedings of the 55th Annual Meeting of
  the Association for Computational Linguistics (Volume 1: Long Papers)}\/}
  (pp. \bibinfo{pages}{1870--1879}).
\bibitem[{Chinchor \& Marsh(1998)}]{chinchor1998muc}
\bibinfo{author}{Chinchor, N.}, \& \bibinfo{author}{Marsh, E.}
  (\bibinfo{year}{1998}).
\newblock \bibinfo{title}{Muc-7 information extraction task definition}.
\newblock In {\it \bibinfo{booktitle}{Proceeding of the 1998 Message
  Understanding Conference (MUC-7)}\/} (pp. \bibinfo{pages}{359--367}).
\bibitem[{Chiu \& Nichols(2016)}]{chiu2016named}
\bibinfo{author}{Chiu, J.~P.}, \& \bibinfo{author}{Nichols, E.}
  (\bibinfo{year}{2016}).
\newblock \bibinfo{title}{Named entity recognition with bidirectional
  lstm-cnns}.
\newblock {\it \bibinfo{journal}{Transactions of the Association for
  Computational Linguistics}\/},  {\it \bibinfo{volume}{4}\/},
  \bibinfo{pages}{357--370}.
\bibitem[{Cifariello et~al.(2019)Cifariello, Ferragina \&
  Ponza}]{cifariello2019wiser}
\bibinfo{author}{Cifariello, P.}, \bibinfo{author}{Ferragina, P.}, \&
  \bibinfo{author}{Ponza, M.} (\bibinfo{year}{2019}).
\newblock \bibinfo{title}{Wiser: A semantic approach for expert finding in
  academia based on entity linking}.
\newblock {\it \bibinfo{journal}{Information Systems}\/},  {\it
  \bibinfo{volume}{82}\/}, \bibinfo{pages}{1--16}.
\bibitem[{Clark et~al.(2018)Clark, Luong, Manning \& Le}]{clark2018semi}
\bibinfo{author}{Clark, K.}, \bibinfo{author}{Luong, M.-T.},
  \bibinfo{author}{Manning, C.~D.}, \& \bibinfo{author}{Le, Q.}
  (\bibinfo{year}{2018}).
\newblock \bibinfo{title}{Semi-supervised sequence modeling with cross-view
  training}.
\newblock In {\it \bibinfo{booktitle}{Proceedings of the 2018 Conference on
  Empirical Methods in Natural Language Processing}\/} (pp.
  \bibinfo{pages}{1914--1925}).
\bibitem[{Cohen(1960)}]{cohen1960coefficient}
\bibinfo{author}{Cohen, J.} (\bibinfo{year}{1960}).
\newblock \bibinfo{title}{A coefficient of agreement for nominal scales}.
\newblock {\it \bibinfo{journal}{Educational and psychological measurement}\/},
   {\it \bibinfo{volume}{20}\/}, \bibinfo{pages}{37--46}.
\bibitem[{Derczynski et~al.(2017)Derczynski, Nichols, van Erp \&
  Limsopatham}]{derczynski2017results}
\bibinfo{author}{Derczynski, L.}, \bibinfo{author}{Nichols, E.},
  \bibinfo{author}{van Erp, M.}, \& \bibinfo{author}{Limsopatham, N.}
  (\bibinfo{year}{2017}).
\newblock \bibinfo{title}{Results of the wnut2017 shared task on novel and
  emerging entity recognition}.
\newblock In {\it \bibinfo{booktitle}{Proceedings of the 3rd Workshop on Noisy
  User-generated Text}\/} (pp. \bibinfo{pages}{140--147}).
\bibitem[{Devlin et~al.(2019)Devlin, Chang, Lee \& Toutanova}]{devlin2019bert}
\bibinfo{author}{Devlin, J.}, \bibinfo{author}{Chang, M.-W.},
  \bibinfo{author}{Lee, K.}, \& \bibinfo{author}{Toutanova, K.}
  (\bibinfo{year}{2019}).
\newblock \bibinfo{title}{Bert: Pre-training of deep bidirectional transformers
  for language understanding}.
\newblock In {\it \bibinfo{booktitle}{Proceedings of the 2019 Conference of the
  North American Chapter of the Association for Computational Linguistics:
  Human Language Technologies}\/} (pp. \bibinfo{pages}{4171--4186}).
\bibitem[{Dixit \& Al-Onaizan(2019)}]{dixit2019span}
\bibinfo{author}{Dixit, K.}, \& \bibinfo{author}{Al-Onaizan, Y.}
  (\bibinfo{year}{2019}).
\newblock \bibinfo{title}{Span-level model for relation extraction}.
\newblock In {\it \bibinfo{booktitle}{Proceedings of the 2019 Annual Meeting of
  the Association for Computational Linguistics}\/} (pp.
  \bibinfo{pages}{5308--5314}).
\bibitem[{Doddington et~al.(2004)Doddington, Mitchell, Przybocki, Ramshaw,
  Strassel \& Weischedel}]{doddington2004automatic}
\bibinfo{author}{Doddington, G.~R.}, \bibinfo{author}{Mitchell, A.},
  \bibinfo{author}{Przybocki, M.~A.}, \bibinfo{author}{Ramshaw, L.~A.},
  \bibinfo{author}{Strassel, S.~M.}, \& \bibinfo{author}{Weischedel, R.~M.}
  (\bibinfo{year}{2004}).
\newblock \bibinfo{title}{The automatic content extraction (ace) program -
  tasks, data, and evaluation}.
\newblock In {\it \bibinfo{booktitle}{Proceedings of the 2004 International
  Conference on Language Resources and Evaluation Workshop on Linguistics}\/}
  (pp. \bibinfo{pages}{837--840}).
\bibitem[{Durrett \& Klein(2013)}]{durrett2013easy}
\bibinfo{author}{Durrett, G.}, \& \bibinfo{author}{Klein, D.}
  (\bibinfo{year}{2013}).
\newblock \bibinfo{title}{Easy victories and uphill battles in coreference
  resolution}.
\newblock In {\it \bibinfo{booktitle}{Proceedings of the 2013 Conference on
  Empirical Methods in Natural Language Processing}\/} (pp.
  \bibinfo{pages}{1971--1982}).
\bibitem[{Ellis et~al.(2015)Ellis, Getman, Fore, Kuster, Song, Bies \&
  Strassel}]{ellis2015overview}
\bibinfo{author}{Ellis, J.}, \bibinfo{author}{Getman, J.},
  \bibinfo{author}{Fore, D.}, \bibinfo{author}{Kuster, N.},
  \bibinfo{author}{Song, Z.}, \bibinfo{author}{Bies, A.}, \&
  \bibinfo{author}{Strassel, S.~M.} (\bibinfo{year}{2015}).
\newblock \bibinfo{title}{Overview of linguistic resources for the {TAC} {KBP}
  2015 evaluations: Methodologies and results.}
\newblock In {\it \bibinfo{booktitle}{Proceedings of the 2015 Text Analysis
  Conference}\/}.
\bibitem[{Ellis et~al.(2014)Ellis, Getman \& Strassel}]{ellis2014overview}
\bibinfo{author}{Ellis, J.}, \bibinfo{author}{Getman, J.}, \&
  \bibinfo{author}{Strassel, S.~M.} (\bibinfo{year}{2014}).
\newblock \bibinfo{title}{Overview of linguistic resources for the tac kbp 2014
  evaluations: Planning, execution, and results}.
\newblock In {\it \bibinfo{booktitle}{Proceedings of TAC KBP 2014 Workshop,
  National Institute of Standards and Technology}\/} (pp.
  \bibinfo{pages}{17--18}).
\bibitem[{Eshel et~al.(2017)Eshel, Cohen, Radinsky, Markovitch, Yamada \&
  Levy}]{eshel2017named}
\bibinfo{author}{Eshel, Y.}, \bibinfo{author}{Cohen, N.},
  \bibinfo{author}{Radinsky, K.}, \bibinfo{author}{Markovitch, S.},
  \bibinfo{author}{Yamada, I.}, \& \bibinfo{author}{Levy, O.}
  (\bibinfo{year}{2017}).
\newblock \bibinfo{title}{Named entity disambiguation for noisy text}.
\newblock In {\it \bibinfo{booktitle}{Proceedings of the 2017 Conference on
  Computational Natural Language Learning}\/} (pp. \bibinfo{pages}{58--68}).
\bibitem[{Fei et~al.(2020)Fei, Ren \& Ji}]{fei2020boundaries}
\bibinfo{author}{Fei, H.}, \bibinfo{author}{Ren, Y.}, \& \bibinfo{author}{Ji,
  D.} (\bibinfo{year}{2020}).
\newblock \bibinfo{title}{Boundaries and edges rethinking: An end-to-end neural
  model for overlapping entity relation extraction}.
\newblock {\it \bibinfo{journal}{Information Processing \& Management}\/},
  {\it \bibinfo{volume}{57}\/}, \bibinfo{pages}{102311}.
\bibitem[{Fu et~al.(2019)Fu, Li \& Ma}]{fu2019graphrel}
\bibinfo{author}{Fu, T.-J.}, \bibinfo{author}{Li, P.-H.}, \&
  \bibinfo{author}{Ma, W.-Y.} (\bibinfo{year}{2019}).
\newblock \bibinfo{title}{Graphrel: Modeling text as relational graphs for
  joint entity and relation extraction}.
\newblock In {\it \bibinfo{booktitle}{Proceedings of the 2019 Annual Meeting of
  the Association for Computational Linguistics}\/} (pp.
  \bibinfo{pages}{1409--1418}).
\bibitem[{Gao et~al.(2019)Gao, Li, King \& Lyu}]{gao2019interconnected}
\bibinfo{author}{Gao, Y.}, \bibinfo{author}{Li, P.}, \bibinfo{author}{King,
  I.}, \& \bibinfo{author}{Lyu, M.~R.} (\bibinfo{year}{2019}).
\newblock \bibinfo{title}{Interconnected question generation with coreference
  alignment and conversation flow modeling}.
\newblock In {\it \bibinfo{booktitle}{Proceedings of the 57th Annual Meeting of
  the Association for Computational Linguistics}\/} (pp.
  \bibinfo{pages}{4853--4862}).
\bibitem[{Guo et~al.(2019)Guo, Zhang \& Lu}]{guo2019attention}
\bibinfo{author}{Guo, Z.}, \bibinfo{author}{Zhang, Y.}, \& \bibinfo{author}{Lu,
  W.} (\bibinfo{year}{2019}).
\newblock \bibinfo{title}{Attention guided graph convolutional networks for
  relation extraction}.
\newblock In {\it \bibinfo{booktitle}{Proceedings of the 2019 Annual Meeting of
  the Association for Computational Linguistics}\/} (pp.
  \bibinfo{pages}{241--251}).
\bibitem[{Han \& Wang(2020)}]{han2020novel}
\bibinfo{author}{Han, X.}, \& \bibinfo{author}{Wang, L.}
  (\bibinfo{year}{2020}).
\newblock \bibinfo{title}{A novel document-level relation extraction method
  based on bert and entity information}.
\newblock {\it \bibinfo{journal}{IEEE Access}\/}, .
\bibitem[{Han et~al.(2018)Han, Zhu, Yu, Wang, Yao, Liu \& Sun}]{han2018fewrel}
\bibinfo{author}{Han, X.}, \bibinfo{author}{Zhu, H.}, \bibinfo{author}{Yu, P.},
  \bibinfo{author}{Wang, Z.}, \bibinfo{author}{Yao, Y.}, \bibinfo{author}{Liu,
  Z.}, \& \bibinfo{author}{Sun, M.} (\bibinfo{year}{2018}).
\newblock \bibinfo{title}{Fewrel: A large-scale supervised few-shot relation
  classification dataset with state-of-the-art evaluation}.
\newblock In {\it \bibinfo{booktitle}{Proceedings of the 2018 Conference on
  Empirical Methods in Natural Language Processing}\/} (pp.
  \bibinfo{pages}{4803--4809}).
\bibitem[{Hendrickx et~al.(2010)Hendrickx, Kim, Kozareva, Nakov, S{\'e}aghdha,
  Pad{\'o}, Pennacchiotti, Romano \& Szpakowicz}]{hendrickx2010semeval}
\bibinfo{author}{Hendrickx, I.}, \bibinfo{author}{Kim, S.~N.},
  \bibinfo{author}{Kozareva, Z.}, \bibinfo{author}{Nakov, P.},
  \bibinfo{author}{S{\'e}aghdha, D.~{\'O}.}, \bibinfo{author}{Pad{\'o}, S.},
  \bibinfo{author}{Pennacchiotti, M.}, \bibinfo{author}{Romano, L.}, \&
  \bibinfo{author}{Szpakowicz, S.} (\bibinfo{year}{2010}).
\newblock \bibinfo{title}{Semeval-2010 task 8: Multi-way classification of
  semantic relations between pairs of nominals}.
\newblock In {\it \bibinfo{booktitle}{Proceedings of the 5th International
  Workshop on Semantic Evaluation}\/} (pp. \bibinfo{pages}{33--38}).
\bibitem[{Hoffart et~al.(2011)Hoffart, Yosef, Bordino, F{\"u}rstenau, Pinkal,
  Spaniol, Taneva, Thater \& Weikum}]{hoffart2011robust}
\bibinfo{author}{Hoffart, J.}, \bibinfo{author}{Yosef, M.~A.},
  \bibinfo{author}{Bordino, I.}, \bibinfo{author}{F{\"u}rstenau, H.},
  \bibinfo{author}{Pinkal, M.}, \bibinfo{author}{Spaniol, M.},
  \bibinfo{author}{Taneva, B.}, \bibinfo{author}{Thater, S.}, \&
  \bibinfo{author}{Weikum, G.} (\bibinfo{year}{2011}).
\newblock \bibinfo{title}{Robust disambiguation of named entities in text}.
\newblock In {\it \bibinfo{booktitle}{Proceedings of the 2011 Conference on
  Empirical Methods in Natural Language Processing}\/} (pp.
  \bibinfo{pages}{782--792}).
\bibitem[{Hovy et~al.(2006)Hovy, Marcus, Palmer, Ramshaw \&
  Weischedel}]{hovy2006ontonotes}
\bibinfo{author}{Hovy, E.}, \bibinfo{author}{Marcus, M.},
  \bibinfo{author}{Palmer, M.}, \bibinfo{author}{Ramshaw, L.}, \&
  \bibinfo{author}{Weischedel, R.} (\bibinfo{year}{2006}).
\newblock \bibinfo{title}{Ontonotes: the 90\% solution}.
\newblock In {\it \bibinfo{booktitle}{Proceedings of the 2006 Conference of the
  North American Chapter of the Association for Computational Linguistics:
  Human Language Technologies}\/} (pp. \bibinfo{pages}{57--60}).
\bibitem[{Hu et~al.(2019)Hu, Rohrbach, Darrell \& Saenko}]{hu2019language}
\bibinfo{author}{Hu, R.}, \bibinfo{author}{Rohrbach, A.},
  \bibinfo{author}{Darrell, T.}, \& \bibinfo{author}{Saenko, K.}
  (\bibinfo{year}{2019}).
\newblock \bibinfo{title}{Language-conditioned graph networks for relational
  reasoning}.
\newblock In {\it \bibinfo{booktitle}{Proceedings of the IEEE International
  Conference on Computer Vision}\/} (pp. \bibinfo{pages}{10294--10303}).
\bibitem[{Hu et~al.(2020)Hu, Ma, Li, Li \& Wang}]{hu2020cross}
\bibinfo{author}{Hu, W.}, \bibinfo{author}{Ma, B.}, \bibinfo{author}{Li, Z.},
  \bibinfo{author}{Li, Y.}, \& \bibinfo{author}{Wang, Y.}
  (\bibinfo{year}{2020}).
\newblock \bibinfo{title}{A cross-media deep relationship classification method
  using discrimination information}.
\newblock {\it \bibinfo{journal}{Information Processing \& Management}\/},
  {\it \bibinfo{volume}{57}\/}, \bibinfo{pages}{102344}.
\bibitem[{Ji et~al.(2010)Ji, Grishman, Dang, Griffitt \&
  Ellis}]{ji2010overview}
\bibinfo{author}{Ji, H.}, \bibinfo{author}{Grishman, R.},
  \bibinfo{author}{Dang, H.~T.}, \bibinfo{author}{Griffitt, K.}, \&
  \bibinfo{author}{Ellis, J.} (\bibinfo{year}{2010}).
\newblock \bibinfo{title}{Overview of the tac 2010 knowledge base population
  track}.
\newblock In {\it \bibinfo{booktitle}{Proceedings of the 2010 Text Analysis
  Conference}\/} (pp. \bibinfo{pages}{3--3}).
\bibitem[{Ji et~al.(2015)Ji, Nothman, Hachey \& Florian}]{ji2015overview}
\bibinfo{author}{Ji, H.}, \bibinfo{author}{Nothman, J.},
  \bibinfo{author}{Hachey, B.}, \& \bibinfo{author}{Florian, R.}
  (\bibinfo{year}{2015}).
\newblock \bibinfo{title}{Overview of tac-kbp2015 tri-lingual entity discovery
  and linking.}
\newblock In {\it \bibinfo{booktitle}{Proceedings of the 2015 Text Analysis
  Conference}\/}.
\bibitem[{Ji et~al.(2017)Ji, Pan, Zhang, Nothman, Mayfield, McNamee, Costello
  \& Hub}]{ji2017overview}
\bibinfo{author}{Ji, H.}, \bibinfo{author}{Pan, X.}, \bibinfo{author}{Zhang,
  B.}, \bibinfo{author}{Nothman, J.}, \bibinfo{author}{Mayfield, J.},
  \bibinfo{author}{McNamee, P.}, \bibinfo{author}{Costello, C.}, \&
  \bibinfo{author}{Hub, S.~I.} (\bibinfo{year}{2017}).
\newblock \bibinfo{title}{Overview of tac-kbp2017 13 languages entity discovery
  and linking.}
\newblock In {\it \bibinfo{booktitle}{Proceedings of the 2017 Text Analysis
  Conference}\/}.
\bibitem[{Kantor \& Globerson(2019)}]{kantor2019coreference}
\bibinfo{author}{Kantor, B.}, \& \bibinfo{author}{Globerson, A.}
  (\bibinfo{year}{2019}).
\newblock \bibinfo{title}{Coreference resolution with entity equalization}.
\newblock In {\it \bibinfo{booktitle}{Proceedings of the 2019 Annual Meeting of
  the Association for Computational Linguistics}\/} (pp.
  \bibinfo{pages}{673--677}).
\bibitem[{Karimi et~al.(2018)Karimi, Jannach \& Jugovac}]{karimi2018news}
\bibinfo{author}{Karimi, M.}, \bibinfo{author}{Jannach, D.}, \&
  \bibinfo{author}{Jugovac, M.} (\bibinfo{year}{2018}).
\newblock \bibinfo{title}{News recommender systems--survey and roads ahead}.
\newblock {\it \bibinfo{journal}{Information Processing \& Management}\/},
  {\it \bibinfo{volume}{54}\/}, \bibinfo{pages}{1203--1227}.
\bibitem[{Katiyar \& Cardie(2018)}]{katiyar2018nested}
\bibinfo{author}{Katiyar, A.}, \& \bibinfo{author}{Cardie, C.}
  (\bibinfo{year}{2018}).
\newblock \bibinfo{title}{Nested named entity recognition revisited}.
\newblock In {\it \bibinfo{booktitle}{Proceedings of the 2018 Conference of the
  North American Chapter of the Association for Computational Linguistics:
  Human Language Technologies}\/} (pp. \bibinfo{pages}{861--871}).
\bibitem[{Kim et~al.(2003)Kim, Ohta, Tateisi \& Tsujii}]{kim2003genia}
\bibinfo{author}{Kim, J.-D.}, \bibinfo{author}{Ohta, T.},
  \bibinfo{author}{Tateisi, Y.}, \& \bibinfo{author}{Tsujii, J.}
  (\bibinfo{year}{2003}).
\newblock \bibinfo{title}{Genia corpus - a semantically annotated corpus for
  bio-textmining}.
\newblock {\it \bibinfo{journal}{Bioinformatics}\/},  {\it
  \bibinfo{volume}{19}\/}, \bibinfo{pages}{180--182}.
\bibitem[{Kulkarni et~al.(2018)Kulkarni, Xu, Ritter \&
  Machiraju}]{kulkarni2018annotated}
\bibinfo{author}{Kulkarni, C.}, \bibinfo{author}{Xu, W.},
  \bibinfo{author}{Ritter, A.}, \& \bibinfo{author}{Machiraju, R.}
  (\bibinfo{year}{2018}).
\newblock \bibinfo{title}{An annotated corpus for machine reading of
  instructions in wet lab protocols}.
\newblock In {\it \bibinfo{booktitle}{Proceedings of the 2018 Conference of the
  North American Chapter of the Association for Computational Linguistics:
  Human Language Technologies}\/} (pp. \bibinfo{pages}{97--106}).
\bibitem[{Kulkarni et~al.(2009)Kulkarni, Singh, Ramakrishnan \&
  Chakrabarti}]{kulkarni2009collective}
\bibinfo{author}{Kulkarni, S.}, \bibinfo{author}{Singh, A.},
  \bibinfo{author}{Ramakrishnan, G.}, \& \bibinfo{author}{Chakrabarti, S.}
  (\bibinfo{year}{2009}).
\newblock \bibinfo{title}{Collective annotation of wikipedia entities in web
  text}.
\newblock In {\it \bibinfo{booktitle}{Proceedings of the 15th ACM SIGKDD
  international conference on Knowledge discovery and data mining}\/} (pp.
  \bibinfo{pages}{457--466}).
\bibitem[{Lample et~al.(2016)Lample, Ballesteros, Subramanian, Kawakami \&
  Dyer}]{lample2016neural}
\bibinfo{author}{Lample, G.}, \bibinfo{author}{Ballesteros, M.},
  \bibinfo{author}{Subramanian, S.}, \bibinfo{author}{Kawakami, K.}, \&
  \bibinfo{author}{Dyer, C.} (\bibinfo{year}{2016}).
\newblock \bibinfo{title}{Neural architectures for named entity recognition}.
\newblock In {\it \bibinfo{booktitle}{Proceedings of the 2016 Conference of the
  North American Chapter of the Association for Computational Linguistics:
  Human Language Technologies}\/} (pp. \bibinfo{pages}{260--270}).
\bibitem[{Landis \& Koch(1977)}]{landis1977measurement}
\bibinfo{author}{Landis, J.}, \& \bibinfo{author}{Koch, G.}
  (\bibinfo{year}{1977}).
\newblock \bibinfo{title}{The measurement of observer agreement for categorical
  data.}
\newblock {\it \bibinfo{journal}{Biometrics}\/},  {\it \bibinfo{volume}{33}\/},
  \bibinfo{pages}{159--174}.
\bibitem[{Lee et~al.(2017)Lee, He, Lewis \& Zettlemoyer}]{lee2017end}
\bibinfo{author}{Lee, K.}, \bibinfo{author}{He, L.}, \bibinfo{author}{Lewis,
  M.}, \& \bibinfo{author}{Zettlemoyer, L.} (\bibinfo{year}{2017}).
\newblock \bibinfo{title}{End-to-end neural coreference resolution}.
\newblock In {\it \bibinfo{booktitle}{Proceedings of the 2017 Conference on
  Empirical Methods in Natural Language Processing}\/} (pp.
  \bibinfo{pages}{188--197}).
\bibitem[{Lee et~al.(2018)Lee, He \& Zettlemoyer}]{lee2018higher}
\bibinfo{author}{Lee, K.}, \bibinfo{author}{He, L.}, \&
  \bibinfo{author}{Zettlemoyer, L.} (\bibinfo{year}{2018}).
\newblock \bibinfo{title}{Higher-order coreference resolution with
  coarse-to-fine inference}.
\newblock In {\it \bibinfo{booktitle}{Proceedings of the 2018 Conference of the
  North American Chapter of the Association for Computational Linguistics:
  Human Language Technologies}\/} (pp. \bibinfo{pages}{687--692}).
\bibitem[{Li et~al.(2016{\natexlab{a}})Li, Sun, Johnson, Sciaky, Wei, Leaman,
  Davis, Mattingly, Wiegers \& Lu}]{li2016biocreative}
\bibinfo{author}{Li, J.}, \bibinfo{author}{Sun, Y.}, \bibinfo{author}{Johnson,
  R.~J.}, \bibinfo{author}{Sciaky, D.}, \bibinfo{author}{Wei, C.-H.},
  \bibinfo{author}{Leaman, R.}, \bibinfo{author}{Davis, A.~P.},
  \bibinfo{author}{Mattingly, C.~J.}, \bibinfo{author}{Wiegers, T.~C.}, \&
  \bibinfo{author}{Lu, Z.} (\bibinfo{year}{2016}{\natexlab{a}}).
\newblock \bibinfo{title}{Biocreative v cdr task corpus: a resource for
  chemical disease relation extraction}.
\newblock {\it \bibinfo{journal}{Database}\/},  {\it \bibinfo{volume}{2016}\/}.
\bibitem[{Li \& Ji(2014)}]{li2014incremental}
\bibinfo{author}{Li, Q.}, \& \bibinfo{author}{Ji, H.} (\bibinfo{year}{2014}).
\newblock \bibinfo{title}{Incremental joint extraction of entity mentions and
  relations}.
\newblock In {\it \bibinfo{booktitle}{Proceedings of the 2014 Annual Meeting of
  the Association for Computational Linguistics}\/} (pp.
  \bibinfo{pages}{402--412}).
\bibitem[{Li et~al.(2016{\natexlab{b}})Li, Tarlow, Brockschmidt \&
  Zemel}]{li2015gated}
\bibinfo{author}{Li, Y.}, \bibinfo{author}{Tarlow, D.},
  \bibinfo{author}{Brockschmidt, M.}, \& \bibinfo{author}{Zemel, R.}
  (\bibinfo{year}{2016}{\natexlab{b}}).
\newblock \bibinfo{title}{Gated graph sequence neural networks}.
\newblock In {\it \bibinfo{booktitle}{Proceedings of the 2016 International
  Conference on Learning Representations}\/}.
\bibitem[{Luan et~al.(2018)Luan, He, Ostendorf \& Hajishirzi}]{luan2018multi}
\bibinfo{author}{Luan, Y.}, \bibinfo{author}{He, L.},
  \bibinfo{author}{Ostendorf, M.}, \& \bibinfo{author}{Hajishirzi, H.}
  (\bibinfo{year}{2018}).
\newblock \bibinfo{title}{Multi-task identification of entities, relations, and
  coreference for scientific knowledge graph construction}.
\newblock In {\it \bibinfo{booktitle}{Proceedings of the 2018 Conference on
  Empirical Methods in Natural Language Processing}\/} (pp.
  \bibinfo{pages}{3219--3232}).
\bibitem[{Luan et~al.(2017)Luan, Ostendorf \& Hajishirzi}]{luan2017scientific}
\bibinfo{author}{Luan, Y.}, \bibinfo{author}{Ostendorf, M.}, \&
  \bibinfo{author}{Hajishirzi, H.} (\bibinfo{year}{2017}).
\newblock \bibinfo{title}{Scientific information extraction with
  semi-supervised neural tagging}.
\newblock In {\it \bibinfo{booktitle}{Proceedings of the 2017 Conference on
  Empirical Methods in Natural Language Processing}\/} (pp.
  \bibinfo{pages}{2641--2651}).
\bibitem[{Luan et~al.(2019)Luan, Wadden, He, Shah, Ostendorf \&
  Hajishirzi}]{luan2019general}
\bibinfo{author}{Luan, Y.}, \bibinfo{author}{Wadden, D.}, \bibinfo{author}{He,
  L.}, \bibinfo{author}{Shah, A.}, \bibinfo{author}{Ostendorf, M.}, \&
  \bibinfo{author}{Hajishirzi, H.} (\bibinfo{year}{2019}).
\newblock \bibinfo{title}{A general framework for information extraction using
  dynamic span graphs}.
\newblock In {\it \bibinfo{booktitle}{Proceedings of the 2019 Conference of the
  North American Chapter of the Association for Computational Linguistics:
  Human Language Technologies}\/} (pp. \bibinfo{pages}{3036--3046}).
\bibitem[{Luo(2005)}]{luo2005coreference}
\bibinfo{author}{Luo, X.} (\bibinfo{year}{2005}).
\newblock \bibinfo{title}{On coreference resolution performance metrics}.
\newblock In {\it \bibinfo{booktitle}{Proceedings of the 2005 Conference on
  Human Language Technology and Empirical Methods in Natural Language
  Processing}\/} (pp. \bibinfo{pages}{25--32}).
\bibitem[{Ma \& Hovy(2016)}]{ma2016end}
\bibinfo{author}{Ma, X.}, \& \bibinfo{author}{Hovy, E.} (\bibinfo{year}{2016}).
\newblock \bibinfo{title}{End-to-end sequence labeling via bi-directional
  lstm-cnns-crf}.
\newblock In {\it \bibinfo{booktitle}{Proceedings of the 2016 Annual Meeting of
  the Association for Computational Linguistics}\/} (pp.
  \bibinfo{pages}{1064--1074}).
\bibitem[{McHugh(2012)}]{mchugh2012interrater}
\bibinfo{author}{McHugh, M.~L.} (\bibinfo{year}{2012}).
\newblock \bibinfo{title}{Interrater reliability: the kappa statistic}.
\newblock {\it \bibinfo{journal}{Biochemia medica: Biochemia medica}\/},  {\it
  \bibinfo{volume}{22}\/}, \bibinfo{pages}{276--282}.
\bibitem[{Molla et~al.(2006)Molla, van Zaanen \& Smith}]{molla2006named}
\bibinfo{author}{Molla, D.}, \bibinfo{author}{van Zaanen, M.}, \&
  \bibinfo{author}{Smith, D.} (\bibinfo{year}{2006}).
\newblock \bibinfo{title}{Named entity recognition for question answering}.
\newblock In {\it \bibinfo{booktitle}{Proceedings of the Australasian Language
  Technology Workshop 2006}\/} (pp. \bibinfo{pages}{51--58}).
\bibitem[{Peng et~al.(2017)Peng, Poon, Quirk, Toutanova \& Yih}]{peng2017cross}
\bibinfo{author}{Peng, N.}, \bibinfo{author}{Poon, H.}, \bibinfo{author}{Quirk,
  C.}, \bibinfo{author}{Toutanova, K.}, \& \bibinfo{author}{Yih, W.-t.}
  (\bibinfo{year}{2017}).
\newblock \bibinfo{title}{Cross-sentence n-ary relation extraction with graph
  lstms}.
\newblock {\it \bibinfo{journal}{Transactions of the Association for
  Computational Linguistics}\/},  {\it \bibinfo{volume}{5}\/},
  \bibinfo{pages}{101--115}.
\bibitem[{Pennington et~al.(2014)Pennington, Socher \&
  Manning}]{pennington2014glove}
\bibinfo{author}{Pennington, J.}, \bibinfo{author}{Socher, R.}, \&
  \bibinfo{author}{Manning, C.~D.} (\bibinfo{year}{2014}).
\newblock \bibinfo{title}{Glove: Global vectors for word representation}.
\newblock In {\it \bibinfo{booktitle}{Proceedings of the 2014 conference on
  empirical methods in natural language processing}\/} (pp.
  \bibinfo{pages}{1532--1543}).
\bibitem[{Peters et~al.(2019)Peters, Neumann, Logan, Schwartz, Joshi, Singh \&
  Smith}]{peters2019knowledge}
\bibinfo{author}{Peters, M.~E.}, \bibinfo{author}{Neumann, M.},
  \bibinfo{author}{Logan, R.}, \bibinfo{author}{Schwartz, R.},
  \bibinfo{author}{Joshi, V.}, \bibinfo{author}{Singh, S.}, \&
  \bibinfo{author}{Smith, N.~A.} (\bibinfo{year}{2019}).
\newblock \bibinfo{title}{Knowledge enhanced contextual word representations}.
\newblock In {\it \bibinfo{booktitle}{Proceedings of the 2019 Conference on
  Empirical Methods in Natural Language Processing and the International Joint
  Conference on Natural Language Processing}\/} (pp. \bibinfo{pages}{43--54}).
\bibitem[{Pradhan et~al.(2014)Pradhan, Luo, Recasens, Hovy, Ng \&
  Strube}]{pradhan2014scoring}
\bibinfo{author}{Pradhan, S.}, \bibinfo{author}{Luo, X.},
  \bibinfo{author}{Recasens, M.}, \bibinfo{author}{Hovy, E.},
  \bibinfo{author}{Ng, V.}, \& \bibinfo{author}{Strube, M.}
  (\bibinfo{year}{2014}).
\newblock \bibinfo{title}{Scoring coreference partitions of predicted mentions:
  A reference implementation}.
\newblock In {\it \bibinfo{booktitle}{Proceedings of the 2014 Annual Meeting of
  the Association for Computational Linguistics}\/} (pp.
  \bibinfo{pages}{30--35}).
\bibitem[{Pradhan et~al.(2012)Pradhan, Moschitti, Xue, Uryupina \&
  Zhang}]{pradhan2012conll}
\bibinfo{author}{Pradhan, S.}, \bibinfo{author}{Moschitti, A.},
  \bibinfo{author}{Xue, N.}, \bibinfo{author}{Uryupina, O.}, \&
  \bibinfo{author}{Zhang, Y.} (\bibinfo{year}{2012}).
\newblock \bibinfo{title}{Conll-2012 shared task: Modeling multilingual
  unrestricted coreference in ontonotes}.
\newblock In {\it \bibinfo{booktitle}{Proceedings of the 2012 Conference on
  Computational Natural Language Learning}\/} (pp. \bibinfo{pages}{1--40}).
\bibitem[{Quirk \& Poon(2017)}]{quirk2017distant}
\bibinfo{author}{Quirk, C.}, \& \bibinfo{author}{Poon, H.}
  (\bibinfo{year}{2017}).
\newblock \bibinfo{title}{Distant supervision for relation extraction beyond
  the sentence boundary}.
\newblock In {\it \bibinfo{booktitle}{Proceedings of the 2017 Conference of the
  European Chapter of the Association for Computational Linguistics}\/} (pp.
  \bibinfo{pages}{1171--1182}).
\bibitem[{Riedel et~al.(2010)Riedel, Yao \& McCallum}]{riedel2010modeling}
\bibinfo{author}{Riedel, S.}, \bibinfo{author}{Yao, L.}, \&
  \bibinfo{author}{McCallum, A.} (\bibinfo{year}{2010}).
\newblock \bibinfo{title}{Modeling relations and their mentions without labeled
  text}.
\newblock In {\it \bibinfo{booktitle}{Proceedings of the 2010 European
  Conference on Machine Learning and Knowledge Discovery in Databases}\/} (pp.
  \bibinfo{pages}{148--163}).
\bibitem[{Roller et~al.(2020)Roller, Dinan, Goyal, Ju, Williamson, Liu, Xu,
  Ott, Shuster, Smith et~al.}]{roller2020recipes}
\bibinfo{author}{Roller, S.}, \bibinfo{author}{Dinan, E.},
  \bibinfo{author}{Goyal, N.}, \bibinfo{author}{Ju, D.},
  \bibinfo{author}{Williamson, M.}, \bibinfo{author}{Liu, Y.},
  \bibinfo{author}{Xu, J.}, \bibinfo{author}{Ott, M.},
  \bibinfo{author}{Shuster, K.}, \bibinfo{author}{Smith, E.~M.} et~al.
  (\bibinfo{year}{2020}).
\newblock \bibinfo{title}{Recipes for building an open-domain chatbot}.
\newblock {\it \bibinfo{journal}{arXiv preprint arXiv:2004.13637}\/}, .
\bibitem[{Sang \& De~Meulder(2003)}]{sang2003introduction}
\bibinfo{author}{Sang, E. F. T.~K.}, \& \bibinfo{author}{De~Meulder, F.}
  (\bibinfo{year}{2003}).
\newblock \bibinfo{title}{Introduction to the conll-2003 shared task:
  Language-independent named entity recognition}.
\newblock In {\it \bibinfo{booktitle}{Proceedings of the 2003 Conference of the
  North American Chapter of the Association for Computational Linguistics:
  Human Language Technologies}\/} (pp. \bibinfo{pages}{142--147}).
\bibitem[{Scarselli et~al.(2008)Scarselli, Gori, Tsoi, Hagenbuchner \&
  Monfardini}]{scarselli2008graph}
\bibinfo{author}{Scarselli, F.}, \bibinfo{author}{Gori, M.},
  \bibinfo{author}{Tsoi, A.~C.}, \bibinfo{author}{Hagenbuchner, M.}, \&
  \bibinfo{author}{Monfardini, G.} (\bibinfo{year}{2008}).
\newblock \bibinfo{title}{The graph neural network model}.
\newblock {\it \bibinfo{journal}{IEEE Transactions on Neural Networks}\/},
  {\it \bibinfo{volume}{20}\/}, \bibinfo{pages}{61--80}.
\bibitem[{Scott(1955)}]{scott1955reliability}
\bibinfo{author}{Scott, W.~A.} (\bibinfo{year}{1955}).
\newblock \bibinfo{title}{Reliability of content analysis: The case of nominal
  scale coding}.
\newblock {\it \bibinfo{journal}{Public opinion quarterly}\/},  (pp.
  \bibinfo{pages}{321--325}).
\bibitem[{Singh et~al.(2018)Singh, Radhakrishna, Both, Shekarpour, Lytra,
  Usbeck, Vyas, Khikmatullaev, Punjani, Lange et~al.}]{singh2018reinvent}
\bibinfo{author}{Singh, K.}, \bibinfo{author}{Radhakrishna, A.~S.},
  \bibinfo{author}{Both, A.}, \bibinfo{author}{Shekarpour, S.},
  \bibinfo{author}{Lytra, I.}, \bibinfo{author}{Usbeck, R.},
  \bibinfo{author}{Vyas, A.}, \bibinfo{author}{Khikmatullaev, A.},
  \bibinfo{author}{Punjani, D.}, \bibinfo{author}{Lange, C.} et~al.
  (\bibinfo{year}{2018}).
\newblock \bibinfo{title}{Why reinvent the wheel: Let's build question
  answering systems together}.
\newblock In {\it \bibinfo{booktitle}{Proceedings of the 2018 World Wide Web
  Conference}\/} (pp. \bibinfo{pages}{1247--1256}).
\bibitem[{Soares et~al.(2019)Soares, FitzGerald, Ling \&
  Kwiatkowski}]{soares2019matching}
\bibinfo{author}{Soares, L.~B.}, \bibinfo{author}{FitzGerald, N.},
  \bibinfo{author}{Ling, J.}, \& \bibinfo{author}{Kwiatkowski, T.}
  (\bibinfo{year}{2019}).
\newblock \bibinfo{title}{Matching the blanks: Distributional similarity for
  relation learning}.
\newblock In {\it \bibinfo{booktitle}{Proceedings of the 2019 Annual Meeting of
  the Association for Computational Linguistics}\/} (pp.
  \bibinfo{pages}{2895--2905}).
\bibitem[{Song et~al.(2015)Song, Bies, Strassel, Riese, Mott, Ellis, Wright,
  Kulick, Ryant \& Ma}]{song2015light}
\bibinfo{author}{Song, Z.}, \bibinfo{author}{Bies, A.},
  \bibinfo{author}{Strassel, S.}, \bibinfo{author}{Riese, T.},
  \bibinfo{author}{Mott, J.}, \bibinfo{author}{Ellis, J.},
  \bibinfo{author}{Wright, J.}, \bibinfo{author}{Kulick, S.},
  \bibinfo{author}{Ryant, N.}, \& \bibinfo{author}{Ma, X.}
  (\bibinfo{year}{2015}).
\newblock \bibinfo{title}{From light to rich ere: annotation of entities,
  relations, and events}.
\newblock In {\it \bibinfo{booktitle}{Proceedings of the the 3rd Workshop on
  EVENTS: Definition, Detection, Coreference, and Representation}\/} (pp.
  \bibinfo{pages}{89--98}).
\bibitem[{Strubell et~al.(2017)Strubell, Verga, Belanger \&
  McCallum}]{strubell2017fast}
\bibinfo{author}{Strubell, E.}, \bibinfo{author}{Verga, P.},
  \bibinfo{author}{Belanger, D.}, \& \bibinfo{author}{McCallum, A.}
  (\bibinfo{year}{2017}).
\newblock \bibinfo{title}{Fast and accurate entity recognition with iterated
  dilated convolutions}.
\newblock In {\it \bibinfo{booktitle}{Proceedings of the 2017 Conference on
  Empirical Methods in Natural Language Processing}\/} (pp.
  \bibinfo{pages}{2670--2680}).
\bibitem[{Sun et~al.(2017)Sun, Luo \& Chen}]{sun2017review}
\bibinfo{author}{Sun, S.}, \bibinfo{author}{Luo, C.}, \& \bibinfo{author}{Chen,
  J.} (\bibinfo{year}{2017}).
\newblock \bibinfo{title}{A review of natural language processing techniques
  for opinion mining systems}.
\newblock {\it \bibinfo{journal}{Information fusion}\/},  {\it
  \bibinfo{volume}{36}\/}, \bibinfo{pages}{10--25}.
\bibitem[{Thorne \& Vlachos(2018)}]{thorne2018automated}
\bibinfo{author}{Thorne, J.}, \& \bibinfo{author}{Vlachos, A.}
  (\bibinfo{year}{2018}).
\newblock \bibinfo{title}{Automated fact checking: Task formulations, methods
  and future directions}.
\newblock In {\it \bibinfo{booktitle}{Proceedings of the 27th International
  Conference on Computational Linguistics}\/} (pp.
  \bibinfo{pages}{3346--3359}).
\bibitem[{Wadden et~al.(2019)Wadden, Wennberg, Luan \&
  Hajishirzi}]{wadden2019entity}
\bibinfo{author}{Wadden, D.}, \bibinfo{author}{Wennberg, U.},
  \bibinfo{author}{Luan, Y.}, \& \bibinfo{author}{Hajishirzi, H.}
  (\bibinfo{year}{2019}).
\newblock \bibinfo{title}{Entity, relation, and event extraction with
  contextualized span representations}.
\newblock In {\it \bibinfo{booktitle}{Proceedings of the 2019 Conference on
  Empirical Methods in Natural Language Processing and International Joint
  Conference on Natural Language Processing}\/} (pp.
  \bibinfo{pages}{5788--5793}).
\bibitem[{Walker et~al.(2006)Walker, Strassel, Medero \& Maeda}]{walker2006ace}
\bibinfo{author}{Walker, C.}, \bibinfo{author}{Strassel, S.},
  \bibinfo{author}{Medero, J.}, \& \bibinfo{author}{Maeda, K.}
  (\bibinfo{year}{2006}).
\newblock \bibinfo{title}{Ace 2005 multilingual training corpus}.
\newblock {\it \bibinfo{journal}{Linguistic Data Consortium, Philadelphia}\/},
  {\it \bibinfo{volume}{57}\/}.
\bibitem[{Wang et~al.(2018)Wang, Zhang, Xie \& Guo}]{wang2018dkn}
\bibinfo{author}{Wang, H.}, \bibinfo{author}{Zhang, F.}, \bibinfo{author}{Xie,
  X.}, \& \bibinfo{author}{Guo, M.} (\bibinfo{year}{2018}).
\newblock \bibinfo{title}{Dkn: Deep knowledge-aware network for news
  recommendation}.
\newblock In {\it \bibinfo{booktitle}{Proceedings of the 2018 world wide web
  conference}\/} (pp. \bibinfo{pages}{1835--1844}).
\bibitem[{Wang et~al.(2019)Wang, Zhang, Zhao, Li, Xie \& Guo}]{wang2019multi}
\bibinfo{author}{Wang, H.}, \bibinfo{author}{Zhang, F.}, \bibinfo{author}{Zhao,
  M.}, \bibinfo{author}{Li, W.}, \bibinfo{author}{Xie, X.}, \&
  \bibinfo{author}{Guo, M.} (\bibinfo{year}{2019}).
\newblock \bibinfo{title}{Multi-task feature learning for knowledge graph
  enhanced recommendation}.
\newblock In {\it \bibinfo{booktitle}{The World Wide Web Conference}\/} (pp.
  \bibinfo{pages}{2000--2010}).
\bibitem[{Webster et~al.(2018)Webster, Recasens, Axelrod \&
  Baldridge}]{webster2018mind}
\bibinfo{author}{Webster, K.}, \bibinfo{author}{Recasens, M.},
  \bibinfo{author}{Axelrod, V.}, \& \bibinfo{author}{Baldridge, J.}
  (\bibinfo{year}{2018}).
\newblock \bibinfo{title}{Mind the gap: A balanced corpus of gendered ambiguous
  pronouns}.
\newblock {\it \bibinfo{journal}{Transactions of the Association for
  Computational Linguistics}\/},  {\it \bibinfo{volume}{6}\/},
  \bibinfo{pages}{605--617}.
\bibitem[{Wei et~al.(2015)Wei, Peng, Leaman, Davis, Mattingly, Li, Wiegers \&
  Lu}]{wei2015overview}
\bibinfo{author}{Wei, C.-H.}, \bibinfo{author}{Peng, Y.},
  \bibinfo{author}{Leaman, R.}, \bibinfo{author}{Davis, A.~P.},
  \bibinfo{author}{Mattingly, C.~J.}, \bibinfo{author}{Li, J.},
  \bibinfo{author}{Wiegers, T.~C.}, \& \bibinfo{author}{Lu, Z.}
  (\bibinfo{year}{2015}).
\newblock \bibinfo{title}{Overview of the biocreative v chemical disease
  relation (cdr) task}.
\newblock In {\it \bibinfo{booktitle}{Proceedings of the 5th BioCreative
  Challenge Evaluation Workshop}\/}.
\bibitem[{Weischedel et~al.(2011)Weischedel, Hovy, Marcus, Palmer, Belvin,
  Pradhan, Ramshaw \& Xue}]{weischedel2011ontonotes}
\bibinfo{author}{Weischedel, R.}, \bibinfo{author}{Hovy, E.},
  \bibinfo{author}{Marcus, M.}, \bibinfo{author}{Palmer, M.},
  \bibinfo{author}{Belvin, R.}, \bibinfo{author}{Pradhan, S.},
  \bibinfo{author}{Ramshaw, L.}, \& \bibinfo{author}{Xue, N.}
  (\bibinfo{year}{2011}).
\newblock \bibinfo{title}{Ontonotes: A large training corpus for enhanced
  processing}.
\newblock {\it \bibinfo{journal}{Handbook of Natural Language Processing and
  Machine Translation. Springer}\/},  (p.~\bibinfo{pages}{59}).
\bibitem[{Weischedel et~al.(2013)Weischedel, Palmer, Marcus, Hovy, Pradhan,
  Ramshaw, Xue, Taylor, Kaufman, Franchini et~al.}]{weischedel2013ontonotes}
\bibinfo{author}{Weischedel, R.}, \bibinfo{author}{Palmer, M.},
  \bibinfo{author}{Marcus, M.}, \bibinfo{author}{Hovy, E.},
  \bibinfo{author}{Pradhan, S.}, \bibinfo{author}{Ramshaw, L.},
  \bibinfo{author}{Xue, N.}, \bibinfo{author}{Taylor, A.},
  \bibinfo{author}{Kaufman, J.}, \bibinfo{author}{Franchini, M.} et~al.
  (\bibinfo{year}{2013}).
\newblock \bibinfo{title}{Ontonotes release 5.0 ldc2013t19}.
\newblock {\it \bibinfo{journal}{Linguistic Data Consortium, Philadelphia,
  PA}\/},  {\it \bibinfo{volume}{23}\/}.
\bibitem[{Wiseman et~al.(2015)Wiseman, Rush, Shieber \&
  Weston}]{wiseman2015learning}
\bibinfo{author}{Wiseman, S.}, \bibinfo{author}{Rush, A.~M.},
  \bibinfo{author}{Shieber, S.~M.}, \& \bibinfo{author}{Weston, J.}
  (\bibinfo{year}{2015}).
\newblock \bibinfo{title}{Learning anaphoricity and antecedent ranking features
  for coreference resolution}.
\newblock In {\it \bibinfo{booktitle}{Proceedings of the 2015 Annual Meeting of
  the Association for Computational Linguistics and International Joint
  Conference on Natural Language Processing}\/} (pp.
  \bibinfo{pages}{1416--1426}).
\bibitem[{Wu \& He(2019)}]{wu2019enriching}
\bibinfo{author}{Wu, S.}, \& \bibinfo{author}{He, Y.} (\bibinfo{year}{2019}).
\newblock \bibinfo{title}{Enriching pre-trained language model with entity
  information for relation classification}.
\newblock In {\it \bibinfo{booktitle}{Proceedings of the 2019 ACM International
  Conference on Information and Knowledge Management}\/} (pp.
  \bibinfo{pages}{2361--2364}).
\bibitem[{Wu et~al.(2020)Wu, Pan, Chen, Long, Zhang \&
  Philip}]{wu2020comprehensive}
\bibinfo{author}{Wu, Z.}, \bibinfo{author}{Pan, S.}, \bibinfo{author}{Chen,
  F.}, \bibinfo{author}{Long, G.}, \bibinfo{author}{Zhang, C.}, \&
  \bibinfo{author}{Philip, S.~Y.} (\bibinfo{year}{2020}).
\newblock \bibinfo{title}{A comprehensive survey on graph neural networks}.
\newblock {\it \bibinfo{journal}{IEEE Transactions on Neural Networks and
  Learning Systems}\/},  (pp. \bibinfo{pages}{1--21}).
\bibitem[{Xu et~al.(2018)Xu, Hu, Leskovec \& Jegelka}]{xu2018powerful}
\bibinfo{author}{Xu, K.}, \bibinfo{author}{Hu, W.}, \bibinfo{author}{Leskovec,
  J.}, \& \bibinfo{author}{Jegelka, S.} (\bibinfo{year}{2018}).
\newblock \bibinfo{title}{How powerful are graph neural networks?}
\newblock In {\it \bibinfo{booktitle}{Proceedings of the 2018 International
  Conference on Learning Representations}\/}.
\bibitem[{Yao et~al.(2019)Yao, Ye, Li, Han, Lin, Liu, Liu, Huang, Zhou \&
  Sun}]{yao2019docred}
\bibinfo{author}{Yao, Y.}, \bibinfo{author}{Ye, D.}, \bibinfo{author}{Li, P.},
  \bibinfo{author}{Han, X.}, \bibinfo{author}{Lin, Y.}, \bibinfo{author}{Liu,
  Z.}, \bibinfo{author}{Liu, Z.}, \bibinfo{author}{Huang, L.},
  \bibinfo{author}{Zhou, J.}, \& \bibinfo{author}{Sun, M.}
  (\bibinfo{year}{2019}).
\newblock \bibinfo{title}{Docred: A large-scale document-level relation
  extraction dataset}.
\newblock In {\it \bibinfo{booktitle}{Proceedings of the 2019 Annual Meeting of
  the Association for Computational Linguistics}\/} (pp.
  \bibinfo{pages}{764--777}).
\bibitem[{Yu et~al.(2017)Yu, Yin, Hasan, dos Santos, Xiang \&
  Zhou}]{yu2017improved}
\bibinfo{author}{Yu, M.}, \bibinfo{author}{Yin, W.}, \bibinfo{author}{Hasan,
  K.~S.}, \bibinfo{author}{dos Santos, C.}, \bibinfo{author}{Xiang, B.}, \&
  \bibinfo{author}{Zhou, B.} (\bibinfo{year}{2017}).
\newblock \bibinfo{title}{Improved neural relation detection for knowledge base
  question answering}.
\newblock In {\it \bibinfo{booktitle}{Proceedings of the 55th Annual Meeting of
  the Association for Computational Linguistics (Volume 1: Long Papers)}\/}
  (pp. \bibinfo{pages}{571--581}).
\bibitem[{Zhang et~al.(2017{\natexlab{a}})Zhang, Zhang \& Fu}]{zhang2017end}
\bibinfo{author}{Zhang, M.}, \bibinfo{author}{Zhang, Y.}, \&
  \bibinfo{author}{Fu, G.} (\bibinfo{year}{2017}{\natexlab{a}}).
\newblock \bibinfo{title}{End-to-end neural relation extraction with global
  optimization}.
\newblock In {\it \bibinfo{booktitle}{Proceedings of the 2017 Conference on
  Empirical Methods in Natural Language Processing}\/} (pp.
  \bibinfo{pages}{1730--1740}).
\bibitem[{Zhang \& Ghorbani(2020)}]{zhang2020overview}
\bibinfo{author}{Zhang, X.}, \& \bibinfo{author}{Ghorbani, A.~A.}
  (\bibinfo{year}{2020}).
\newblock \bibinfo{title}{An overview of online fake news: Characterization,
  detection, and discussion}.
\newblock {\it \bibinfo{journal}{Information Processing \& Management}\/},
  {\it \bibinfo{volume}{57}\/}, \bibinfo{pages}{102025}.
\bibitem[{Zhang et~al.(2018)Zhang, Qi \& Manning}]{zhang2018graph}
\bibinfo{author}{Zhang, Y.}, \bibinfo{author}{Qi, P.}, \&
  \bibinfo{author}{Manning, C.~D.} (\bibinfo{year}{2018}).
\newblock \bibinfo{title}{Graph convolution over pruned dependency trees
  improves relation extraction}.
\newblock In {\it \bibinfo{booktitle}{Proceedings of the 2018 Conference on
  Empirical Methods in Natural Language Processing}\/} (pp.
  \bibinfo{pages}{2205--2215}).
\bibitem[{Zhang et~al.(2017{\natexlab{b}})Zhang, Zhong, Chen, Angeli \&
  Manning}]{zhang2017position}
\bibinfo{author}{Zhang, Y.}, \bibinfo{author}{Zhong, V.},
  \bibinfo{author}{Chen, D.}, \bibinfo{author}{Angeli, G.}, \&
  \bibinfo{author}{Manning, C.~D.} (\bibinfo{year}{2017}{\natexlab{b}}).
\newblock \bibinfo{title}{Position-aware attention and supervised data improve
  slot filling}.
\newblock In {\it \bibinfo{booktitle}{Proceedings of the 2017 Conference on
  Empirical Methods in Natural Language Processing}\/} (pp.
  \bibinfo{pages}{35--45}).

\end{thebibliography}

\end{document}